\documentclass{scrartcl}
\usepackage{amsmath}
\usepackage{amsfonts}
\usepackage{amssymb}
\usepackage{dsfont}
\usepackage[T1]{fontenc}
\usepackage[latin1]{inputenc}
\usepackage{epsfig}
\usepackage{graphicx}

\usepackage[boxed, lined]{algorithm2e}
\usepackage{float}
\usepackage{comment}
\usepackage{mathtools}

\newcommand{\bj}{\mathbf{j}}

\newcommand{\ba}{\mathbf{a}}

\newcommand{\be}{\mathbf{e}}
\newcommand{\bx}{\mathbf{x}}

\newcommand{\bw}{\mathbf{w}}

\newcommand{\bW}{\mathbf{W}}
\newcommand{\bV}{\mathbf{V}}

\newcommand{\D}{{\mathcal{D}}}

\newcommand{\W}{{\mathcal{W}}}

\newcommand{\Mu}{{\mathcal{M}}}

\newcommand{\btheta}{\mathbf{\vartheta}}
\newcommand{\bTheta}{\mathbf{\Theta}}

\newcommand{\N}{\mathbb{N}}

\newcommand{\R}{\mathbb{R}}
\newcommand{\Z}{\mathbb{Z}}
\newcommand{\Rd}{\mathbb{R}^d}

\newcommand{\beq}{\begin{eqnarray*}}

\newcommand{\eeq}{\end{eqnarray*}}

\newcommand{\beqm}{\begin{eqnarray}}

\newcommand{\eeqm}{\end{eqnarray}}

\newtheorem{theorem}{Theorem}

\newtheorem{lemma}{Lemma}
\newtheorem{definition}{Definition}
\newtheorem{remark}{Remark}

\newcommand{\EXP}{{\mathbf E}}
\newcommand{\PROB}{{\mathbf P}}

\renewcommand{\P}{{\cal P}}
\newcommand{\bf}{\normalfont \bfseries}
\newcommand{\it}{\normalfont \itshape}

\allowdisplaybreaks
\begin{document}
\renewcommand{\thefootnote}{\fnsymbol{footnote}}
\newcommand{\F}{{\cal F}}
\newcommand{\Sp}{{\cal S}}
\newcommand{\G}{{\cal G}}
\newcommand{\HH}{{\cal H}}

\begin{center}

  {\LARGE \bf
On the rate of convergence of an over-parametrized
Transformer classifier
learned by gradient descent
  }
\footnote{
Running title: {\it Over-parametrized transformers}}
\vspace{0.5cm}

Michael Kohler$^{1}$
and Adam Krzy\.zak$^{2,}$\footnote{Corresponding author. Tel:
  +1-514-848-2424 ext. 3007, Fax:+1-514-848-2830}\\

{\it $^1$
Fachbereich Mathematik, Technische Universit\"at Darmstadt,
Schlossgartenstr. 7, 64289 Darmstadt, Germany,
email: kohler@mathematik.tu-darmstadt.de}

{\it $^2$ Department of Computer Science and Software Engineering, Concordia University, 1455 De Maisonneuve Blvd. West, Montreal, Quebec, Canada H3G 1M8, email: krzyzak@cs.concordia.ca}

\end{center}
\vspace{0.5cm}

\begin{center}
June 20, 2024
\end{center}
\vspace{0.5cm}

\noindent
    {\bf Abstract}\\
Classification from independent and identically
distributed random variables is considered. 
Classifiers based on over-parametrized
transformer encoders are defined where
all the weights are learned by gradient descent. Under suitable
conditions on the a posteriori probability an upper bound on the
rate of convergence of the difference of the misclassification
probability of the estimate and the optimal misclassification probability
is derived.

    \vspace*{0.2cm}


\vspace*{0.2cm}


\section{Introduction}
\label{se1}

\subsection{Scope of this article}
One of the most recent and fascinating breakthroughs
in artificial intelligence is ChatGPT, a chatbot which
can simulate human conversation. ChatGPT is an instance
of GPT4, which is a language model based on
{\bf g}enerative {\bf p}redictive {\bf t}ransformers.
So if one wants to study from a theoretical point of view,
how powerful such artificial intelligence can be, one
approach is to consider transformer networks and to study which
problems one can solve with these networks theoretically.
Here it is not only important what kind of models these network
can {\it approximate}, or how they can {\it generalize} their
knowledge learned by choosing the best possible approximation to
a concrete data set, but also how well {\it optimization}
of such transformer network based on concrete data set works.
In this article we consider all these three different aspects
simultaneously and show a theoretical upper bound on
the missclassification probability of a transformer
network fitted to the observed data. For simplicity we focus in this
context on transformer encoder networks which can be applied to
define an estimate in the context of a classification
problem involving natural language.

\subsection{Pattern recognition}
We study these estimates in the context of pattern recognition.
Given $(X,Y)$, $(X_1,Y_1)$, \dots, $(X_n,Y_n)$ independent and identically distributed random variables
with values in $\R^{d \cdot l} \times \{-1,1\}$, and given the data set
\[
\D_n = \{ (X_1,Y_1), \dots, (X_n,Y_n)\}
\]
the goal is to construct a classifier
\[
\eta_n(\cdot) = \eta_n(\cdot,\D_n):\R^{d \cdot l} \rightarrow \{-1,1\}
\]
such that its misclassification probability
\[
\PROB \{ \eta_n(X) \neq Y | \D_n\}
\]
is as small as possible.
Here the predictor variable $X$ describes the encoding of a sequence
of length $l$ consisting of words or tokens, and each word or token
is encoded by a value in $\R^d$. The goal is to predict
the label $Y$ corresponding to the sentence described by $X$.

Let
\begin{equation}
\label{se1eq1}
m(x) = \PROB\{ Y=1|X=x\}
\quad
( x \in \R^{d \cdot l})
\end{equation}
be the a posteriori probability of class 1. Then
\[
\eta^*(x)=
\begin{cases}
  1, & \mbox{if } m(x) \geq \frac{1}{2} \\
  -1, & \mbox{elsewhere}
  \end{cases}
\]
is the Bayes classifier, i.e., the classifier satisfying 
\[
\PROB\{ \eta^*(X) \neq Y\}
=
\min_{\eta : \R^{d \cdot l} \rightarrow \{-1,1\}}
\PROB\{\eta(X) \neq Y\}
\]
(cf., e.g., Theorem 2.1 in Devroye, Gy\"orfi and Lugosi (1996)).

In this paper we derive upper bounds on
\begin{eqnarray}
  \label{se1eq2}
  &&
  \EXP \left\{
\PROB \{ \eta_n(X) \neq Y | \D_n\}
-
\PROB\{\eta^*(X) \neq Y\}
\right\}
\nonumber \\
&&
=
\PROB \{ \eta_n(X) \neq Y \}
-
\min_{\eta : \R^{d \cdot l} \rightarrow \{-1,1\}}
\PROB\{\eta(X) \neq Y\}
.
\end{eqnarray}

It is well-known that in order to derive
nontrivial rate of convergence results
on the difference between the misclassification probability of any
estimate and the minimal possible value it is necessary
to restrict the class of distributions (cf., e.g., Section 3.1 in
Gy\"orfi et al. (2002)).
In this context we will assume that the
a posteriori probability is smooth in the following sense:

\begin{definition}
\label{intde2} 
  Let $p=q+s$ for some $q \in \N_0$ and $0< s \leq 1$.
A function $m:\R^{d \cdot l} \rightarrow \R$ is called
$(p,C)$-smooth, if for every $\mathbf{\alpha}=(\alpha_1, \dots, \alpha_{d \cdot l}) \in
\N_0^{d \cdot l}$
with $\sum_{j=1}^{d \cdot l} \alpha_j = q$ the partial derivative
$\partial^q m/(\partial x_1^{\alpha_1}
\dots
\partial x_{d \cdot l}^{\alpha_{d \cdot l}}
)$
exists and satisfies
\[
\left|
\frac{
\partial^q m
}{
\partial x_1^{\alpha_1}
\dots
\partial x_{d \cdot l}^{\alpha_{d \cdot l}}
}
(x)
-
\frac{
\partial^q m
}{
\partial x_1^{\alpha_1}
\dots
\partial x_{d \cdot l}^{\alpha_{d \cdot l}}
}
(z)
\right|
\leq
C
\|\bold{x}-\bold{z}\|^s
\]
for all $\bold{x},\bold{z} \in \R^{d \cdot l}$, where $\Vert\cdot\Vert$ denotes the Euclidean norm.
\end{definition}

In order to show good rates of convergence
even for high-dimensional predictors we use a hierarchical composition
model as in Schmidt-Hieber (2020), where the a posteriori probability
is represented by a composition of several functions and where each
of these functions depends only on a few variables. We use the
following definition of Kohler and Langer (2021) to formalize
this assumption.

\begin{definition}
\label{de2}
Let $d,l \in \N$, $m: \R^{d \cdot l} \to \R$ and let
$\P$ be a subset
of $(0,\infty) \times \N$.\\
\noindent
\textbf{a)}
We say that $m$ satisfies a hierarchical composition model of level $0$
with order and smoothness constraint $\mathcal{P}$, if there exists $K \in \{1, \dots, {d \cdot l}\}$ such that
\[
m(\bold{x}) = x^{(K)} \quad \mbox{for all } \bold{x}
= (x^{(1)}, \dots, x^{({d \cdot l})})^{\top} \in \R^{d \cdot l}.
\]
\noindent
\textbf{b)}
Let $\kappa \in \N_0$.
We say that $m$ satisfies a hierarchical composition model
of level $\kappa+1$ with order and smoothness constraint $\mathcal{P}$, if there exist $(p,K)  \in \P$, $C>0$, \linebreak $g: \R^{K} \to \R$ and $f_{1}, \dots, f_{K}: \R^{d \cdot l} \to \R$, such that
$g$ is $(p,C)$--smooth,
$f_{1}, \dots, f_{K}$ satisfy a  hierarchical composition model of level $\kappa$
with order and smoothness constraint $\mathcal{P}$
and 
\[m(\bold{x})=g(f_{1}(\bold{x}), \dots, f_{K}(\bold{x})) \quad \mbox{for all } \bold{x} \in \R^{d \cdot l}.\]
\end{definition}
\noindent
Let $\HH(\kappa,\P)$ be the set of all functions $m:\R^{d \cdot l} \rightarrow \R$
which satisfy a  hierarchical composition model of level $\kappa$
with order and smoothness constraint $\mathcal{P}$.

A motivation of hierarchical models from an applied point of view
can be found in Kohler and Langer (2020a).

\subsection{Learning of a transformer encoder}
We apply gradient descent to an over-parametrized model of a transformer
encoder in order to learn its parameter. More precisely, let $\bTheta$
be the set of parameters of the transformer networks
$\{f_\btheta : \btheta \in \bTheta \}$ (which we will introduce
in detail in Section \ref{se2} below), and consider a linear
combination
\[
f(x)= f_{(w_k)_{k=1, \dots, K}, (\btheta_k)_{k=1, \dots, K}}
  (x)
  =
  \sum_{k=1}^{K} w_k \cdot f_{\btheta_k}(x)
\]
of transformer networks $f_{\btheta_k}$. Here $(w_k)_{k=1, \dots, K}$
are weights satisfying
\begin{equation}
  \label{se1eq1}
w_k \geq 0 \quad \mbox{and} \sum_{k=1}^K w_k =1,
\end{equation}
$\btheta_k$ are the weights of the transformer networks $f_{\btheta_k}$
$(k=1,\dots, K_n)$,
and by choosing $K$ very large our model becomes 
 over-parametrized in the sense that the number of its parameters
is much larger than the sample size. We will use
\[
\eta_n(X)=sgn(f(X))
\]
as our prediction of $Y$, and in order to achieve a small missclassification
probability our aim will be to choose the parameters
$(w_k)_{k=1, \dots, K}$ and $(\btheta_k)_{k=1, \dots, K}$ of $f$ such that its
  logistic loss
  \[
  \EXP \left\{
  \log \left(
1 + \exp(-Y \cdot f(X)
  \right)
  \right\}
  \]
  is small. To do this, we will randomly initialize its parameter
  in a proper way and then perform $t_n$ gradient descent steps
  in view of minimization of the empirical logistic loss
  \[
  \frac{1}{n}
  \sum_{i=1}^n
  \log \left(
1 + \exp(-Y_i \cdot f(X_i)
  \right),
  \]
  where proper projection steps will ensure that (\ref{se1eq1}) is satisfied
  and that the parameters $\btheta_k$ will not move away too far from
  their random starting values
  (see Section \ref{se2} for details).

\subsection{Main results}
 We show, that
in case that the a posteriori probability satisfies a hierarchical
composition model with smoothness and order constraint $\P$, the
corresponding estimate $\eta_n$ satisfies
\[
  \PROB\{ \eta_n(X) \neq Y \}
  -
  \min_{\eta:\R^{d \cdot l} \rightarrow \{-1,1\}}  \PROB\{ \eta(X) \neq Y \}
  \leq
  c_1 \cdot (\log n)^{3} \cdot
  \max_{(p,K) \in \P} n^{- \min \left\{ \frac{p}{2 \cdot (2p+K)}, \frac{1}{6} \right\}}.
\]
And if, in addition, 
\[
\PROB \left\{
\max
\left\{ \frac{\PROB\{Y=1|X\}}{1-\PROB\{Y=1|X\}},
\frac{1-\PROB\{Y=1|X\}}{\PROB\{Y=1|X\}} \right\}
> n^{1/3}
\right\}
 \geq
1 - \frac{1}{n^{1/3}}
\quad (n \in \N)
\]
holds (which implies that with high probability $\PROB\{Y=1|X\}$
is either close to one or close to zero)
then we show that the estimates achieve the improved rate of convergence
  \[
  \PROB\{ \eta_n(X) \neq Y \}
  -
  \min_{\eta:\R^{d \cdot l} \rightarrow \{-1,1\}}  \PROB\{ \eta(X) \neq Y \}
  \leq
  c_2 \cdot (\log n)^{6} \cdot
  \max_{(p,K) \in \P} n^{-  \min \left\{ \frac{p}{2p+K}, \frac{1}{3} \right\}}.
  \]
  In order to prove these results we derive a general result
  which gives an upper bound on the expected logistic loss of an
  over-parametrized linear combination of deep networks learned
  by minimizing an empirical logistic loss via gradient descent.
  In the proof of this result we show that the projection of the outer
  weights enables us to bound the generalization error of our
  over-parametrized linear combination of deep networks
  by the Rademacher complexity of a class of single deep networks.
  And in the application of this general result, we derive new approximation
  properties of Transformer networks with slightly disturbed weight
  matrices.
  
\subsection{Discussion of related results}
Transformers have been introduced 
by Vaswani et al. (2017). In applications they are usually
 combined with unsupervised pre-training and the same pre-trained
transformer encoder is then fine-tuned to a variety of
natural language processing tasks, see
Devlin et al. (2019).

Approximation and generalization of Transformer encoder networks
has been studied in Gurevych et al. (2022). Their estimates are
defined as plug-in classifiers of the least squares estimates
based on transformer networks, and similar rate of convergence results
as in the current paper are shown. The main difference between our
result in this paper and the result in Gurevych et al. (2022) is that
we define our estimates using gradient descent, and consequently we
have to take the optimization error into account too,
which forces us  to derive technically
much more complex approximation
results for transformer networks.

Much more is known about the deep neural network estimates.
There exist quite a few approximation
results for neural networks (cf., e.g.,
Yarotsky (2018),
Yarotsky and Zhevnerchute (2019),
Lu et al. (2020), Langer (2021) and the literature cited therein),
and generalization of deep neural networks can either
be analyzed within the framework of the classical VC theory
(using e.g. the result of
Bartlett et al. (2019) to bound the VC dimension of classes
of neural networks) or in case of over-parametrized deep
neural networks
(where the number of free parameters adjusted
to the observed data set is much larger than the sample size)
by using bounds on the Rademacher complexity
(cf., e.g., Liang, Rakhlin and Sridharan (2015), Golowich, Rakhlin and
Shamir (2019), Lin and Zhang (2019),
Wang and Ma (2022)
and the literature cited therein).

Combining such results leads to a rich theory
showing that owing to the network structure the least squares
neural network estimates can achieve suitable dimension
reduction in hierarchical composition models for the function to be
estimated. For a simple model this was first shown by Kohler and Krzy\.zak
(2017) for H\"older smooth function and later extended to arbitrary
smooth functions by Bauer and Kohler (2019). For a more complex
hierarchical composition model and the ReLU activation function this
was shown in Schmidt-Hieber (2020) under the assumption that the networks
satisfy some sparsity constraint. Kohler and Langer (2021) showed
that this is also possible for fully connected neural networks, i.e.,
without imposing a sparsity constraint on the network.
Adaptation of deep neural network to especially
weak smoothness assumptions was shown in
Imaizumi and Fukamizu (2018), Suzuki (2018) and Suzuki and Nitanda (2019).

Less well understood is the optimization of deep neural networks.
As was shown, e.g., in Zou et al. (2018), Du et al. (2019),
Allen-Zhu, Li and Song (2019) and Kawaguchi and Huang (2019)
 application of gradient descent to over-parameterized deep
neural networks 
leads to neural network which (globally) minimizes the empirical risk
considered. However, as was shown in Kohler and Krzy\.zak (2021),
the corresponding estimates do not behave well on new independent
data.
So the main question is why gradient descent (and its variants
like stochastic gradient descent) can be used to fit a neural
network to observed data in such a way that the resulting estimate
achieves good results on new independent data.
 The challenge here is not only to analyze optimization
but to consider it simultaneously with approximation and generalization.

In case of shallow neural networks (i.e., neural networks with only
one hidden layer) this has been done successfully in Braun et al. (2023).
Here it was possible to show that the classical dimension free
rate of convergence of Barron (1994) for estimation of a regression function
where its Fourier transform has a finite moment can also  be achieved by
shallow neural networks learned by gradient descent. The main idea
here is that the gradient descent  selects a subset of the neural
network where random initialization of the inner weights leads
to values with good approximation properties, and that it adjusts
the outer weights for these neurons properly. A similar idea
was also applied in Gonon (2021).
Kohler and Krzy\.zak
(2022) applied this idea in the context of over-parametrized deep
neural networks where a linear combination of a huge number of 
deep neural networks of fixed size are computed in parallel.
Here the gradient descent selects again a subset of the
neural networks computed in parallel and chooses a proper linear
combination of the networks. By using metric entropy bounds
(cf., e.g.,
Birman and Solomnjak (1967) and
Li, Gu and Ding (2021)) it is possible to control generalization of
the over-parametrized neural networks, and as a result the rate of
convergence of order close to $n^{-1/(1+d)}$ (or $n^{1/(1+d^*)}$ in case
of interaction models, where it is assumed that the regression function
is a sum of functions applied to only $d^*$ of the $d$ components
of the predictor variable) can be shown
for H\"older-smooth regression function
with H\"older exponent $p \in [1/2,1]$. Universal consistency of
such estimates for bounded $X$ was shown in Drews and Kohler (2022).

In all those results adjusting the inner weights with gradient
descent is not important. In fact, Gonon (2021)
does not do this at all, while Braun et al. (2023) and Kohler and Krzy\.zak
(2022) use that the relevant inner weights do not move too far away from
their starting values during gradient descent.
Similar ideas have also been applied in
Andoni et al. (2014) and Daniely (2017).
This whole approach is related to random feature networks
(cf., e.g., Huang, Chen and Siew (2006) and Rahimi and Recht (2008a, 2008b, 2009)),
where the inner weights are chosen randomly and
only the outer weights are learned during gradient descent.
Yehudai and Shamir (2022) present a lower bound which implies
that either the number of neurons or the absolute value
of the coefficients must grow exponential
in the dimension in order to learn a single ReLU neuron with
random feature networks. But since Braun et al. (2023) was able to
prove a useful rate of convergence result for networks similar
to random feature networks, the practical relevance of this lower
bound is not clear.

The estimates in Kohler and Krzy\.zak (2022) use a $L_2$
regularization on the outer weights during gradient descent.
As was shown in Drews and Kohler (2023), it is possible to
achieve similar results without $L_2$ regularization.

Often gradient descent in neural networks is studied in the
neural tangent kernel setting proposed by Jacot, Gabriel and Hongler (2020),
where instead of a neural network
estimate a kernel estimate is studied and its error is used
to bound the error of the neural network estimate. For further
results in this context see Hanin and Nica (2019) and the literature
cited therein.
Nitanda and Suzuki (2021) were able to analyze the global error
of 
an over-parametrized shallow neural network
learned by gradient descent based on this approach. However, due to the
use of the neural tangent kernel, also the smoothness assumption
of the function to be estimated has to be defined with the aid of
a norm involving the kernel, which does not lead to the classical
smoothness conditions of our paper. Another approach where the estimate
is studied  in some asymptotically equivalent model
is the  mean field approach, cf.,
Mei, Montanari, and Nguyen (2018), Chizat and Bach (2018) or Nguyen and Pham (2020).
A survey of various results on over-parametrized deep neural network
estimates learned by gradient descent can be found in
Bartlett, Montanari and Rakhlin (2021).

\subsection{Notation}
\label{se1sub5}
  The sets of natural numbers, natural numbers including zero, real numbers and nonnegative real numbers
  are denoted by $\N$, $\N_0$, $\R$ and $\R_+$, respectively.
We set $\bar{\R}=\R \cup \{-\infty,\infty\}$.
  For $z \in \R$, we denote
the smallest integer greater than or equal to $z$ by
$\lceil z \rceil$, and we set $z_+=\max\{z,0\}$ and
$z_-=\max\{-z,0\}$.
The Euclidean norm of $x \in \Rd$
is denoted by $\|x\|$ and for $x,z \in \Rd$ its scalar product
is denoted by $<x,z>$. For a closed and convex set $A \subseteq \R^d$
we denote by $Proj_A x$ that element $Proj_A x \in A$ such that
\[
\|x-Proj_A x\|= \min_{z \in A} \|x-z\|.
\]
For $f:\R^d \rightarrow \R$
\[
\|f\|_\infty = \sup_{x \in \R^d} |f(x)|
\]
is its supremum norm, and for $A \subseteq \Rd$ we set
\[
\|f\|_{\infty, A} = \sup_{x \in A} |f(x)|.
\]
For a vector $x=(x^{(1)}, \dots, x^{(d)})^T$ we denote by
\[
\|x\|_\infty = \max_{i=1, \dots, n} |x^{(i)}|
\]
its supremum norm, and if $A=(a_{i,j})_{i=1, \dots, I, j=1, \dots, J}$
we set
\[
\|A\|_\infty = \max_{i=1, \dots, I, j=1, \dots, J} |a_{i,j}|.
\]
For $\bj=(j^{(1)},\dots,j^{(d)}) \in \N_0^d$ we write
\[
\|\bj\|_1=j^{(1)}+ \dots + j^{(d)}
\]
and for $f:\R^d \rightarrow \R$ we set
\[
\partial^{\bj} f =
\frac{\partial^\|\bj\|_1 f}{(\partial x^{(1)})^{j^{(1)}} \dots
  (\partial x^{(d)})^{j^{(d)}}}.
\]
For $q \in \N_0$ and $f:\Rd \rightarrow \R$ we set
\[
\|f\|_{C^q(\Rd)}= \max\left\{ \|\partial^\bj f\|_\infty \, : \, \bj \in \N_0^d,
\, \|\bj\|_1 \leq q \right\}.
\]
Let $\F$ be a set of functions $f:\Rd \rightarrow \R$,
let $x_1, \dots, x_n \in \Rd$, set $x_1^n=(x_1,\dots,x_n)$ and let
$p \geq 1$.
A finite collection $f_1, \dots, f_N:\Rd \rightarrow \R$
  is called an $L_p$ $\varepsilon$--packing in $\F$ on $x_1^n$
  if $f_1, \dots, f_N \in \F$ and
  \[
  \min_{1 \leq i<j\leq N}
  \left(
  \frac{1}{n} \sum_{k=1}^n |f_i(x_k)-f_j(x_k)|^p
  \right)^{1/p} \geq \varepsilon
  \]
  holds.
  The $L_p$ $\varepsilon$--packing number of $\F$ on $x_1^n$
  is the  size $N$ of the largest $L_p$ $\varepsilon$--packing
  of $\F$ on $x_1^n$ and is denoted by $\Mu_p(\varepsilon,\F,x_1^n)$.

For $z \in \R$ and $\beta>0$ we define
$T_\beta z = \max\{-\beta, \min\{\beta,z\}\}$. If $f:\R^d \rightarrow
\R$
is a function  then we set
$
(T_{\beta} f)(x)=
T_{\beta} \left( f(x) \right)$.
For $z \in \bar{\R}$ we denote by
\[
sgn(z)=
\begin{cases}
  1 & \mbox{if } z >0,\\
  0 & \mbox{if } z =0,\\
  -1 & \mbox{if } z <0
  \end{cases}
\]
its sign. For $i,j \in \N_0$ we set
\[
\delta_{i,j}=
\begin{cases}
  1 & \mbox{if } i=j,\\
  0 & \mbox{if } i \neq j.
\end{cases}
\]

\subsection{Outline}
\label{se1sub6}
The over-parametrized transformer classifiers considered
in this paper are introduced in Section \ref{se2}. The main result
is presented in Section \ref{se3}. In Section \ref{se4}
we present a general result concerning the expected logistic loss
of an over-parametrized estimate defined by a
linear combination of deep networks. The proof of our main result is given in Section \ref{se5}.

\section{Definition of the estimate}
\label{se2}

\subsection{Topology of the Transformer networks}
\label{se2sub1}
Let $K_n \in \N$ be the number of transformer networks
which we compute in parallel.
The over-parametrized transformer networks which we use
for our classifier are of the form
\begin{equation}
  \label{se2eq1}
  f_{(w_k)_{k=1, \dots, K_n}, (\bW_k)_{k=1, \dots, K_n}, (\bV_k)_{k=1, \dots, K_n}}
  (x)
  =
  \sum_{k=1}^{K_n} w_k \cdot T_{\beta_n}( f_{\bW_k,\bV_k}(x)),
  \end{equation}
where the outer weights $(w_k)_{k=1,\dots,K_n}$ will be chosen such that
\begin{equation}
  \label{se2eq2}
  w_k \geq 0 \quad (k=1, \dots, K_n) \quad \mbox{and} \quad
  \sum_{k=1}^{K_n} w_k \leq 1
  \end{equation}
hold and where $\bW_k$ and $\bV_k$ are the weights used in the $k$-th
Transformer network $T_{\beta_n}(f_{\bW_k,\bV_k})$ and $\beta_n=c_3 \cdot \log n$. This Transformer network is
defined as follows:

For input
\[
x=(x_1, \dots, x_l) \in \R^{l \cdot d}
\]
it computes in a first step a new representation
\[
z_{k,0} = (z_{k,0,1}, \dots, z_{k,0,l }) \in \R^{d_{model} \times l}
\]
for some $d_{model} \in \N$ (which will be done in the same way for all
$k \in \{1, \dots, K_n\}$). Here $z_{k,0,j}$ is a new representation
of
$x_j \in \R^d$ of dimension
\begin{equation}
  \label{se2eq3}
  d_{model}
  =
  h \cdot I
\end{equation}
(where $h, I \in \N$ with $I \geq d+l+4$)
 which includes the
original data, coding of the position and additional
auxiliary values used for
later computation of function values.
More precisely, we set for  $s \in \{1, \dots, h \cdot I\}$
\[
z_{k,0,j}^{(s)}
=
\begin{cases}
  x_j^{(s)} & \mbox{if } s \in \{1, \dots, d\} \\
  1 & \mbox{if } s=d+1 \\
  \delta_{s-d-1,j} & \mbox{if } s \in \{d+2, \dots, d+1+l\}\\
  0 & \mbox{if } s \in \{d+l+2, d+l+3, \dots, h \cdot I\} \\
  \end{cases}
\]
For $d=2$, $l=4$, $I=10$ and $h=2$ the transformation of
the input is illustrated in Figure \ref{trans_input}.
\begin{figure}
\begin{equation*}
  x
  =
\begin{pmatrix}
x_1^{(1)} & x_2^{(1)} & x_3^{(1)} & x_4^{(1)}\\
x_1^{(2)} & x_2^{(2)} & x_3^{(2)} & x_4^{(2)}\\
\end{pmatrix}
\quad
\mapsto
\quad
z_{k,0}=
\begin{pmatrix}
x_1^{(1)} & x_2^{(1)} & x_3^{(1)} & x_4^{(1)} \\
x_1^{(2)} & x_2^{(2)} & x_3^{(2)} & x_4^{(2)}\\
1        &  1      &    1 & 1\\
1        &  0      &    0  & 0\\
0        &  1      &    0  & 0\\
0        &  0      &    1  & 0\\
0        &  0      &    0  & 1\\
0        &  0      &    0  & 0\\
\vdots   & \vdots  & \vdots & \vdots \\
0        &  0      &    0  & 0\\
\end{pmatrix}
\in \R^{20 \times 4}
\end{equation*}
\caption{Illustration of the transformation of the input
  in case $d=2$, $l=4$, $I=10$ and $h=2$.}
  \label{trans_input}
\end{figure}

After that  we compute successive
representations
\begin{equation}
  \label{se2eq4b}
z_{k,r}=(z_{k,r,1}, \dots, z_{k,r,l})  \in \R^{d_{model} \times l}
\end{equation}
of the input for $r=1, \dots, N$, and apply a feedforward
neural network to $z_{k,N}$. Here $z_{k,r}$ is the representation
of the input in the $k$-th transformer network in level $r$.
It depends on $l$ parts which correspond to $x_1$, \dots, $x_l$.
And
$N$ is the number
of pairs of attention and pointwise feedforward layers
of our transformer encoder.

Given $z_{k,r-1}$ for some $r \in \{1, \dots, N\}$
we compute $z_{k,r}$ by applying first
a multi-head attention and afterwards a pointwise feedforward neural
network with one hidden layer.
Both times we will use an additional
residual connection.

The computation of the multi-head attention depends on matrices
\begin{equation}
  \label{se2eq4}
  W_{query,k,r,s}, W_{key,k,r,s} \in \R^{d_{key} \times d_{model}}
    \quad \mbox{and} \quad
    W_{value,k,r,s} \in \R^{d_v \times d_{model}}
    \quad (s=1, \dots, h),
\end{equation}
where $h \in \N$ is the number of attentions which we compute
in parallel, where $d_{key} \in \N$ is the dimension of the queries
and the keys, and where $d_v=d_{model} /h= I$
is the dimension of the values. Here each of the $h$ attention heads
will be used to compute a new part of length $d_v=I$ of the
representation $z_{k,r,i}$ of $x_i$ for $i=1, \dots, l$.
 We use the above matrices to compute
 for each component $z_{k,r-1,i}$ of $z_{k,r-1}$
 (i.e., for each representation of $x_i$ at level
 $r-1$ $(i=1, \dots, l)$)
 corresponding queries
\begin{equation}
  \label{se2eq5}
  q_{k,r-1,s,i} = W_{query,k,r,s} \cdot z_{k,r-1,i},
\end{equation}
keys
\begin{equation}
  \label{se2eq6}
  k_{k,r-1,s,i} = W_{key,k,r,s} \cdot z_{k,r-1,i}
\end{equation}
and values
\begin{equation}
  \label{se2eq7}
  v_{k,r-1,s,i} = W_{value,k,r,s} \cdot z_{k,r-1,i}
\end{equation}
$(s \in \{1, \dots, h\}, i \in \{1, \dots, l\})$.
Then the so-called attention between the component
$i$ of $z_{k,r-1}$ and the component $j$ of $z_{k,r-1}$
(i.e., between the representations of $x_i$ and $x_j$ at
level $r-1$)
is defined
as the scalar product
\begin{equation}
  \label{se2eq8}
<q_{k,r-1,s,i}, k_{k,r-1,s,j}>
\end{equation}
of the corresponding query and key,
and the index $\hat{j}_{k,r-1,s,i}$ for which the maximal value occurs, i.e.,
\begin{equation}
  \label{se2eq9}
  \hat{j}_{k,r-1,s,i} = \arg \max_{j \in \{1, \dots, l\}}
  <q_{k,r-1,s,i}, k_{k,r-1,s,j}>,
\end{equation}
is determined. The value corresponding to  this index is multiplied with the
maximal attention in (\ref{se2eq8}) in order to define
\begin{eqnarray}
  \label{se2eq10}
  \hspace*{-0.5cm}
  \bar{y}_{k,r,s,i}&=&v_{k,r-1,s,\hat{j}_{k,r-1,s,i}} \cdot \max_{j \in \{1, \dots, l\}}
  <q_{k,r-1,s,i}, k_{k,r-1,s,j}> \nonumber \\
  \hspace*{-0.5cm}
  &=& v_{k,r-1,s,\hat{j}_{r-1,s,i}} \cdot  <q_{k,r-1,s,i}, k_{k,r-1,s,\hat{j}_{k,r-1,s,i}}>
\end{eqnarray}
$(s \in \{1, \dots, h\}, i \in \{1, \dots, l\})$.
Using a residual connection 
we  compute the output of the multi-head attention
by
\begin{equation}
  \label{se2eq10b}
  y_{k,r}=z_{k,r-1}+(\bar{y}_{k,r,1}, \dots, \bar{y}_{k,r,l})
\end{equation}
where
\[
\bar{y}_{k,r,i}
= (\bar{y}_{k,r,1,i}, \dots, \bar{y}_{k,r,h,i})
\in \R^{d_v \cdot h} = \R^{d_{model}}
\quad (i \in \{1, \dots, l\}).
\]
Here $y_{k,r} \in \R^{d_{model} \times l}$ has the same dimension as $z_{k,r-1}$.

The output of the pointwise feedforward neural network depends
on parameters
\begin{equation}
  \label{se2eq11}
  W_{k,r,1} \in \R^{d_{ff} \times d_{model}}, b_{k,r,1} \in \R^{d_{ff}},
  W_{k,r,2} \in \R^{d_{model} \times d_{ff}}, b_{k,r,2} \in \R^{d_{model}},
\end{equation}
which describe the weights in a feedforward neural network with
one hidden layer and $d_{ff} \in \N$ hidden neurons. This feedforward
neural network is applied to each component of (\ref{se2eq10b})
(which is analogous to a convolutionary neural network), i.e., to
each 
representation of $x_1$, \dots, $x_l$ computed up to this point
on level $r$, and computes
\begin{equation}
  \label{se2eq12}
  z_{k,r,s}=y_{k,r,s}+W_{k,r,2} \cdot \sigma \left(
  W_{k,r,1} \cdot y_{k,r,s} + b_{k,r,1}
  \right) + b_{k,r,2} \quad (s \in \{1, \dots, l\}),
\end{equation}
where we use again a residual connection.
Here
\[
\sigma(x) = \max\{x,0\}
\]
is the ReLU activation function, which is applied
to a vector by applying it
to each component
of the vector separately. After computing $z_{k,r,s}$
$(s \in \{1, \dots, l\})$ we define $z_{k,r}$
by (\ref{se2eq4b}).

Given the output $z_{k,N}$ of the sequence of $N$ multi-head
attention and pointwise feedforward layers, we apply
a (shallow) feedforward neural network with one hidden layer
and $J_n$ neurons to $z_{k,N,1}^{(d+l+2)}$, i.e., we set
\[
f_{\bV_k}(z_{k,N,1}^{(d+l+2)})
=
f_{net, J_n, \bV_k}(z_{k,N,1}^{(d+l+2)}),
\]
where for $z \in \R$
we define
\[
f_{net, J_n, \bV_k}(z)=
\sum_{j=1}^{J_n} v_{k,j}^{(1)} \cdot
\sigma \left(
v_{k,j,1}^{(0)} \cdot z 
+ v_{k,j,0}^{(0)}
\right).
\]
Here $\sigma(x)=\max\{x,0\}$ is again the ReLU activation function, and
\[
\bV_k
=
\left(
( v_{k,j}^{(1)})_{j=1, \dots, J_n},
( v_{k,j,1}^{(0)})_{j=1, \dots, J_n},
( v_{k,j,0}^{(0)})_{j=1, \dots, J_n}
\right)
\]
is the matrix of the weights of this feedforward neural network.

Because of
\begin{eqnarray*}
T_{\beta_n}(z)&=&\max\{-\beta_n, \min\{\beta_n,z\}\}
= \max\{0, \beta_n - \max\{-\beta_n,-z\}\} -\beta_n
\\
&=& \max\{0, 2 \beta_n - \max\{0,-z+ \beta_n\}\} -\beta_n
= \sigma(2 \beta_n -  \sigma( (-1) \cdot z + \beta_n)) - \beta_n,
\end{eqnarray*}
$z \mapsto T_{\beta_n}(z)$ is a neural network with two layers,
one hidden neuron per layer and ReLU activation function. This implies
that
$T_{\beta_n}(f_{\bV_k}(z))$ is a feedforward neural
network with $3$  hidden layers, $J_n$ neurons in layer
$1$ and one hidden neuron in layers $2$ and $3$, resp.

The output of our $k$-th transformer network is then
\[
T_{\beta_n} (f_{\bW_k,\bV_k}(x))
=
T_{\beta_n}(f_{\bV_k}(z_{k,N,1}^{(d+l+2)})),
\]
where
$z_{k,N,1}^{(d+l+2)}$
is one component of
 the output $z_{k,N}$ of the $N$ pairs of attention layers
and pointwise
feedforward layers.

\subsection{Initialization of the weights}
\label{se2sub2}
We initialize the weights $\bw^{(0)}=(w_k^{(0)},\bW_k^{(0)},\bV_k^{(0)})_{k=1, \dots, K_n}$ as
follows: We set
\[
w_k^{(0)} =0
\quad (k=1, \dots, K_n)
\]
and choose the components of all other weight matrices independently
from uniform distributions on the interval
\[
\left[ - c_{4} \cdot n^{c_{5}}, c_{4} \cdot n^{c_{5}} \right],
\]
where $c_{4}, c_{5}>0$ are suitably large constants.
After that we make a pruning step which depends on a parameter
$\tau \in \N$ chosen  in Theorem \ref{th1} below: We choose 
for each $k \in \{1, \dots, K_n\}$ and for each attention head
in each matrix in
each row $ \tau \in \N$ of its weights
randomly by independent uniform distributions and set all weights
not chosen to zero.
Similarly, we choose for each $k \in \{1, \dots, K_n\}$
and for each matrix $W_{k,r,1}$ in each row and for each matrix
$W_{k,r,2}$ in each column $\tau$ of its weights randomly by uniform
distributions and set all weights not chosen to zero.
Furthermore we set all entries
in
$W_{query,k,r,1}$ and $W_{key,k,r,1}$, all entries in the first
$d+l+1$ columns of $W_{k,r,2}$,
and all entries in the last two rows
of $W_{query,k,r,s}$ and $W_{key,k,r,s}$ in columns
greater than
$d+l+1$ to zero.

\subsection{Learning of the weights of the transformer network}
\label{se2sub3}
The aim in choosing the weights
$\bw=(w_k,\bW_k,\bV_k)_{k=1, \dots, K_n}$
of our transformer network is
the minimization of the empirical logistic loss. Let
\[
\varphi(z)=\log( 1 + \exp(-z))
\]
be the logistic loss (or cross entropy loss). Then the
empirical logistic loss of
$  f_{\bw}=f_{(w_k)_{k=1, \dots, K_n}, (\bW_k)_{k=1, \dots, K_n}, (\bV_k)_{k=1, \dots, K_n}}$
is defined by
\begin{equation}
  \label{se2eq13}
F_n(\bw)=
\frac{1}{n} \sum_{i=1}^n \varphi(Y_i \cdot  f_{\bw}(X_i)).
\end{equation}
We use gradient descent together with a projection step in order
to minimize (\ref{se2eq13}). Let $A$ be the set of all
$(w_k)_{k=1, \dots, K_n}$ which satisfy (\ref{se2eq2}) and let $B$
be the set of all $(\bW_k, \bV_k)_{k=1, \dots, K_n}$
which have nonzero components only in components which
have not been set to zero in the pruning step of the
initialization of the weights
 and which satisfy
\begin{equation}
  \label{se2eq14}
\| (\bW_k, \bV_k)_{k=1, \dots, K_n} - (\bW_k^{(0)}, \bV_k^{(0)})_{k=1, \dots, K_n}\| \leq c_6,
\end{equation}
where $c_6>0$ is a constant which will be chosen sufficiently small
in Theorem \ref{th1} below.
Let $\lambda_n>0$ be the stepsize of the gradient descent
and let
$\bw^{(0)}=(\bw_k^{(0)},\bW_k^{(0)},\bV_k^{(0)})_{k=1, \dots, K_n}$
be defined as in Subsection \ref{se2sub2}. Then we define
$\bw^{(t)}=(\bw_k^{(t)},\bW_k^{(t)},\bV_k^{(t)})_{k=1, \dots, K_n}$
recursively by setting
\begin{eqnarray*}
  &&
  \left(
  w_k^{(t+1)}
  \right)_{k=1, \dots, K_n}
  =
  Proj_A \left(
  \left(
  w_k^{(t)}
-
\lambda_n
\cdot
\frac{\partial F_n(\bw^{(t)})}{
\partial
w_k
}
 \right)_{k=1, \dots, K_n}
\right)
\end{eqnarray*}
and
\begin{eqnarray*}
  &&
  (\bW_k^{(t+1)}, \bV_k^{(t+1)})_{k=1, \dots, K_n}
  \\
  &&
  =
  Proj_B \left(
  \left(
(\bW_k^{(t)}, \bV_k^{(t)})
-
\lambda_n
\cdot
\nabla_{
(\bW_k, \bV_k)
}
(
F_n(\bw^{(t)}))
\right)_{k=1, \dots, K_n}
\right)
\end{eqnarray*}
$(t=0, \dots, t_n-1)$, where $t_n \in \N$ is the number of gradient
descent steps which will be chosen in Theorem \ref{th1} below.

\subsection{Definition of the estimate}
\label{se2sub4}
We define our estimate as the plug-in classifier
corresponding to the over-parametrized Transformer network
network with weight vector $\bw^{(\hat{t})}$ where $\hat{t} \in \{0,1, \dots, t_n\}$ is the index for which the empirical logistic loss is minimal during the
training, i.e., we set
\begin{equation}
  \label{se2eq15}
  \hat{t} = \arg \min_{t \in \{0,1, \dots, t_n\}}
F_n(\bw^{(t)}),
  \end{equation}
\begin{equation}
\label{se2eq16}
f_n(x)= f_{\bw^{(\hat{t})}}(x)
\end{equation}
and
\begin{equation}
\label{se2eq17}
\eta_n(x)= sgn(f_n(x)).
\end{equation}

\section{Main result}
\label{se3}

Our main result is the following bound on the difference of the
misclassification probability of our estimate and the minimal 
misclassification probability.

\begin{theorem}
  \label{th1}
  Let $A \geq 1$.
  Let $(X,Y)$, $(X_1, Y_1)$, \dots, $(X_n,Y_n)$ be
  independent and identically distributed $[-A,A]^{d \cdot l} \times \{-1,1\}$--valued
  random variables, and let \linebreak
  $m(x)= \PROB\{Y=1|X=x\}$ be the corresponding
  a posteriori probability. Let $\P$ be a finite subset
  of $[1,\infty) \times \N$ and assume that $m$ satisfies a hierarchical
  composition model with some finite level and smoothness and order constraint
  $\P$ and that all functions $g:\R^K \rightarrow \R$ in this hierarchical
  composition model are Lipschitz continuous and satisfy
  \[
\|g\|_{C^q(\R^K)} \leq c_7 < \infty,
  \]
  where $p=q+s$ with $s \in (0,1]$ and $q \in \N_0$
  (here $(p,K) \in \P$ is the smoothness and order
  corresponding to $g$ in the hierarchical composition model). 
Let $K_n \in \N$ be such that
\begin{equation}
  \label{th1eq2}
\frac{K_n}{e^{(\log n)^3 \cdot\sqrt{n}}} \rightarrow \infty
\quad (n \rightarrow \infty).
\end{equation}
Set $\beta_n= c_3 \cdot \log n$ for some $c_3 \geq 1$,
  \[
  h= \left\lceil \max_{(p,K) \in \P} n^{\frac{K}{2p+K}} \right\rceil,
  \quad
  d_{ff}= 2 \cdot h + 2,
  \quad
  I=\lceil \log n \rceil,
  \quad
J_n = \lceil c_8 \cdot n^{1/3} \rceil,
 \quad
t_n = n \cdot K_n
\]
and
\[
\lambda_n=\frac{1}{t_n},
\]
choose $\tau \in \{l+1, l+2,, \dots, l+d+1\}$ and choose
$N \in \N$ sufficiently large,
$c_6>0$ sufficiently small, $c_{4}, c_{5}>0$ sufficiently large,
  $d_{key} \geq 4$,  and define the estimate
  $\eta_n$ as in Section \ref{se2}.

  \noindent
  {\bf a)}
  We have
  for $n$ sufficiently large
  \[
  \PROB\{ \eta_n(X) \neq Y \}
  -
  \min_{\eta:\R^{d \cdot l} \rightarrow \{-1,1\}}  \PROB\{ \eta(X) \neq Y \}
  \leq
  c_9 \cdot (\log n)^{3} \cdot
  \max_{(p,K) \in \P} n^{- \min \left\{ \frac{p}{2 \cdot (2p+K)}, \frac{1}{6} \right\}}.
  \]
  
  \noindent
  {\bf b)}
    If, in addition, 
    \begin{equation}
\label{th1eq1}
\PROB \left\{
\max
\left\{
\frac{\PROB\{Y=1|X\}}{1-\PROB\{Y=1|X\}}
, \frac{1-\PROB\{Y=1|X\}}{\PROB\{Y=1|X\}}
\right\}
> n^{1/3}
\right\}
 \geq
1 - \frac{1}{n^{1/3}}
\quad (n \in \N)
\end{equation}
    holds, then we have
  for $n$ sufficiently large
  \[
  \PROB\{ \eta_n(X) \neq Y \}
  -
  \min_{\eta:\R^{d \cdot l} \rightarrow \{-1,1\}}  \PROB\{ \eta(X) \neq Y \}
  \leq
  c_{10} \cdot (\log n)^{6} \cdot
  \max_{(p,K) \in \P} n^{-  \min \left\{ \frac{p}{(2p+K)}, \frac{1}{3} \right\}}.
  \]
 
\end{theorem}

\noindent
\begin{remark}
The upper bound in parts a) and b) of Theorem \ref{th1}
do not depend on the dimension $d \cdot l$ of $X$, hence
the Transformer encoder estimate is able to circumvent the
curse of dimensionality in case that the a posteriori probability
satisfies a suitable hierarchical composition model.
\end{remark}

\noindent
\begin{remark}
In the definition of the estimate we use twice  a projection
step in the definition of the gradient descent. Here the projection
on the outer weights $(w_k)_{k=1, \dots, K_n}$ is our main tool
which enables us to show that the over-parametrization of the estimate
does not hurt the generalization. The second projection is used
to ensure that the change of the inner weights during gradient
descent does not hurt the approximation properties of the estimate.
For neural networks with smooth activation function it is possible to
show that such a projection step is automatically satisfied during
gradient descent steps for suitable chosen stepsizes and number
of gradient descent steps, cf. Lemma 1 in Drews and Kohler (2023).
\end{remark}

\noindent
\begin{remark}
The proof of Theorem \ref{th1} implies that the result also holds
for an estimate where gradient descent is only applied to the outer
weights $(w_k)_{k=1, \dots, K_n}$ and for all other weights their
initial randomly chosen values are not changed. Consequently, our
estimate is based on representation guessing and not representation
learning.
\end{remark}

\noindent
\begin{remark}
By assumption (\ref{th1eq2}) the number of parameters of our
estimate grows exponential in the sample size, so as in many
modern applications of deep learning our estimate uses a massive
over-parametrization.
\end{remark}

\section{A general result}
\label{se4}
\label{se4sub1}

Let $W \in \N$ and let
$\bTheta \subseteq \R^W$ be a closed and convex set of parameter values
(weights) for
a deep network of a given topology.
In the sequel we assume that our aim is to learn the parameter
$\btheta \in \bTheta$ (vector of weights) for a  deep network
\[
f_\btheta: \R^{d \cdot l} \rightarrow \R
\]
from the data $\D_n$ such that
\[
sgn(f_\btheta(x))
\]
is a good classifier. We do this by considering linear
combinations
\begin{equation}
  \label{se4sub1eq1}
f_{(\bw,\btheta)}(x)= \sum_{k=1}^{K_n} w_k \cdot T_{\beta_n}(f_{\btheta_k}(x))
  \end{equation}
of truncated versions of estimates $f_{\btheta_k}(x)$ $(k=1, \dots, K_n)$,
where $\bw=(w_k)_{k=1, \dots, K_n}$ satisfies
\begin{equation}
  \label{se4sub1eq2}
  w_k \geq 0 \quad (k=1, \dots, K_n) \quad \mbox{and} \quad
  \sum_{k=1}^{K_n} w_k \leq 1
  \end{equation}
and $\btheta=(\btheta_1, \dots, \btheta_{K_n}) \in \bTheta^{K_n}$.
Observe that by choosing $w_1=1$ and $w_k=0$ for $k>1$ we get
\[
f_{(\bw,\btheta)}(x)=T_{\beta_n}(f_{\btheta_1}(x))
\]
and in this way we can construct an estimate which satisfies
\[
sgn(f_{(\bw,\btheta)}(x))
=
sgn(f_{\btheta_1}(x))
\]
for any $\btheta_1 \in \bTheta$. And by choosing $K_n$ very large
our estimate will be over-parametrized in the sense that the number
of parameters of the estimate is much larger than the sample size.

Let
\[
\varphi(z)=\log( 1 + \exp(-z))
\]
be the logistic loss (or cross entropy loss)
and let $m(x)=\PROB\{Y=1|X=x\}$. Then
\[
f_{\varphi^*}(x)=
\begin{cases}
  \infty & \mbox{if } m(x)=1, \\
  \log \frac{m(x)}{1-m(x)} & \mbox{if } 0 < m(x) < 1, \\
  -\infty & \mbox{if } m(x)=0\\
  \end{cases}
\]
minimizes the expected logistic loss, i.e.,
\begin{eqnarray*}
  &&
\EXP\{ \varphi( Y \cdot f_{\varphi^*}(X)) \}
\\
&&
=
\EXP\left\{
m(X) \cdot \log( 1 + \exp(-  f_{\varphi^*}(X) ))
+
(1-m(X)) \cdot \log( 1 + \exp( f_{\varphi^*}(X) ))
\right\}
\\
&&
=
\min_{
f:\R^{d \cdot l} \rightarrow \bar{\R}
}
\EXP\{ \varphi( Y \cdot f(X)) \}
\end{eqnarray*}
holds. Because of
\[
sgn( f_{\varphi^*}(x))
=
\begin{cases}
  1 & \mbox{if } m(x)>\frac{1}{2}, \\
  -1 & \mbox{if } m(x)<\frac{1}{2},
  \end{cases}
\]
this implies
\[
\PROB\{ sgn(f_{\varphi^*}(X)) \neq Y\}
=
\min_{f: \R^{d \cdot l} \rightarrow \{-1,1\}}
\PROB\{ f(X) \neq Y\},
\]
i.e., we can compute the optimal predictor of $Y$ given $X$
by minimizing
the expected logistic loss.

Our aim in choosing
$(\bw,\btheta)$ is the minimization of the empirical logistic loss
\[
F_n((\bw,\btheta))
=
\frac{1}{n} \sum_{i=1}^n \varphi(Y_i \cdot f_{(\bw,\btheta)}(X_i)).
\]
In order to achieve this, we start with a random
initialization of $(\bw,\btheta)$: We choose
\begin{equation}
  \label{se4sub1eq3}
\btheta_1^{(0)}, \dots, \btheta_{K_n}^{(0)}
\end{equation}
randomly from some set $\bTheta^0 \subseteq \bTheta$ such that
the random variables in (\ref{se4sub1eq3}) are independent and also
independent from $(X,Y)$, $(X_1,Y_1)$, \dots, $(X_n,Y_n)$, and we set
\[
w_k^{(0)}=0 \quad (k=1, \dots, K_n).
\]
Then we perform $t_n \in \N$ gradient descent steps starting with
\[
\btheta^{(0)}=(\btheta_1^{(0)}, \dots, \btheta_{K_n}^{(0)})
\quad \mbox{and} \quad \bw^{(0)}=(w_1^{(0)}, \dots, w_{K_n}^{(0)}).
\]
To do this, we choose a stepsize $\lambda_n>0$ and set
\begin{eqnarray*}
  \bw^{(t+1)}
  &=&
  Proj_A \left(
 \bw^{(t)} - \lambda_n \cdot \nabla_\bw F_n((\bw^{(t)},\btheta^{(t)}))
 \right), \\
 \btheta^{(t+1)}
 &=&
 Proj_B\left(
 \btheta^{(t)} - \lambda_n \cdot \nabla_\btheta F_n((\bw^{(t)},\btheta^{(t)}))
 \right) \\
\end{eqnarray*}
for $t=1, \dots, t_n$.
Here $A$ is the set of all $\bw$ which satisfy (\ref{se4sub1eq2}), and
\[
B= \left\{
\btheta \in \bTheta^{K_n} \, : \, \|\btheta-\btheta^{(0)}\| \leq c_6
\right\},
\]
where $c_6>0$ is a constant,
and $Proj_A$ and $Proj_B$ is the $L_2$ projection on the
closed and convex sets $A$ and $B$. (Here closeness and convexity
of $B$ is implied by the closeness and convexity of $\bTheta$.)
Our estimate is then defined by
\begin{equation}
  \label{se4sub1eq4}
  \hat{t} = \arg \min_{t \in \{0,1, \dots, t_n\}}
F_n((\bw^{(t)},\btheta^{(t)}))
  \end{equation}
and
\begin{equation}
\label{se4sub1eq5}
f_n(x)= f_{(\bw^{(\hat{t})},\btheta^{(\hat{t})})}(x).
\end{equation}

\begin{theorem}
  \label{th2}
  Let $(X,Y)$, $(X_1,Y_1)$, \dots, $(X_n,Y_n)$
  be independent and identically distributed random variables
  with values in $\R^{d \cdot l} \times \{-1,1\}$.
Let $t_n, N_n, I_n \in \N$,
  set
  \[
\lambda_n=\frac{1}{t_n}, \quad K_n = N_n \cdot I_n,
\]
choose $c_6>0$,
and define the estimate $f_n$ as above.

Let
$\bTheta^* \subset \bTheta^0$ and set
\[
\bar{\bTheta}= \left\{ \btheta \in \bTheta \; : \,
\inf_{\tilde{\btheta} \in \bTheta^0} \|\btheta - \tilde{\btheta}\| \leq c_6 \right\}.
\]
Let $C_n, D_n \geq 0$ and assume
\begin{equation}
  \label{th2eq1}
  \| f_\btheta - f_{\btheta^*} \|_{\infty,supp(X)}
  \leq
  C_n \cdot \| \btheta-\btheta^*\|
\end{equation}
for all 
$\btheta^* \in \bTheta^*$ and all
$\btheta \in \{ \bar{\btheta} \in \bTheta \, : \,
\| \bar{\btheta} - \btheta^* \| \leq c_6 \}$,
\begin{equation}
  \label{th2eq2}
  \epsilon_n = \PROB \left\{
\btheta^{(0)}_1 \in \bTheta^*
\right\}
>0,
\end{equation}
\begin{equation}
  \label{th2eq3}
N_n \cdot (1- \epsilon_n)^{I_n} \leq \frac{1}{n} 
\end{equation}
and
\begin{equation}
  \label{th2eq4}
  \| \nabla_{\bw} F_n((\bw,\btheta)) \| \leq D_n
  \quad
  \mbox{for all }
  \bw \in A, \btheta \in \bar{\bTheta}.
\end{equation}
Then we have
\begin{eqnarray*}
  &&
  \EXP \left\{ \varphi(Y \cdot f_n(X)) \right\}
  -
  \min_{f : \R^{d \cdot l} \rightarrow \bar{\R}}
  \EXP \left\{ \varphi(Y \cdot f(X)) \right\}
 \\
  &&
  \leq
 c_{11} \cdot  \Bigg(
  \frac{\log n}{n}
  +
 \EXP \left\{
  \sup_{\btheta \in \bar{\bTheta}}
  \left|
  \frac{1}{n} \sum_{i=1}^n \epsilon_i \cdot T_{\beta_n}(f_\btheta(X_i))
  \right|
  \right\}
  +
  \frac{C_n +1}{\sqrt{N_n}}
  +
  \frac{D_n^2}{t_n}
  \\
  &&
  \hspace*{3cm}
  +
  \sup_{\btheta \in \bTheta^*}
  \EXP \left\{ \varphi(Y \cdot T_{\beta_n} f_\btheta(X)) \right\}
  -
  \min_{f : \R^{d \cdot l} \rightarrow \bar{\R}}
  \EXP \left\{ \varphi(Y \cdot f(X)) \right\}
  \Bigg),
\end{eqnarray*}
where $\epsilon_1$, \dots, $\epsilon_n$ are independent and
uniformly distributed
on $\{-1,1\}$ (so-called Rademacher random variables)
and independent from $X_1, \dots, X_n$.
  \end{theorem}

\noindent
\begin{remark}
    In Theorem \ref{th2} the Rademacher complexity
    \[
     \EXP \left\{
  \sup_{\btheta \in \bar{\bTheta}}
  \left|
  \frac{1}{n} \sum_{i=1}^n \epsilon_i \cdot T_{\beta_n}(f_\btheta(X_i))
  \right|
  \right\}
  \]
  is used to control the generalization error of the estimate,
  \[
  \sup_{\btheta \in \bTheta^*}
  \EXP \left\{ \varphi(Y \cdot f_\btheta(X)) \right\}
  -
  \min_{f : \R^{d \cdot l} \rightarrow \bar{\R}}
  \EXP \left\{ \varphi(Y \cdot f(X)) \right\},
  \]
  which describes the worst error occuring in the set
  $\Theta^*$ of ''good'' parameter values, is used to measure
  the approximation error, and
  \[
  \frac{C_n +1}{\sqrt{N_n}}
  +
  \frac{D_n^2}{t_n}
  \]
  is used to bound the error occuring due to gradient descent.
\end{remark}

    \noindent
        {\bf Proof of Theorem \ref{th2}.}
        Let $E_n$ be the event that there exist pairwise distinct
        $j_1, \dots, j_{N_n} \in \{1, \dots, K_n\}$ such that
        \[
\btheta_{j_i}^{(0)} \in \bTheta^*
       \]
       holds for all $i=1, \dots, N_n$. If $E_n$ holds set
       \[
       w_{j_i}^*= \frac{1}{N_n} \quad (i=1, \dots, N_n)
       \quad \mbox{and} \quad
       w_k^*=0 \quad (k \in \{1, \dots, K_n\} \setminus \{j_1, \dots, j_{N_n} \})
       \]
       and $\bw^*=(w_k^*)_{k=1, \dots, K_n}$, otherwise set $\bw^*=0$.

       We will use the following error decomposition:
       \begin{eqnarray*}
         &&
           \EXP \left\{ \varphi(Y \cdot f_n(X)) \right\}
  -
  \min_{f : \R^{d \cdot l} \rightarrow \bar{\R} }
  \EXP \left\{ \varphi(Y \cdot f(X)) \right\}
  \\
  &&
  = \EXP \left\{ \varphi(Y \cdot f_n(X)) \cdot 1_{E_n^c}\right\}
  \\
  &&
  \quad
  + 
  \EXP \left\{
  \left(
\EXP \left\{  
\varphi(Y \cdot f_n(X)) 
\big| \btheta^{(0)}, \D_n \right\}
-
\frac{1}{n} \sum_{i=1}^n \varphi(Y_i \cdot f_n(X_i))
\right) \cdot 1_{E_n}
\right\}
\\
&&
\quad
+
\EXP \left\{ \frac{1}{n} \sum_{i=1}^n \varphi(Y_i \cdot f_n(X_i))\cdot 1_{E_n}
\right\}
-
  \min_{f : \R^{d \cdot l} \rightarrow \bar{\R} }
  \EXP \left\{ \varphi(Y \cdot f(X)) \right\}
  \\
  &&
  =: T_{1,n} + T_{2,n} + T_{3,n}.
       \end{eqnarray*}
       In the {\it first step of the proof} we show
       \begin{equation}
         \label{pth1eq1}
         \PROB\{E_n^c\} \leq \frac{1}{n}.
         \end{equation}
       To do this we consider a sequential choice of the
       initial weights $\btheta_1^{(0)}$, \dots, $\btheta_{K_n}^{(0)}$.
       By definition of $\epsilon_n$ we know that the probability
       that none of $\btheta_1^{(0)}$, \dots, $\btheta_{I_n}^{(0)}$
       is contained in $\bTheta^*$ is given by
       \[
(1-\epsilon_n)^{I_n}.
       \]
       This implies that the probability that there exists $l \in \{1, \dots, N_n\}$
       such that  none of $\btheta_{(l-1)*I_n+1}^{(0)}$,
       \dots, $\btheta_{l \cdot I_n}^{(0)}$
       is contained in $\bTheta^*$ is upper bounded by
       \[
N_n \cdot (1-\epsilon_n)^{I_n}.
       \]
       Using (\ref{th2eq3}) we can conclude
       \[
\PROB\{ E_n^c \} \leq N_n \cdot (1-\epsilon_n)^{I_n} \leq \frac{1}{n}.
\]

In the {\it second step of the proof} we show
\[
T_{1,n} \leq c_{12} \cdot \frac{(\log n)}{n}.
\]
To do this, we observe that for $|z| \leq \beta_n$ we have
\[
\varphi(z)=\log( 1 + \exp(-z))
\leq (\log 4) \cdot I_{\{z>-1\}} + \log( 2 \cdot \exp(-z)) \cdot I_{\{z \leq -1\}}
\leq
3 + |z| \leq c_{13} \cdot \log n,
\]
from which we can conclude by the first step of the proof
\[
T_{1,n} \leq c_{13} \cdot (\log n) \cdot \PROB\{E_n^c\}
\leq c_{13} \cdot \frac{(\log n)}{n}.
\]

Let $\F$ be the set of all $f_{(\bw,\btheta)}$ where $\bw \in A$ and
$\btheta \in \bar{\bTheta}^{K_n}$. In the
{\it third step of the proof} we show
\[
T_{2,n}
\leq \EXP \left\{
 \EXP \left\{
    \sup_{f \in \F}
    \left(
    \EXP \{ \varphi( f(X) \cdot Y) \}
    -
    \frac{1}{n} \sum_{i=1}^{n} \varphi( f(X_i) \cdot Y_i)
    \right)
    \right\}
    \cdot 1_{E_n}
\right\}.
\]
This follows from
\begin{eqnarray*}
  T_{2,n}
  &=&
  \EXP \left\{
  \EXP \left\{
  \left(
\EXP \left\{  
\varphi(Y \cdot f_n(X)) 
\big| \btheta^{(0)}, \D_n \right\}
-
\frac{1}{n} \sum_{i=1}^n \varphi(Y_i \cdot f_n(X_i))
\right)
\big| \btheta^{(0)} \right\}
\cdot 1_{E_n}
\right\}
\\
&\leq&
\EXP \left\{
 \EXP \left\{
    \sup_{f \in \F}
    \left(
    \EXP \{ \varphi( f(X) \cdot Y) \}
    -
    \frac{1}{n} \sum_{i=1}^{n} \varphi( f(X_i) \cdot Y_i)
    \right)
\bigg| \btheta^{(0)} \right\}    
    \cdot 1_{E_n}
    \right\}
    \\
    &=&
\EXP \left\{
 \EXP \left\{
    \sup_{f \in \F}
    \left(
    \EXP \{ \varphi( f(X) \cdot Y) \}
    -
    \frac{1}{n} \sum_{i=1}^{n} \varphi( f(X_i) \cdot Y_i)
    \right)
    \right\}
    \cdot 1_{E_n}
\right\}.    
\end{eqnarray*}
Here the first inequality followed from
$\bw^{(t)} \in A$ and
$\btheta^{(t)} \in \bar{\bTheta}^{K_n}$
$(t \in \{0,1, \dots, t_n\})$.
       
            In the {\it fourth step of the proof} we show
   \begin{eqnarray}
    &&
    \EXP \left\{
    \sup_{f \in \F}
    \left(
    \EXP \{ \varphi( f(X) \cdot Y) \}
    -
    \frac{1}{n} \sum_{i=1}^{n} \varphi( f(X_i) \cdot Y_i)
    \right)
    \right\}
    \nonumber \\
    && \leq
    2 \cdot
    \EXP \left\{
    \sup_{f \in \F}
    \frac{1}{n} \sum_{i=1}^n \epsilon_i \cdot f(X_i)
    \right\}.
    \label{ple3eq1}
   \end{eqnarray}
  
   Choose random variables
   $(X_1^\prime, Y_1^\prime)$, \dots, $(X_n^\prime,Y_n^\prime)$ such that
   \[
   (X_1,Y_1), \dots, (X_n,Y_n), \epsilon_1, \dots, \epsilon_n,
   (X_1^\prime, Y_1^\prime), \dots, (X_n^\prime,Y_n^\prime)
   \]
   are independent and such that
    \[
   (X_1,Y_1), \dots, (X_n,Y_n), 
   (X_1^\prime, Y_1^\prime), \dots, (X_n^\prime,Y_n^\prime)
   \]
   are identically distributed and set $(X,Y)_1^n =((X_1,Y_1), \dots, (X_n,Y_n))$.
   We have
   \begin{eqnarray*}
     &&
  \EXP \left\{
    \sup_{f \in \F}
    \left(
    \EXP \{ \varphi( f(X) \cdot Y) \}
    -
    \frac{1}{n} \sum_{i=1}^{n} \varphi( f(X_i) \cdot Y_i)
    \right)
    \right\}
    \\
    &&
    =
      \EXP \left\{
    \sup_{f \in \F}
    \left(
    \EXP \{
    \frac{1}{n} \sum_{i=1}^n 
\varphi( f(X_i^\prime) \cdot Y_i^\prime)
    | (X,Y)_1^n
    \}
    -
    \frac{1}{n} \sum_{i=1}^{n} \varphi( f(X_i) \cdot Y_i)
    \right)
    \right\}
    \\
    &&
    \leq
    \EXP \left\{
    \EXP \left\{
    \sup_{f \in \F}
    \left(
    \frac{1}{n} \sum_{i=1}^n
\varphi( f(X_i^\prime) \cdot Y_i^\prime)
    -
    \frac{1}{n} \sum_{i=1}^{n} \varphi( f(X_i) \cdot Y_i)
    \right)
      | (X,Y)_1^n
    \right\}
    \right\}
    \\
    &&
    =
        \EXP \left\{
    \sup_{f \in \F}
    \left(
    \frac{1}{n} \sum_{i=1}^{n}
\varphi( f(X_i^\prime) \cdot Y_i^\prime)
    -
    \frac{1}{n} \sum_{i=1}^{n} \varphi( f(X_i) \cdot Y_i)
    \right)
    \right\}.
     \end{eqnarray*}
   Since the joint distribution of $(X_1,Y_1), \dots, (X_n,Y_n),
   (X_1^\prime, Y_1^\prime), \dots, (X_n^\prime,Y_n^\prime)$ does not change
   if we (randomly) interchange $(X_i,Y_i)$ and $(X_i^\prime, Y_i^\prime)$,
   the last term is equal to
   \begin{eqnarray*}
     &&
        \EXP \left\{
    \sup_{f \in \F}
    \left(
    \frac{1}{n} \sum_{i=1}^n
\epsilon_i \cdot \left(
\varphi( f(X_i^\prime) \cdot Y_i^\prime)
-
\varphi( f(X_i) \cdot Y_i)
    \right)
    \right)
    \right\}
    \\
    &&
    \leq
        \EXP \left\{
    \sup_{f \in \F}
    \left(
    \frac{1}{n} \sum_{i=1}^n
\epsilon_i \cdot 
\varphi( f(X_i^\prime) \cdot Y_i^\prime)
    \right)
    \right\}
    +
        \EXP \left\{
    \sup_{f \in \F}
    \left(
    \frac{1}{n} \sum_{i=1}^n
(-\epsilon_i) \cdot 
\varphi( f(X_i) \cdot Y_i)
    \right)
    \right\}
    \\
    &&
    =
    2 \cdot
          \EXP \left\{
    \sup_{f \in \F}
    \left(
    \frac{1}{n} \sum_{i=1}^n
\epsilon_i \cdot 
\varphi( f(X_i) \cdot Y_i)
    \right)
    \right\}.
     \end{eqnarray*}
   Next we use a contraction-style argument. Because of the independence
   of the random variables we can compute the expectation by first computing
   the expectation with respect to $\epsilon_1$ and then computing
   the expectation with respect to all other random variables.
   Consequently, the last term above is equal to
   \begin{eqnarray*}
     &&
         2 \cdot
         \EXP \Bigg\{
        \frac{1}{2} \cdot
    \sup_{f \in \F}
    \left(
    \frac{1}{n} \sum_{i=2}^n
\epsilon_i \cdot 
\varphi( f(X_i) \cdot Y_i)
    +
    \frac{1}{n} \cdot
\varphi( f(X_1) \cdot Y_1)
    \right)
    \\
    &&
    \quad \quad
    +
            \frac{1}{2} \cdot
    \sup_{g \in \F}
    \left(
    \frac{1}{n} \sum_{i=2}^n
\epsilon_i \cdot 
\varphi( g(X_i) \cdot Y_i)
    -
    \frac{1}{n} \cdot
    \varphi( g(X_1) \cdot Y_1)
    \right)
\Bigg\}
\\
&&
=
         \EXP \Bigg\{
    \sup_{f,g  \in \F}
    \Bigg(
    \frac{1}{n} \sum_{i=2}^n
\epsilon_i \cdot 
\varphi( f(X_i) \cdot Y_i)
    +         
    \frac{1}{n} \sum_{i=2}^n
\epsilon_i \cdot 
\varphi( g(X_i) \cdot Y_i)
    \\
    &&
    \quad \quad \quad \quad
    +
    \frac{1}{n} \cdot
\varphi( f(X_1) \cdot Y_1)
 -
    \frac{1}{n} \cdot
\varphi( g(X_1) \cdot Y_1)
    \Bigg)
    \Bigg\}.
   \end{eqnarray*}
   Because of
   \[
\varphi^\prime(z)=\frac{1}{1 + \exp(-z)} \cdot \exp(-z) \cdot (-1) \in [-1,0],
   \]
   $\varphi$ is Lipschitz continuous with Lipschitz constant $1$ which implies
   \begin{eqnarray*}
   \frac{1}{n} \cdot
\varphi( f(X_1) \cdot Y_1)
 -
    \frac{1}{n} \cdot
    \varphi( g(X_1) \cdot Y_1)
    &\leq& \frac{1}{n} \cdot | f(X_1) \cdot Y_1 -  g(X_1) \cdot Y_1|
    \\
    &\leq& \frac{1}{n} \cdot |f(X_1)-g(X_1)|.
   \end{eqnarray*}
Hence the last expectation above is upper bounded by
   \begin{eqnarray*}
    &&
         \EXP \Bigg\{
    \sup_{f,g  \in \F}
    \Bigg(
    \frac{1}{n} \sum_{i=2}^n
\epsilon_i \cdot 
\varphi( f(X_i) \cdot Y_i)
    +         
    \frac{1}{n} \sum_{i=2}^n
\epsilon_i \cdot 
\varphi( g(X_i) \cdot Y_i)
    +
    \frac{1}{n} \cdot
    |f(X_1)-g(X_1)|
    \Bigg)
    \Bigg\}
.
   \end{eqnarray*}
   For fixed $(X_1,Y_1)$, \dots, $(X_n,Y_n)$, $\epsilon_2$, \dots, $\epsilon_n$
   the term
   \[
    \frac{1}{n} \sum_{i=2}^n
\epsilon_i \cdot 
\varphi( f(X_i) \cdot Y_i)
    +         
    \frac{1}{n} \sum_{i=2}^n
\epsilon_i \cdot 
\varphi( g(X_i) \cdot Y_i)
    +
    \frac{1}{n} \cdot
    |f(X_1)-g(X_1)|
    \]
    is symmetric in $f$ and $g$. Therefore we can assume w.l.o.g. that
    $f(X_1) \geq g(X_1)$ holds which implies that we have
    \begin{eqnarray*}
      &&
    \sup_{f,g  \in \F}
    \Bigg(
    \frac{1}{n} \sum_{i=2}^n
\epsilon_i \cdot 
\varphi( f(X_i) \cdot Y_i)
    +         
    \frac{1}{n} \sum_{i=2}^n
\epsilon_i \cdot 
\varphi( g(X_i) \cdot Y_i)
    +
    \frac{1}{n} \cdot
    |f(X_1)-g(X_1)|
    \Bigg)
    \\
    &&
    =
\sup_{f,g  \in \F}
    \Bigg(
    \frac{1}{n} \sum_{i=2}^n
\epsilon_i \cdot 
\varphi( f(X_i) \cdot Y_i)
    +         
    \frac{1}{n} \sum_{i=2}^n
\epsilon_i \cdot 
\varphi( g(X_i) \cdot Y_i)
    +
    \frac{1}{n} \cdot
    (f(X_1)-g(X_1))
    \Bigg).
    \end{eqnarray*}
    In the same way we see that the term above is also equal to
\[
\sup_{f,g  \in \F}
    \Bigg(
    \frac{1}{n} \sum_{i=2}^n
\epsilon_i \cdot 
\varphi( f(X_i) \cdot Y_i)
    +         
    \frac{1}{n} \sum_{i=2}^n
\epsilon_i \cdot 
\varphi( g(X_i) \cdot Y_i)
    -
    \frac{1}{n} \cdot
    (f(X_1)-g(X_1))
    \Bigg),
    \]
    and we get
    \begin{eqnarray*}
      &&
         \EXP \Bigg\{
    \sup_{f,g  \in \F}
    \Bigg(
    \frac{1}{n} \sum_{i=2}^n
\epsilon_i \cdot 
\varphi( f(X_i) \cdot Y_i)
    +         
    \frac{1}{n} \sum_{i=2}^n
\epsilon_i \cdot 
\varphi( g(X_i) \cdot Y_i)
\\
&&
\hspace*{2cm}
    +
    \frac{1}{n} \cdot
    |f(X_1)-g(X_1)|
    \Bigg)
    \Bigg\}
    \\
    &&
    =
    \EXP \Bigg\{
    \frac{1}{2} \cdot
    \sup_{f,g  \in \F}
    \Bigg(
    \frac{1}{n} \sum_{i=2}^n
\epsilon_i \cdot 
\varphi( f(X_i) \cdot Y_i)
    +         
    \frac{1}{n} \sum_{i=2}^n
\epsilon_i \cdot 
\varphi( g(X_i) \cdot Y_i)
\\
&&
\hspace*{2cm}
    +
    \frac{1}{n} \cdot
    (f(X_1)-g(X_1))
    \Bigg)
    \\
    &&
    \quad \quad
+    \frac{1}{2} \cdot
    \sup_{f,g  \in \F}
    \Bigg(
    \frac{1}{n} \sum_{i=2}^n
\epsilon_i \cdot 
\varphi( f(X_i) \cdot Y_i)
    +         
    \frac{1}{n} \sum_{i=2}^n
\epsilon_i \cdot 
\varphi( g(X_i) \cdot Y_i)
\\
&&
\hspace*{2cm}
    -
    \frac{1}{n} \cdot
    (f(X_1)-g(X_1))
    \Bigg)    
    \Bigg\}
    \\
    &&
    =
  \EXP \Bigg\{
    \sup_{f,g  \in \F}
    \Bigg(
    \frac{1}{n} \sum_{i=2}^n
\epsilon_i \cdot 
\varphi( f(X_i) \cdot Y_i)
    +         
    \frac{1}{n} \sum_{i=2}^n
\epsilon_i \cdot 
\varphi( g(X_i) \cdot Y_i)
  \\
&&
\hspace*{2cm}
  +
    \frac{1}{n} \cdot \epsilon_1 \cdot
    (f(X_1)-g(X_1))
    \Bigg)
    \Bigg\}
    \\
    &&
    \leq
  \EXP \Bigg\{
    \sup_{f,g  \in \F}
    \Bigg(
    \frac{1}{n} \sum_{i=2}^n
\epsilon_i \cdot 
\varphi( f(X_i) \cdot Y_i)
    +
    \frac{1}{n} \cdot \epsilon_1 \cdot
    f(X_1)
    \Bigg)
    \Bigg\}
    \\
    &&
    \quad \quad
    +
  \sup_{f,g  \in \F}
    \Bigg(
    \frac{1}{n} \sum_{i=2}^n
\epsilon_i \cdot 
\varphi( g(X_i) \cdot Y_i)
    +
    \frac{1}{n} \cdot (-\epsilon_1) \cdot g(X_1)
    \Bigg)
    \Bigg\}
    \\
    &&
    =
    2 \cdot
  \EXP \Bigg\{
    \sup_{f  \in \F}
\Bigg(
    \frac{1}{n} \sum_{i=2}^n
\epsilon_i \cdot 
\varphi( f(X_i) \cdot Y_i)
    +
    \frac{1}{n} \cdot \epsilon_1 \cdot
    f(X_1)
    \Bigg)
      \Bigg\},
    \end{eqnarray*}
    where we have used that $-\epsilon_1$ has the same distribution as
    $\epsilon_1$.

    Arguing in the same way for $i=2, \dots, n$ we get
    \begin{eqnarray*}
      &&
    2 \cdot
          \EXP \left\{
    \sup_{f \in \F}
    \left(
    \frac{1}{n} \sum_{i=1}^n
\epsilon_i \cdot 
\varphi( f(X_i) \cdot Y_i)
    \right)
    \right\}
    \\
    &&
    \leq
    2 \cdot
  \EXP \Bigg\{
    \sup_{f  \in \F}
    \Bigg(
    \frac{1}{n} \sum_{i=2}^n
\epsilon_i \cdot 
\varphi( f(X_i) \cdot Y_i)
    +
    \frac{1}{n} \cdot \epsilon_1 \cdot
    f(X_1)
    \Bigg)
    \Bigg\}
    \\
    &&
    \leq
    2 \cdot
  \EXP \Bigg\{
    \sup_{f  \in \F}
    \Bigg(
    \frac{1}{n} \sum_{i=3}^n
\epsilon_i \cdot 
\varphi( f(X_i) \cdot Y_i)
    +
    \frac{1}{n} \cdot ( \epsilon_1 \cdot
    f(X_1) + \epsilon_2 \cdot
    f(X_2))
    \Bigg)
    \Bigg\}
    \\
    &&
    \leq \dots
    \\
    && \leq
    2 \cdot
    \EXP \left\{
    \sup_{f \in \F}
    \frac{1}{n} \cdot \sum_{i=1}^n \epsilon_i \cdot f(X_i)
    \right\},    
      \end{eqnarray*}
    which finishes the fourth step of the proof.
    
    In the {\it fifth step of the proof} we show
    \begin{eqnarray*}
      &&
   \EXP \left\{
    \sup_{f \in \F}
    \frac{1}{n} \sum_{i=1}^n \epsilon_i \cdot f(X_i)
    \right\} \cdot 1_{E_n}
    \leq
    \EXP \left\{
    \sup_{\btheta \in \bar{\bTheta}}
    \left|
    \frac{1}{n} \sum_{i=1}^n \epsilon_i \cdot (T_{\beta_n}(f_{\btheta}(X_i))
    \right|
    \right\}
.
    \end{eqnarray*}
        Let $\W$ be the set of all weight vectors $\bw=( (w_k)_{k=1, \dots, K_n},
    (\btheta_k)_{k=1, \dots, K_n})$
    which satisfy $\btheta=(\btheta_k)_{k=1, \dots, K_n} \in \bar{\bTheta}^{K_n}$
    and (\ref{se4sub1eq2}). 
    Because of $f=0$ is contained in $\F$ it implies
    \[
\sup_{f \in \F}
    \frac{1}{n} \sum_{i=1}^n \epsilon_i \cdot f(X_i)
    \geq 0
    \]
    from which we can conclude
    \begin{eqnarray*}
      &&
       \EXP \left\{
    \sup_{f \in \F}
    \frac{1}{n} \sum_{i=1}^n \epsilon_i \cdot f(X_i)
    \right\} \cdot 1_{E_n}
    \\
    &&
   \leq
       \EXP \Bigg\{
       \sup_{\bw \in \W}
    \frac{1}{n} \sum_{i=1}^n \epsilon_i \cdot 
    \sum_{j=1}^{K_n} w_j \cdot (T_{\beta_n} f_{\btheta_j}(X_i)))
    \Bigg\}
    \\
    &&
   =
       \EXP \Bigg\{
       \sup_{\bw \in \W}
    \sum_{j=1}^{K_n} w_j \cdot  \frac{1}{n} \sum_{i=1}^n \epsilon_i \cdot 
   (T_{\beta_n} f_{\btheta_j}(X_i)))
    \Bigg\}
    \\
   &&
   \leq
       \EXP \Bigg\{
       \sup_{\bw \in \W}
    \sum_{j=1}^{K_n} |w_j| \cdot  \left| \frac{1}{n} \sum_{i=1}^n \epsilon_i \cdot 
    (T_{\beta_n} f_{\btheta_j}(X_i)))
    \right|
    \Bigg\}
    \\
   &&
   \leq
       \EXP \Bigg\{
       \sup_{\bw \in \W}
       \sum_{j=1}^{K_n} |w_j| \cdot
       \sup_{\btheta \in \bar{\bTheta}^{K_n}, k \in \{1, \dots, K_n\}}
       \left| \frac{1}{n} \sum_{i=1}^n \epsilon_i \cdot 
    (T_{\beta_n} f_{\btheta_k}(X_i)))
    \right|
    \Bigg\}
    \\
   &&
    \leq
    1 \cdot
       \EXP \Bigg\{
       \sup_{\btheta \in \bar{\bTheta}^{K_n}, k \in \{1, \dots, K_n\}}
       \left| \frac{1}{n} \sum_{i=1}^n \epsilon_i \cdot 
    (T_{\beta_n} f_{\btheta_k}(X_i)))
    \right|
    \Bigg\}
    \\
    &&
    =
       \EXP \Bigg\{
       \sup_{\btheta \in \bar{\bTheta}^{K_n}}
       \left| \frac{1}{n} \sum_{i=1}^n \epsilon_i \cdot 
    (T_{\beta_n} f_{\btheta_1}(X_i)))
    \right|
    \Bigg\}
    \\
    &&
    =
        \EXP \left\{
    \sup_{\btheta \in \bar{\bTheta}}
    \left|
    \frac{1}{n} \sum_{i=1}^n \epsilon_i \cdot (T_{\beta_n}(f_{\btheta}(X_i))
    \right|
    \right\}
,
      \end{eqnarray*}
    where the last inequality followed from
    \[
    \{ T_{\beta_n} f_{\btheta_k} \, : \, \btheta \in \bar{\bTheta}^{K_n}, k \in \{1, \dots, K_n\}\}
    =
    \{ T_{\beta_n} f_{\btheta_1} \, : \,  \btheta \in \bar{\bTheta} \}. 
    \]

    In the {\it sixth step of the proof} we show
    \begin{eqnarray*}
      T_{3,n}
      & \leq &
      c_{14} \cdot \Bigg(
        \frac{C_n +1}{\sqrt{N_n}}
  +
  \frac{D_n^2}{t_n}
  +
  \sup_{\btheta \in \bTheta^*}
  \EXP \left\{ \varphi(Y \cdot f_\btheta(X)) \right\}
  \\
  &&
  \hspace*{3cm}
  -
  \min_{f : \R^{d \cdot l} \rightarrow \bar{\R}}
  \EXP \left\{ \varphi(Y \cdot f(X)) \right\}
  \Bigg).
      \end{eqnarray*}
    Application of standard techniques concerning the analysis of
    gradient descent in case of convex function (cf., Lemma \ref{le1new}) yields
\begin{eqnarray*}
  &&
  \frac{1}{n} \sum_{i=1}^n \varphi(Y_i \cdot f_n(X_i))\cdot 1_{E_n}
  \\
  &&
  =
  \min_{t=0, \dots, t_n}
  F_n((\bw^{(t)},\btheta^{(t)})) \cdot 1_{E_n}
  \\
  &&
  \leq
  F_n((\bw^*,\btheta^{(0)})) \cdot 1_{E_n}
  +
  \frac{1}{t_n} \cdot
  \sum_{t=1}^{t_n}
  | F_n((\bw^*,\btheta^{(t)}))-F_n((\bw^*,\btheta^{(0)}))|
\cdot 1_{E_n}
\\
&&
\hspace*{3cm}
  +
  \frac{\|\bw^*\|^2}{2} + \frac{D_n^2}{2 \cdot t_n}.
  \end{eqnarray*}
By the definition of $\bw^*$ we know
\[
\frac{\|\bw^*\|^2}{2} \leq \frac{1}{2 \cdot N_n}.
\]
The logistic loss is convex (since $\varphi^{\prime \prime}(z) \geq 0$
for all $z \in \R$) from which we can conclude
\begin{eqnarray*}
  &&
  \EXP\left\{ F_n((\bw^*,\btheta^{(0)})) \cdot 1_{E_n} \right\}
  \\
  &&
  =
  \EXP\left\{  \EXP\left\{ F_n((\bw^*,\btheta^{(0)}))
  \big| \btheta^{(0)} \right\} \cdot 1_{E_n} \right\}
  \\
  &&
  =
  \EXP \left\{
  \EXP \left\{
  \frac{1}{n} \sum_{i=1}^n
  \varphi( \frac{1}{N_n} \sum_{k=1}^{N_n} Y_i \cdot
  T_{\beta_n} f_{\btheta_{j_k}^{(0)}}(X_i))
    \big| \btheta^{(0)} \right\} \cdot 1_{E_n}
  \right\}
  \\
  &&
  \leq
   \EXP \left\{
  \frac{1}{N_n} \sum_{k=1}^{N_n}
    \EXP \left\{
  \frac{1}{n} \sum_{i=1}^n
  \varphi(  Y_i \cdot T_{\beta_n} f_{\btheta_{j_k}^{(0)}}(X_i))
  \big| \btheta^{(0)} \right\} \cdot 1_{E_n}
  \right\}
  \\
  &&
  \leq
  \sup_{\btheta \in \Theta^*}
  \EXP \left\{
\varphi( Y \cdot T_{\beta_n} f_{\btheta}(X)
  \right\}.
  \end{eqnarray*}

Finally we conclude from the fact that $\varphi$ is Lipschitz
continuous with Lipschitz constant $1$, the Cauchy-Schwarz
inequality and assumption (\ref{th2eq1}) that we have
\begin{eqnarray*}
  &&
  \frac{1}{t_n} \cdot
  \sum_{t=1}^{t_n}
  | F_n((\bw^*,\btheta^{(t)}))-F_n((\bw^*,\btheta^{(0)}))| \cdot 1_{E_n}
  \\
  &&
  =
  \frac{1}{t_n} \cdot
  \sum_{t=1}^{t_n}
  \left|
  \frac{1}{n} \sum_{i=1}^n
  \left(
  \varphi(  Y_i \cdot f_{(\bw^*,\btheta^{(t)})}(X_i))
  -
  \varphi(  Y_i \cdot f_{(\bw^*,\btheta^{(0)})}(X_i))
  \right)
  \right| \cdot 1_{E_n}
  \\
  &&
  \leq
  \max_{t=1, \dots, t_n} \max_{i=1, \dots, n}
  |f_{(\bw^*,\btheta^{(t)})}(X_i)-f_{(\bw^*,\btheta^{(0)})}(X_i)| \cdot 1_{E_n}
  \\
  &&
  \leq
  \max_{t=1, \dots, t_n} \max_{i=1, \dots, n}
  \sqrt{ \sum_{k=1}^{K_n} |w_k^*|^2}
  \cdot
  \sqrt{
\sum_{i=1}^{N_n} | f_{\btheta^{(t)}_{j_i}}(X_i)-f_{\btheta^{(0)}_{j_i}}(X_i)|^2
  }
\cdot 1_{E_n}
  \\
  &&
  \leq
  \frac{1}{\sqrt{N_n}} \cdot
  \sqrt{
\sum_{i=1}^{N_n} C_n^2 \cdot \| \btheta^{(t)}_{j_i} - \btheta^{(0)}_{j_i}\|^2
  }
    =
    \frac{1}{\sqrt{N_n}} \cdot C_n \cdot
     \sqrt{
\sum_{i=1}^{N_n}  \| \btheta^{(t)}_{j_i} - \btheta^{(0)}_{j_i}\|^2
     }
     \\
     &&
       \leq
       \frac{1}{\sqrt{N_n}} \cdot C_n \cdot \|\btheta^{(t)} - \btheta^{(0)}\|
    \leq c_6 \cdot
  \frac{C_n}{\sqrt{N_n}}.
  \end{eqnarray*}
Here (\ref{th2eq1}) is applicable because
the definition of the estimate implies
\begin{eqnarray*}
\| \btheta^{(t)}_{j_i} - \btheta^{(0)}_{j_i}\|
& \leq &
\sqrt{
  \sum_{s=1}^{N_n}
  \| \btheta^{(t)}_{j_s} - \btheta^{(0)}_{j_s}\|^2
}
\leq
\sqrt{
\| \btheta^{(t)} - \btheta^{(0)}\|^2
}
\leq c_6.
\end{eqnarray*}

  Gathering the above results completes the proof.
        \hfill $\Box$

\section{Proof of Theorem \ref{th1}}
\label{se5}

     In the sequel we show
        \begin{eqnarray}
          \label{pth1eq1}
          &&
          \EXP\left\{
\varphi( Y \cdot \hat{f}_n(X))
\right\}
-
\EXP\left\{
\varphi( Y \cdot f_{\varphi}^*(X))
\right\}
\nonumber \\
&&
\leq
c_{85} \cdot (\log n)^{6} \cdot
  \max_{(p,K) \in \P} n^{- \min \left\{ \frac{p}{(2p+K)}, \frac{1}{3} \right\}}.
          \end{eqnarray}
        This implies the assertion, because by Lemma \ref{le2new} a)
        we conclude from (\ref{pth1eq1})
        \begin{eqnarray*}
          &&
          \PROB \left\{
Y \neq sgn(f_n(X))| \right\}
-
\PROB \left\{
Y \neq \eta^*(X) \right\}
\\
&&
\leq
\EXP \left\{
\frac{1}{\sqrt{2}}
\cdot
\left(
\EXP\left\{
\varphi( Y \cdot f_n(X))
| \D_n
\right\}
-
\EXP\left\{
\varphi( Y \cdot f_{\varphi^*}(X))
\right\}
\right)^{1/2}
\right\}
\\
&&
\leq
\frac{1}{\sqrt{2}}
\cdot
\sqrt{
\EXP\left\{
\varphi( Y \cdot f_n(X))
\right\}
-
\EXP\left\{
\varphi( Y \cdot f_{\varphi^*}(X))
\right\}
}
\\
&&
\leq
  c_{86} \cdot (\log n)^{3} \cdot
  \max_{(p,K) \in \P} n^{- \min \left\{ \frac{p}{2 \cdot (2p+K)}, \frac{1}{6} \right\}}
          \end{eqnarray*}
And from Lemma \ref{le2new} b), (\ref{th1eq1}) and Lemma \ref{le2new} c)
we conclude from (\ref{pth1eq1})
\begin{eqnarray*}
  &&
        \PROB \left\{
Y \neq sgn(f_n(X))| \right\}
-
\PROB \left\{
Y \neq \eta^*(X) \right\}
\\
&&
\leq
2 \cdot
\left(
\EXP\left\{
\varphi( Y \cdot f_n(X))
\right\}
-
\EXP\left\{
\varphi( Y \cdot f_{\varphi^*}(X))
\right\}
\right)
+ 4 \cdot \frac{c_{87} \cdot \log n}{n^{1/3}}
\\
&&
\leq
  c_{88} \cdot (\log n)^{6} \cdot
  \max_{(p,K) \in \P} n^{- \min \left\{ \frac{p}{2p+K}, \frac{1}{3} \right\}}.
\end{eqnarray*}
Here we have used the fact that
\[
\max
\left\{
\frac{\PROB\{Y=1|X\}}{1-\PROB\{Y=1|X\}},
\frac{1-\PROB\{Y=1|X\}}{\PROB\{Y=1|X\}}
\right\}
> n^{1/3}
\]
is equivalent to
\[
|f_{\varphi^*}(X)|
=
\left|
\log \frac{\PROB\{Y=1|X\}}{1-\PROB\{Y=1|X\}}
\right|
> \frac{1}{3} \cdot \log n.
\]

So it suffices to prove (\ref{pth1eq1}), which we do in the sequel by
applying Theorem \ref{th2}.

In the {\it first step of the proof} we define $\bTheta$, $\bTheta^0$
and
$\bTheta^*$.

Let $\bTheta=\bTheta_0$ be the set of all pairs 
$(\bW,\bV)$
of weight matrices of the transformer networks $f_{(\bW_k,\bV_k)}$
introduced in Subsection \ref{se2sub1}.
In the supplement we will introduce Transformer networks with good
approximation properties, and we use here these Transformer
networks for the definition of  $\bTheta^*$:
Let  $\bTheta^*$ be the
set of all weight matrices $(\bW,\bV)$ where $\bW$ is from the
weight matrices introduced in Theorem \ref{th3} in supremum
norm not further away than
\[
\epsilon=\frac{1}{c_{89} \cdot n^{c_{90}}}, 
\]
and where $\bV$ is from the weight matrix introduced in Lemma
\ref{le12new}
in supremum norm not further away than
\[
\bar{\epsilon}=\frac{1}{c_{91} \cdot n^{c_{92}}}.
\]

In the {\it second step of the proof} we show that
\[
C_n = c_{93} \cdot n^{c_{94}}
\]
satisfies (\ref{th2eq1}).
We will show this in the Supplement in Lemma \ref{le13},
which is applicable provided we choose
$c_6 \leq 1/(2 \cdot c_{62})$.

In the {\it third step of the proof} we show that
$\epsilon_n$ defined by  (\ref{th2eq2}) satisfies
\[
\epsilon_n \geq \frac{1}{e^{(\log n)^2 \cdot \sqrt{n}}}.
\]

The event
$\{ \theta_1^{(0)} \in \bTheta^* \}$
occurs if the pruning step selects the right subset of size $L_n \leq
c_{95} \cdot \sqrt{n}$
out of all subsets of size $L_n$ of the possible set of parameters,
which has size less equal than $c_{96} \cdot n$, and if the uniform distributions
(on intervals of length $2 \cdot c_{4} \cdot n^{c_{5}}$)
choose each of
these $L_n$ parameters correctly from an interval of size
$1/(c_{97} \cdot n^{c_{98}})$. This implies for large $n$
\[
\PROB \{ \theta_1^{(0)} \in \bTheta^* \} \geq
\frac{1}{(c_{95} \cdot n)^{c_{96} \cdot \sqrt{n}}} \cdot
\left( \frac{1}{2 \cdot c_{4} \cdot n^{c_{5}} \cdot c_{97} \cdot n^{c_{98}}} \right)^{c_{95} \cdot \sqrt{n}}
\geq
\frac{1}{e^{(\log n)^2 \cdot \sqrt{n}}}.
\]

In the {\it fourth step of the proof} we
show that
\[
N_n=n^{c_{99}}, \quad I_n= \lceil (\log n)^2 \cdot e^{(\log n)^2 \cdot \sqrt{n}} \rceil
\]
satisfies (\ref{th2eq3}) for $n$ large.

This follows from
\begin{eqnarray*}
N_n \cdot (1-\epsilon_n)^{I_n}
& \leq &
n^{c_{99}} \cdot 
\left(
1
-
\frac{1}{e^{(\log n)^2 \cdot \sqrt{n}}}
\right)^{ \lceil (\log n)^2 \cdot e^{(\log n)^2 \cdot \sqrt{n}} \rceil }
\\
&
\leq
&
\frac{1}{n}
\end{eqnarray*}
for $n$ large.

In the {\it fifth step of the proof} we show that
\[
D_n= \sqrt{K_n} \cdot \beta_n
\]
satisfies (\ref{th2eq4}).

We will show this in Lemma \ref{le3new} in the Supplement.

In the {\it sixth step of the proof} we show
\begin{eqnarray*}
&&
 \EXP \left\{
  \left|
  \sup_{\btheta \in \bar{\bTheta}}
  \frac{1}{n} \sum_{i=1}^n \epsilon_i \cdot T_{\beta_n}(f_\btheta(X_i))
  \right|
  \right\}
  \leq  c_{100} \cdot (\log n)^{3} \cdot
  \left(
  \max_{(p,K) \in \P} n^{- \frac{p}{2p+K}}
  + n^{-\frac{1}{3}}
\right)
  .
 \end{eqnarray*}

To see this, we use standard techniques from empirical process
theory which are summarized in Lemma \ref{le4new}
in the Supplement. From this we conclude
\begin{eqnarray*}
&&
 \EXP \left\{
  \left|
  \sup_{\btheta \in \bar{\bTheta}}
  \frac{1}{n} \sum_{i=1}^n \epsilon_i \cdot T_{\beta_n}(f_\btheta(X_i))
  \right|
  \right\}
\\
&&
\leq
 c_{101} \cdot \frac{\sqrt{\max\{h \cdot I, d_{ff}, J_n\}  }\cdot (\log n)^2}
{\sqrt{n}}
\\
&&
\leq
 c_{102} \cdot \frac{ ( \sqrt{\log n}  
  \cdot \max_{(p,K) \in \P} n^{K/(2 \cdot (2  \cdot p+K))} +n^{\frac{1}{6}})   \cdot (\log n)^2}
 {\sqrt{n}}
\\
&&
\leq
  c_{103} \cdot (\log n)^{3} \cdot
\left(  \max_{(p,K) \in \P} n^{- \frac{p}{2p+K}} + n^{- \frac{1}{3}} \right).
\end{eqnarray*}

In the {\it  seventh step of the proof} we show
\begin{eqnarray*}
&&
\sup_{\btheta \in \bTheta^*}
  \EXP \left\{ \varphi(Y \cdot T_{\beta_n} f_\btheta(X)) \right\}
  -
  \min_{f : \Rd \rightarrow \bar{\R}}
  \EXP \left\{ \varphi(Y \cdot f(X)) \right\}
\leq c_{104} \cdot  \frac{\log n}{n^{1/3}} + c_{105} \cdot   \max_{j,i} h^{-p_j^{(i)}/K_j^{(i)}}
\end{eqnarray*}
for $n$ sufficiently large.
We have
\begin{eqnarray*}
&&
\sup_{\btheta \in \bTheta^*}
  \EXP \left\{ \varphi(Y \cdot T_{\beta_n} f_\btheta(X)) \right\}
  -
  \min_{f : \Rd \rightarrow \bar{\R}}
  \EXP \left\{ \varphi(Y \cdot f(X)) \right\}
\\
&&
=
\sup_{\btheta \in \bTheta^*}
  \EXP \left\{ \varphi(Y \cdot T_{\beta_n} f_\btheta(X))  - \varphi( Y \cdot f_{\varphi^*}(X))\right\}
\\
&&
=
\sup_{\btheta \in \bTheta^*}
  \EXP \Bigg\{ 
1_{\{Y=1\}} \cdot
\left(
\varphi(T_{\beta_n} f_\btheta(X))  - \varphi( f_{\varphi^*}(X))
\right)
\\
&&
\hspace*{3cm}
+
1_{\{Y=-1\}} \cdot
\left(
\varphi(- T_{\beta_n} f_\btheta(X))  - \varphi( - f_{\varphi^*}(X))
\right)
\Bigg\}
\\
&&
=
\sup_{\btheta \in \bTheta^*}
  \EXP \Bigg\{ 
m(X) \cdot
\left(
\varphi(T_{\beta_n} f_\btheta(X))  - \varphi( f_{\varphi^*}(X))
\right)
\\
&&
\hspace*{3cm}
+
(1-m(X)) \cdot
\left(
\varphi(- T_{\beta_n} f_\btheta(X))  - \varphi( - f_{\varphi^*}(X))
\right)
\Bigg\}
\\
&&
\leq
\sup_{\btheta \in \bTheta^*}
\sup_{x \in \R^{d \cdot l}}
\Bigg( 
|m(x)| \cdot
\left|
\varphi(T_{\beta_n} f_\btheta(x))  - \varphi( f_{\varphi^*}(x))
\right|
\\
&&
\hspace*{3cm}
+
|1-m(x)| \cdot
\left|
\varphi(- T_{\beta_n} f_\btheta(x))  - \varphi( - f_{\varphi^*}(x))
\right|
\Bigg).
\end{eqnarray*} 
Application of the approximation results for Transformer
networks derived in the Supplement, i.e., application of
Lemma \ref{le12new} and Theorem \ref{th3}
(which is applicable because of the pruning step in the gradient
descent introduced in Section \ref{se2}, which implies 
in particular that some components of $z_{k,r}$ do not change during the computation)
yields the assertion. Here we have used that Lemma \ref{le12new}
yields an approximating function which is bounded in absolute value
by
\[
\log J_n + 11 \cdot (3 \cdot J_n + 9) \cdot J_n \cdot
\frac{1}{
c_{91} \cdot n^{c_{92}}
}
\leq
\beta_n
\]
for $n$ sufficiently large
and consequently the truncation operator above can be ignored.

In the {\it eighth step of the proof} we complete the proof by
showing (\ref{pth1eq1}).

Thanks to the results of the steps 1 through 5 we know that the assumptions
of Theorem \ref{th2} are satisfied. Application of Theorem \ref{th2}
yields
\begin{eqnarray*}
&&
        \EXP\left\{
\varphi( Y \cdot f_n(X))
\right\}
-
\EXP\left\{
\varphi( Y \cdot f_{\varphi^*}(X))
\right\}
\\
&&
\leq
c_{106} \cdot  \Bigg(
  \frac{\log n}{n}
  +
 \EXP \left\{
  \left|
  \sup_{\btheta \in \bar{\bTheta}}
  \frac{1}{n} \sum_{i=1}^n \epsilon_i \cdot T_{\beta_n}(f_\btheta(X_i))
  \right|
  \right\}
  +
  \frac{C_n +1}{\sqrt{N_n}}
  +
  \frac{D_n^2}{t_n}
  \\
  &&
  \hspace*{3cm}
  +
  \sup_{\btheta \in \bTheta^*}
  \EXP \left\{ \varphi(Y \cdot T_{\beta_n} f_\btheta(X)) \right\}
  -
  \min_{f : \R^{d \cdot l} \rightarrow \bar{\R}}
  \EXP \left\{ \varphi(Y \cdot f(X)) \right\}
  \Bigg).
\end{eqnarray*}
Plugging in the results of steps 6 and 7 and the values
of $C_n$, $N_n$ and $D_n$ derived in steps 2, 4 and 5 yields
the assertion.
                        \hfill $\Box$

\newpage
\begin{appendix}
\section{SUPPLEMENTARY MATERIAL}\label{appn} 

\subsection{A result for gradient descent}

\begin{lemma}
  \label{le1new}
  Let $d_1, d_2 \in \N$, let $D_n \geq 0$,
  let $A \subset \R^{d_1}$ and $B \subseteq \R^{d_2}$
  be closed and convex, and let $F:\R^{d_1} \times \R^{d_2} \rightarrow \R_+$
  be a function such that
  \[
  u \mapsto F(u,v) \quad \mbox{is differentiable and convex for all } v \in
  \R^{d_2}
\]
and
  \begin{equation}
    \label{le1neweq1}
\| (\nabla_{u} F)(u,v) \| \leq D_n
  \end{equation}
  for all $(u,v) \in A \times B$. 
  Choose $(u_0,v_0) \in A \times B$, let $v_1, \dots, v_{t_n} \in B$ and set
  \[
  u_{t+1} = Proj_A \left(
u_t - \lambda \cdot \left( \nabla_u F \right)(u_t,v_t)
  \right) \quad (t=0, \dots, t_n-1),
  \]
where
  \[
\lambda = \frac{1}{t_n}. 
\]
Let    $u^* \in A$. Then it holds:
  \begin{eqnarray*}
    &&
\min_{t=0, \dots, t_n}    F(u_t,v_t)
    \leq
    F(u^*,v_0)
    +
\frac{1}{t_n} \sum_{t=1}^{t_n} | F(u^*,v_t) - F(u^*,v_0)|
    +
    \frac{\|u^*-u_0\|^2}{2}
    +
    \frac{D_n^2}{2 \cdot t_n}.
    \end{eqnarray*}
\end{lemma}

\noindent
    {\bf Proof.}
    The result follows from the proof of Lemma 1 in Kohler and Krzy\.zak (2023).
    For the sake of completeness we give nevertheless a complete proof here.

        In the {\it first step of the proof} we show
    \begin{equation}
      \label{ple1neweq1}
      \frac{1}{t_n}
      \sum_{t=0}^{t_n-1} F(u_t,v_t)
      \leq
          \frac{1}{t_n}
      \sum_{t=0}^{t_n-1} F(u^*,v_t)
      +
      \frac{\|u^*-u_0\|^2}{2}
      +
      \frac{1}{2 \cdot t_n^2}
      \sum_{t=0}^{t_n-1}
      \| (\nabla_{u}F)(u_t,v_t) \|^2.
      \end{equation}
    By convexity of $u \mapsto F(u,v_t)$ and because of
    $u^* \in A$ we have
    \begin{eqnarray*}
      &&
      F(u_t,v_t) - F(u^*,v_t)
      \\
      &&
      \leq
      \,
      < (\nabla_u F)(u_t,v_t), u_t-u^* >
      \\
      &&
      =
      \frac{1}{2 \cdot \lambda}
      \cdot 2
      \cdot
      < \lambda \cdot (\nabla_u F)(u_t,v_t), u_t-u^* >
      \\
      &&
      =
      \frac{1}{2 \cdot \lambda}
      \cdot
      \left(
      - \| u_t - u^* - \lambda \cdot (\nabla_u F)(u_t,v_t)\|^2
      +
      \|u_t-u^*\|^2
      +
      \|  \lambda \cdot (\nabla_u F)(u_t,v_t) \|^2
      \right)\\
      &&
      \leq
       \frac{1}{2 \cdot \lambda}
      \cdot
      \left(
      - \| Proj_A (u_t  - \lambda \cdot (\nabla_u F)(u_t,v_t)) - u^*\|^2
      +
      \|u_t-u^*\|^2
      +
      \lambda^2 \cdot \|  (\nabla_u F)(u_t,v_t) \|^2
      \right)
      \\
      &&
      =
       \frac{1}{2 \cdot \lambda}
      \cdot
      \left(
      \|u_t-u^*\|^2
      - \| u_{t+1} - u^*\|^2
      +
      \lambda^2 \cdot \|  (\nabla_u F)(u_t,v_t) \|^2
      \right).
      \end{eqnarray*}
    This implies
    \begin{eqnarray*}
&&      \frac{1}{t_n}
      \sum_{t=0}^{t_n-1} F(u_t,v_t)
      -
          \frac{1}{t_n}
      \sum_{t=0}^{t_n-1} F(u^*,v_t)
      \\
      &&
      =
        \frac{1}{t_n}
      \sum_{t=0}^{t_n-1} \left( F(u_t,v_t)
      -
      F(u^*,v_t) \right)
      \\
      &&
      \leq
      \frac{1}{t_n}
      \sum_{t=0}^{t_n-1}
       \frac{1}{2 \cdot \lambda}
      \cdot
      \left(
      \|u_t-u^*\|^2
      - \| u_{t+1} - u^*\|^2
      \right)
      +
      \frac{1}{t_n}
      \sum_{t=0}^{t_n-1}
      \frac{\lambda}{2} \cdot \|  (\nabla_u F)(u_t,v_t) \|^2
      \\
      &&
      =
        \frac{1}{2}
\cdot
        \sum_{t=0}^{t_n-1}
      \left(
      \|u_t-u^*\|^2
      - \| u_{t+1} - u^*\|^2
      \right)
      +
      \frac{1}{2 \cdot t_n^2}
      \sum_{t=0}^{t_n-1}
      \|  (\nabla_u F)(u_t,v_t) \|^2
      \\
      &&
      \leq
       \frac{ \|u_0-u^*\|^2}{2}
       +
      \frac{1}{2 \cdot t_n^2}
      \sum_{t=0}^{t_n-1}
      \|  (\nabla_u F)(u_t,v_t) \|^2. 
      \end{eqnarray*}

    In the {\it second step of the proof} we show the assertion.

    Using the result of step 1 we get
    \begin{eqnarray*}
      &&
\min_{t=0, \dots, t_n}    F(u_t,v_t)
\\
&&
\leq
      \frac{1}{t_n}
      \sum_{t=0}^{t_n-1} F(u_t,v_t)
      \\
      &&
      \leq
          \frac{1}{t_n}
      \sum_{t=0}^{t_n-1} F(u^*,v_t)
      +
      \frac{\|u^*-u_0\|^2}{2}
      +
      \frac{1}{2 \cdot t_n^2}
      \sum_{t=0}^{t_n-1}
      \| (\nabla_{u}F)(u_t,v_t) \|^2\\
      &&
      \leq
      F(u^*,v_0)
      +
         \frac{1}{t_n}
         \sum_{t=0}^{t_n-1}
         |F(u^*,v_t)-F(u^*,v_0)|
      +
      \frac{\|u^*-u_0\|^2}{2}
      \\
      &&
      \quad
      +
      \frac{1}{2 \cdot t_n^2}
      \sum_{t=0}^{t_n-1}
      \| (\nabla_{u}F)(u_t,v_t) \|^2.
      \end{eqnarray*}
    By  (\ref{le1neweq1}) we get
    \[
\frac{1}{2 \cdot t_n^2}
      \sum_{t=0}^{t_n-1}
      \| (\nabla_{u}F)(u_t,v_t) \|^2
      \leq
      \frac{1}{2 \cdot t_n^2}
      \sum_{t=0}^{t_n-1}
      D_n^2
      =
      \frac{D_n^2}{2 \cdot t_n}.
    \]
    Summarizing the above results, the proof is complete.

    \hfill $\Box$

\subsection{An auxiliary result}

\begin{lemma}
  \label{le2new}
  Let $\varphi$ be the logistic loss. Let $(X,Y), (X_1, Y_1), \dots, (X_n, Y_n)$ and $\eta^*$,
  $\mathcal{D}_n, f_n$ and $\eta_n$ as in Sections 1 and 2, and set
  \[
  f_{\varphi^*}=
  \arg \min_{f: \R^{d \cdot l} \rightarrow \bar{\R}}
  \EXP
  \left\{
\varphi( Y \cdot f(X) )
  \right\}
  .
  \]\\ 
{\bf a)} Then
\begin{eqnarray*}
&&
\PROB \left\{
Y \neq \eta_n(X)| \D_n \right\}
-
\PROB \left\{
Y \neq \eta^*(X) \right\}
\\
&&
\leq
\frac{1}{\sqrt{2}}
\cdot
\left(
\EXP\left\{
\varphi( Y \cdot f_n(X))
| \D_n
\right\}
-
\EXP\left\{
\varphi( Y \cdot f_{\varphi^*}(X))
\right\}
\right)^{1/2}
\end{eqnarray*}
holds.\\
{\bf b)}
Then
\begin{eqnarray*}
&&
\PROB \left\{
Y \neq \eta_n(X) | \D_n \right\}
-
\PROB \left\{
Y \neq \eta^*(X) \right\}
\\
&&
\leq
2
\cdot
\left(
\EXP\left\{
\varphi( Y \cdot f_n(X))
| \D_n
\right\}
-
\EXP\left\{
\varphi( Y \cdot f_{\varphi^*}(X))
\right\}
\right)
+
4 \cdot
\EXP\left\{
\varphi( Y \cdot f_{\varphi^*}(X))
\right\}.
\end{eqnarray*}
holds.
\\
{\bf c)} Assume that 
\[
\PROB \left\{ |f_{\varphi^*}(X)| > \tilde{F}_n\right\} \geq 1 - e^{-\tilde{F}_n}
\]
for a given sequence $\{\tilde{F}_n\}_{n \in \N}$ with $\tilde{F}_n \to \infty$.
Then
\[
\EXP\left\{
\varphi( Y \cdot f_{\varphi^*}(X))
\right\}
\leq c_{15} \cdot \tilde{F}_n \cdot e^{-\tilde{F}_n}
\]
holds.
  \end{lemma}

\noindent
    {\bf Proof.}
    {\bf a)} This result follows from Theorem 2.1 in Zhang (2004), where we choose $s=2$ and $c=2^{-1/2}$. \\
    {\bf b)} This result follows from Lemma 1 b)  in Kohler and Langer (2020b). \\
    {\bf c)}  This result follows from Lemma 3 in Kim, Ohn and Kim (2019).
    \hfill $\Box$

    \subsection{A bound on the gradient}
In the proof of Theorem \ref{th1} we will apply Theorem \ref{th2}.
For this we need the following bound on the gradient
(with respect to the outer weights) of $F_n$.

\begin{lemma}
\label{le3new}
Let $F_n$ be defined by (\ref{se2eq13}). Then we have
\[
\| \nabla_{(w_k)_{k=1, \dots, K_n}} F_n(\bw) \|
\leq \sqrt{K_n} \cdot \beta_n.
\]
\end{lemma}
    
\noindent
{\bf Proof.}
For $k \in \{1, \dots, K_n\}$ we have
\begin{eqnarray*}
&&
\frac{\partial F_n(\bw)}{\partial w_k}
=
\frac{1}{n} \sum_{i=1}^n
\varphi^\prime (Y_i \cdot f_{\bw}(X_i))
\cdot Y_i \cdot T_{\beta_n}(f_{\bW_k,\bV_k}(X_i)).
\end{eqnarray*}

Because of $|\varphi^\prime(z)| \leq 1$ we can conclude
\[
\left|
\frac{\partial F_n(\bw)}{\partial w_k}
\right|
\leq \beta_n
\]
and
\begin{eqnarray*}
&&
\| \nabla_{(w_k)_{k=1, \dots, K_n}} F_n(\bw) \|^2
= \sum_{k=1}^{K_n}
\left|
\frac{\partial F_n(\bw)}{\partial w_k}
\right|^2
\leq K_n \cdot \beta_n^2.
\end{eqnarray*}
\quad \hfill $\Box$

    \subsection{Generalization error}

    \begin{lemma}
  \label{le4new} 
  Let
$d_{model}=h \cdot I$ and let
  $\F$ be the set of all functions
  \[
(x_1, \dots, x_l) \mapsto z_{1,N,1}^{(d+l+2)},
  \]
  where $z_{1,N,1}$ is defined in Section \ref{se2} depending on
  \begin{equation}
    \label{le4neweq1}
  (W_{query,1,r,s}, W_{key,1,r,s}, W_{value,1,r,s})_{r \in \{1, \dots, N\}, s \in \{1, \dots, h \}}
  \end{equation}
  and on
    \begin{equation}
    \label{le4neweq2}
    (W_{1,r,1},b_{1,r,1},W_{1,r,2},b_{1,r,2})_{r \in \{1, \dots, N\}}
    \end{equation}
  and where the total number of nonzero components in each row
  in all the matrices
  in (\ref{le4neweq1}) is bounded by $\tau \in \N$ and where all
  matrices $W_{1,r,1}$ and $W_{1,r,2}$ in (\ref{le4neweq2}) have the
  property that in each row in $W_{1,r,1}$ and in each column in
  $W_{1,r,2}$ there are at most $\tau$ nonzero entries.
  Let $\G$ be the set of all (shallow) feedforward neural networks
  $g:\R \rightarrow \R$
  with one hidden layer and $J_n$ hidden neurons 
  and ReLU activation function.
  Assume 
  \[
  \max\{N, d_{key}, d_v,l\} \leq c_{16}
  \quad \mbox{and} \quad
\max\{J_n,h,I,d_{ff}\}
\leq
c_{17} \cdot n^{c_{17}}.
\]
  Let $A \geq 1$, let $X_1, \dots, X_n$
  be independent and identically distributed $[-A,A]^{d \cdot l}$-valued
  random vectors and let $\epsilon_1, \dots, \epsilon_n$ be independent
  Rademacher random variables, which are independent from $X_1, \dots, X_n$. Then we have for some constant $c_{18}>0$ which depends on $\tau$
  \begin{eqnarray*}
    &&
    \EXP \left\{
    \left|
    \sup_{f \in \G \circ \F}
    \frac{1}{n} \sum_{i=1}^{n} \epsilon_i \cdot T_{\beta_n}(f(X_i)) 
    \right|
    \right\}
    \leq
   c_{18} \cdot \frac{ \sqrt{ \max\{h \cdot I, d_{ff} , J_n\} } \cdot (\log n)^2}
{\sqrt{n}}. 
    \end{eqnarray*}
  \end{lemma}

In order to prove Lemma \ref{le4new} we need the following bound on the covering
number.

\begin{lemma}
  \label{le5new}
Define $\F$ and $\G$ as in Lemma \ref{le4new}.
  Let $\beta \geq 0$ and let $T_\beta \G \circ \F$
  be the set of all functions $g \circ f$ truncated on height $\beta$ and $-\beta$ where $g \in \G$ and $f \in \F$.
  Then we have for any $0<\epsilon<\beta/2$
  \begin{eqnarray*}
&&
  \sup_{z_1^n \in (\R^{d \cdot l})^n} \log \Mu_1(\epsilon, T_\beta \G
   \circ \F, z_1^n) 
\\
&&
\leq
  c_{19} \cdot \tau \cdot \max\{h \cdot I, d_{ff}, J_n\}  \cdot N^3 \cdot \log( \max\{J_n, N,h,d_{ff},I, d_v,l,2\}) \log \left(
\frac{\beta}{\epsilon}
  \right). 
  \end{eqnarray*}
  \end{lemma}

In order to prove Lemma \ref{le5new} we will first show the following
bound on the VC-dimension of subsets of $\F$, where the nonzero
components appear only at fixed positions.

\begin{lemma}
  \label{le6new}
   Let $\F$ be the set of all functions
  \[
(x_1, \dots, x_l) \mapsto z_{1,N,1}^{(d+l+2)},
  \]
  where $z_{1,N,1}$ is defined in Section \ref{se2} depending on
   (\ref{le4neweq1}) and (\ref{le4neweq2}) 
  and where in all matrics in  (\ref{le4neweq1}) there are
  in each row at most $\tau$ fixed components
  where the entries are allowed to be nonzero,
  and where all
  matrices $W_{1,r,1}$ and $W_{1,r,2}$ in (\ref{le4neweq2}) have the
  property that in each row in $W_{1,r,1}$ and in each column in
  $W_{1,r,2}$ there are at most $\tau$ fixed components (depending on $r$)
  where the entries are allowed to be nonzero.
Let $\G$ be defined as in Lemma \ref{le4new}.
  Then we have
  \[
  V_{(\G \circ \F)^+} \leq
  c_{20} \cdot \tau \cdot \max\{ h \cdot I, d_{ff}, J_n\} 
  \cdot N^3 \cdot \log( \max\{J_n,N,h,d_{ff}, d_v,l,2\}). 
\]
  \end{lemma}

The proof of Lemma \ref{le6new} is a modification of the proof of
Lemma 9 in
Gurevych, Kohler and Sarin (2022),
which in turn is based on the proof of Theorem 6 in Bartlett et al. (2019).
In the proof of Lemma \ref{le6new} we will need the following two auxiliary
results.

\begin{lemma}
\label{le7new}
Suppose $W\leq m$ and let $f_1,...,f_m$ be polynomials of degree at most $D$ in $W$ variables. Define
\[
K:=|\{\left(sgn(f_1(\ba)),\dots,sgn(f_m(\ba))\right) : \ba\in\R^{W}\}|.
\]
Then we have 
\[
K\leq2\cdot\left(\frac{2\cdot e\cdot m\cdot D}{W}\right)^{W}.
\]
\end{lemma}
\noindent
    {\bf Proof.} See Theorem 8.3 in Anthony and Bartlett (1999).
		\hfill $\Box$

      \begin{lemma}
\label{le8new}
	Suppose that $2^m\leq2^L\cdot(m\cdot R/w)^w$ for some $R\geq16$ and $m\geq w\geq L\geq0$. Then,
	\[
	m\leq L+w\cdot\log_2(2\cdot R\cdot\log_2(R)).
	\]
\end{lemma}
\noindent
{\bf Proof.}
See Lemma 16 in Bartlett et al. (2019).
\hfill $\Box$          

\noindent
    {\bf Proof of Lemma \ref{le6new}}.
    Let $\HH$ be the set of all functions $h$ defined by
    \[
    h:\R^{d \cdot l} \times \R \rightarrow \R,
    \quad
    h(x,y)=g(x)-y
    \]
    for some $g \in \G \circ \F$.
                    Let $(x_1, y_1)$, \dots
                    $(x_m,y_m) \in \R^{d \cdot l} \times \R$ be such that
                    \begin{equation}
                      \label{ple6neweq1}
                      |
                      \{
(sgn(h(x_1,y_1)), \dots, sgn(h(x_m,y_m))) \, : \, h \in \HH
                      \}
                      |
                      =2^m.
                      \end{equation}
It suffices to show
                         \begin{equation}
                           \label{ple6neweq2}
                           m \leq
                           c_{20} \cdot \tau \cdot
                           \max\{ h \cdot I, d_{ff} , J_n\}
                           \cdot N^3 \cdot \log(
                           \max\{J_n, N,h,d_{ff},d_{key},d_v,l,2\}).
                      \end{equation}
                         To show this we partition $\G \circ \F$ in subsets
                         such that for each subset all
                         \[
g \circ f (x_i) \quad (i=1, \dots, m)
                         \]
                         are polynomials of some fixed degree and use Lemma \ref{le7new}
                         in order to derive an upper bound on the left-hand side of (\ref{ple6neweq1}).
                         This upper bound will depend polynomially on $m$ which will enable
                         us to conclude (\ref{ple6neweq2}) by an application of Lemma \ref{le8new}.
                         
                         Let
                         \begin{eqnarray*}
                         \bar{\theta} &=& \Bigg(
  (W_{query,r,s}, W_{key,r,s}, W_{value,r,s})_{r \in \{1, \dots, N\}, s \in \{1, \dots, h\}},
                        \\
                         && \quad
                     (W_{r,1},b_{r,1},W_{r,2},b_{r,2})_{r \in \{1, \dots, N\}}, 
                         (v_r^{(1)})_{r \in \{1, \dots, J_n\}},
                         (v_{r,s}^{(0)})_{r \in \{1, \dots, J_n\}, s \in \{0, 1\}}
                         \Bigg)
                         \end{eqnarray*}
                         be the parameters which determine a function in
                         $\G \circ \F$.
                         By assumption, each function in $\G \circ \F$
                         can be also described by such a parameter vector.
                         Here
                         only
                         \[
                         \bar{L}_n
                         = N \cdot h \cdot (2 \cdot d_{key} + d_v) \cdot \tau 
                         + N \cdot (d_{ff} \cdot (\tau+1)
                         + d_{ff} \cdot \tau
                         + h \cdot I)  +
                         3 \cdot J_n
                         \]
                         components of the matrices and vectors
                         occuring in the parameter vector
                         are allowed to be nonzero
                         and the positions
                         where these nonzero parameters can occur are fixed.
                         Denote the vector in $\R^{\bar{L}_n}$ which contains all
                         values of these possible nonzero parameters
                         by $\theta$. Then we can write
                         \[
                         \G \circ \F = \{ g(\cdot, \theta):
                         \R^{d \cdot l} \rightarrow \R \, : \,
                         \theta \in \R^{\bar{L}_n}
                         \}.
                         \]
                         In the sequel we construct a partition $\P_{N+1}$ of $\R^{\bar{L}_n}$
                         such that for all $S \in \P_{N+1}$ we have
                         that
                         \[
                         g(x_1, \theta), \dots, g(x_m,\theta)
                         \]
                          (considered as functions of $\theta$) are 
                         polynomials of degree at most $8^{N}+2$
                         for $\theta \in S$.

                         In order to construct this partition we construct
                         first recursively partitions $\P_{0}$, \dots, $\P_N$ of $\R^{\bar{L}_n}$ such that for each $r \in \{1, \dots, N\}$ and all $S \in \P_r$
                         all components in
                         \[
z_r=z_r(x) \quad (x \in \{x_1, \dots, x_m\})
                         \]
                         (considered as a function of $\theta$) are polynomials of degree at most $8^r$ in $\theta$ for
                         $\theta \in S$.

                         Since all components of $z_0$ are constant as
                         functions of $\theta$ this holds
                         for $r=0$ if we set $\P_{0}=\{ \R^{\bar{L}_n}\}$.

                         Let $r \in \{1, \dots, N\}$ and assume that
                         for 
 all $S \in \P_{r-1}$
                         all components in
                         \[
z_{r-1}(x) \quad (x \in \{x_1, \dots, x_m\})
                         \]
                         (considered as a function of $\theta$) are polynomials of degree at most $8^{r-1}$ in $\theta$ for
                         $\theta \in S$.
                         Then all components in
                         \[
q_{r-1,s,i}(x), k_{r-1,s,i}(x) \quad \mbox{and} \quad v_{r-1,s,i}(x) \quad (x \in \{x_1, \dots, x_m\})
                         \]
                         are on each set $S \in \P_{r-1}$ polynomials of degree
                         at most $8^{r-1} +1$.
                         Consequently, for each $S \in \P_{r-1}$ each value
                         \[
<q_{r-1,s,i}(x), k_{r-1,s,j}(x)>  \quad (x \in \{x_1, \dots, x_m\}) 
                         \]
                         is (considered as a function of $\theta$)
                         a polynomial of degree at most $2 \cdot 8^{r-1}+2$ 
                          for
                          $\theta \in S$. Application of
                          Lemma \ref{le7new} yields that
                          \[
                          <q_{r-1,s,i}(x), k_{r-1,s,j_1}(x)>
                          -
                          <q_{r-1,s,i}(x), k_{r-1,s,j_2}(x)>
                          \]
                          $
                          (s \in \{1, \dots, h\}, i, j_1, j_2 \in \{1, \dots, l\}, x \in \{x_1, \dots, x_m\})$
                          has at most
                          \[
                          \Delta
                          =
                          2 \cdot
                          \left(
                          \frac{
2 \cdot e \cdot h \cdot l^3 \cdot m \cdot (2 \cdot 8^{r-1}+2)
                          }{\bar{L}_n}
                          \right)^{\bar{L}_n}
                          \]
                          different sign patterns. If we partition each set
                          in $\P_{r-1}$ according to these sign patterns
                          in $\Delta$ subsets, then on each set in the new partition
                          all components in
                          \[
                          v_{r-1,s,\hat{j}_{r-1,s,i}}(x) \cdot  <q_{r-1,s,i}(x), k_{r-1,s,\hat{j}_{r-1,s,i}}(x)>
\quad (x \in \{x_1, \dots, x_m\}) 
                          \]
                          are polynomials of degree at most
                          $3 \cdot 8^{r-1} + 3$
                          (since on each such set
                          \[
                          <q_{r-1,s,i}(x), k_{r-1,s,\hat{j}_{r-1,s,i}}(x)>
                          \]
                          is equal to one of the
                          $<q_{r-1,s,i}(x), k_{r-1,s,j}(x)>$).
                          On each set within this partition every
                          component of the $\R^{d_{ff}}$-valued vectors
                          \[
                          W_{r,1} \cdot y_{r,s}(x) + b_{r,1}
                          \quad (s=1, \dots, h,
                          x \in \{x_1, \dots, x_m\})
                          \]
                          is (considered as a function of $\theta$)
                          a polynomial of degree at most $3 \cdot 8^{r-1} +4$.

                          By another application of Lemma \ref{le7new}
                          we can refine each set in this partition
                          into
                          \[
2 \cdot
                          \left(
                          \frac{
2 \cdot e \cdot h \cdot d_{ff} \cdot m \cdot (3 \cdot 8^{r-1} + 4)
                          }{\bar{L}_n}
                          \right)^{\bar{L}_n}
                          \]
                          sets such that all components in
                          \begin{equation}
                            \label{ple6neweq3}
W_{r,1} \cdot y_{r,s}(x) + b_{r,1} \quad (x \in \{x_1, \dots, x_m\}) 
                          \end{equation}
                          have the same sign patterns within the refined partition.
                          We call this refined partition $\P_r$.
                          Since on each set of $\P_r$ the sign of all components
                          in (\ref{ple6neweq3}) does not change we can conclude
                          that
 all components in
                          \begin{equation}
                            \label{ple6neweq3b}
\sigma( W_{r,1} \cdot y_{r,s}(x) + b_{r,1}) \quad (x \in \{x_1, \dots, x_m\})
                          \end{equation}                         
                          are either equal to zero or they are equal
                          to a polynomial of degree at most
                          $ 3 \cdot 8^{r-1} + 4$.
                          Consequently we have that on each set
                          in $\P_r$ all components of
                          \[
z_r(x) \quad (x \in \{x_1, \dots, x_m\})
                          \]
                          are equal to a polynomial of degree at most
$ 3 \cdot 8^{r-1} + 5 \leq 8^r$.

                          The partition $\P_N$ satisfies
                          \[
|\P_N| = \prod_{r=1}^N
                          \frac{|\P_r|}{|\P_{r-1}|}
                          \leq
                          \prod_{r=1}^N
                          2 \cdot
                          \left(
                          \frac{
2 \cdot e \cdot h \cdot l^3 \cdot m \cdot 8^r
                          }{\bar{L}_n}
                          \right)^{\bar{L}_n}
                          \cdot
2 \cdot
                          \left(
                          \frac{
2 \cdot e \cdot h \cdot d_{ff} \cdot m \cdot 8^r
                          }{\bar{L}_n}
                          \right)^{\bar{L}_n}.
                          \]

                          Next we construct a partition $\P_{N+1}$ of $\R^{\bar{L}_n}$
 such that for  all
 $S \in \P_{N+1}$
 \[
 g(z_{N,1}^{(d+l+2)}(x))
 =
 \sum_{j=1}^{J_n}
 v_j^{(1)} \cdot
 \sigma \left(
v_{j,1}^{(0)} \cdot z_{N,1}^{(d+l+2)}(x) + v_{j,0}^{(0)}
\right)
\quad (x \in \{x_1, \dots, x_m\})
 \]
(considered as a function of $\theta$) is a polynomial of degree
at most $8^N+2$ for $\vartheta \in S$.

For all $S \in \P_{N}$ all components
in
\[
\left(
v_{j,1}^{(0)} \cdot z_{N,1}^{(d+l+2)}(x) + v_{j,0}^{(0)}
\right)_{j=1, \dots, J_n}
\quad (x \in \{x_1, \dots, x_m\})
\]
(considered as a function of $\theta$) are polynomials of degree
at most $8^N+1$.
Application of
                          Lemma \ref{le7new} implies
                          \[
v_{j,1}^{(0)} \cdot z_{N,1}^{(d+l+2)}(x) + v_{j,0}^{(0)}
                          \quad
                          (j \in \{1, \dots, J_n\}, x \in \{x_1, \dots, x_m\})
                          \]
                          has at most
                          \[
                          \Delta
                          =
                          2 \cdot
                          \left(
                          \frac{
2 \cdot e \cdot J_n \cdot m \cdot (8^{N}+1)
                          }{\bar{L}_n}
                          \right)^{\bar{L}_n}
                          \]
                          different sign patterns. If we partition in each set
                          in $\P_{N}$ according to these sign patterns
                          in $\Delta$ subsets, then on each set in the new partition $\P_{N+1}$
                          all components in
                          \[
                          \sigma \left(
v_{j,1}^{(0)} \cdot z_{N,1}^{(d+l+2)}(x) + v_{j,0}^{(0)} \right)
                          \quad
                          (j \in \{1, \dots, J_n\}, x \in \{x_1, \dots, x_m\})
                          \]
                          are polynomials of degree at most
                          $8^{N} + 1$ (since after the application of
                          $\sigma(z)=\max\{z,0\}$ the component is either
                          equal to zero on the set or equal to the argument
                          of $\sigma$).
                          Consequently on each set in $\P_{N+1}$
                          \[
                          g(x, \theta)
                          =
 \sum_{j=1}^{J_n}
 v_j^{(1)}(x) \cdot
 \sigma \left(
v_{j,1}^{(0)} \cdot (z_{N,1}^{(d+l+2)} + v_{j,0}^{(0)}
\right)
\quad (x \in \{x_1, \dots, x_m\})
                          \]
                          (considered as a function of $\theta$) is a polynomial
                          of degree $8^N+2$.

                          The partition $\P_{N+1}$ satisfies
                          \begin{eqnarray*}
                          |\P_{N+1}| &=& 
                          \frac{|\P_{N+1}|}{|\P_{N}|}
                          \cdot
                          |\P_N|
                          \\
                          &
                          \leq &
                          2 \cdot
                          \left(
                          \frac{
2 \cdot e \cdot J_n  \cdot m \cdot (8^{N}+1)
                          }{\bar{L}_n}
                          \right)^{\bar{L}_n}
                          \cdot
                          \Bigg(
                          \prod_{r=1}^N
                          2 \cdot
                          \left(
                          \frac{
2 \cdot e \cdot h \cdot l^3 \cdot m \cdot 8^r
                          }{\bar{L}_n}
                          \right)^{\bar{L}_n}
                          \\
                          &&
                          \hspace*{2cm}
                          \cdot
2 \cdot
                          \left(
                          \frac{
2 \cdot e \cdot h \cdot d_{ff} \cdot m \cdot 8^r
                          }{\bar{L}_n}
                          \right)^{\bar{L}_n}
                          \Bigg)
                          \end{eqnarray*}
                          and has the property
                          that for
                          all $S \in \P_{N+1}$ and
                          for all $(x,y) \in \{(x_1,y_1), \dots, (x_m,y_m)\}$
\[
g(x)=g(x,\theta)  \quad \mbox{and} \quad h(x,y)=g(x,\theta) -y 
\]
                         (considered as a function of $\theta$) are polynomials of degree at most $8^N+2$ in $\theta$ for
                         $\theta \in S$.

Using
\begin{eqnarray*}
&&
|
                      \{
(sgn(h(x_1,y_1)), \dots, sgn(h(x_m,y_m))) \, : \, h \in \HH
                      \}
                      |
                      \\
                      &&
                      \leq 
                      \sum_{S \in \P_{N+1}}
                      |
                      \{
(sgn(g(x_1, \theta)-y_1), \dots, sgn(g(x_m, \theta)-y_m)) \, : \, \theta \in S
                      \}
                      |
                      \end{eqnarray*}
                      we can apply one more time Lemma \ref{le7new} to conclude
                      \begin{eqnarray*}
                        && 2^m
                        \\
                        &&=
  |
                      \{
(sgn(h(x_1,y_1)), \dots, sgn(h(x_m,y_m))) \, : \, h \in \HH
                      \}
                      |
                      \\
                      &&
                      \leq
                      |\P_{N+1}| \cdot
                      2 \cdot \left(
\frac{2 \cdot e \cdot m \cdot (8^{N}+2)}{\bar{L}_n}
\right)^{\bar{L}_n}
\\
&&
\leq                    
                      2 \cdot \left(
\frac{2 \cdot e \cdot m \cdot (8^{N}+2)}{\bar{L}_n}
\right)^{\bar{L}_n}
\cdot
\Bigg(
\prod_{r=1}^N
                          2 \cdot
                          \left(
                          \frac{
2 \cdot e \cdot h \cdot l^3 \cdot m \cdot 8^r
                          }{\bar{L}_n}
                          \right)^{\bar{L}_n}
                          \\
                          &&
                          \hspace*{1cm}
                          \cdot
2 \cdot
                          \left(
                          \frac{
2 \cdot e \cdot h \cdot d_{ff} \cdot m \cdot 8^r
                          }{\bar{L}_n}
                          \right)^{\bar{L}_n}
                          \Bigg)
                          \cdot
                          2
                          \cdot
                          \left(
\frac{2 \cdot e \cdot J_n \cdot m \cdot (8^N+1)}{\bar{L}_n}
                          \right)^{\bar{L}_n}
                          \\
                          &&
                          \leq 
                          2^{2 \cdot N+2}
                          \cdot
\\
&&
                          \left(
                          \frac{
                            m \cdot 2 e \cdot (2 N +2)  \cdot  \max\{J_n,h\} \cdot (\max\{l, d_{ff}\})^3
                            \cdot (8^{N}+2)
                          }{(2 N +2) \cdot \bar{L}_n}
                          \right)^{(2 N +2) \cdot \bar{L}_n}.
                      \end{eqnarray*}
                      Assume $m \geq  (2 N +2) \cdot
                      (N \cdot h \cdot (2 \cdot d_{key} + d_v) \cdot \tau 
                      + N \cdot (d_{ff} \cdot (\tau+1)
                      + d_{ff} \cdot \tau
                         + h \cdot I ) +
                         3 \cdot J_n)
                         $.
Application of Lemma \ref{le8new} with $L= 2 \cdot N +2$,
                      $R=
2 e \cdot (2 N +2)  \cdot  \max\{ J_n,h\}  \cdot (\max\{l, d_{ff}\})^3
\cdot (8^{N}+2)                   $ and
                      $w=(2 N +2) \cdot \bar{L}_n$ yields
                        \begin{eqnarray*}
                        m &\leq& 
(2 \cdot N +2)
                        + (2 N +2) \cdot \bar{L}_n
                                 \cdot\log_2(2\cdot R\cdot\log_2(R))
                          \\
&\leq&
 c_{21} \cdot \tau \cdot \max\{ h \cdot I, d_{ff}, J_n\}  \cdot N^3 \cdot \log( \max\{J_n,N,h,d_{ff},l,2\}),
                        \end{eqnarray*}
                        which implies (\ref{ple6neweq2}).
         	    \hfill $\Box$

                    \noindent
                        {\bf Proof of Lemma \ref{le5new}.}
                        The functions in the function set $T_\beta \G \circ \F$ depend on
                        at most
                        \[
\lceil c_{22} \cdot (N \cdot h^2+ J_n) \cdot I \cdot \max\{d_k,d_{ff},d_v\} \rceil
                        \]
                        many parameters, and of these parameters at most
                        \[
                        \bar{L}_n= N \cdot h \cdot (2 \cdot d_{key} + d_v) \cdot \tau 
                         + N \cdot (d_{ff} \cdot (\tau+1)
                         + h \cdot I \cdot (\tau+1)) +
                         3 \cdot J_n
                        \]
                        are allowed
                        to be nonzero. We have
                        \begin{eqnarray*}
                          &&
\left( {\lceil c_{22} \cdot (N \cdot h^2+ J_n) \cdot I \cdot \max\{d_{key},d_{ff},d_v\} \rceil
  \atop \bar{L}_n} \right)
\\
&& \leq
\left( \lceil c_{22} \cdot (N \cdot h^2+ J_n) \cdot I \cdot
\max\{d_{key},d_{ff},d_v\} \rceil
\right)^{\bar{L}_n}
\end{eqnarray*}
many possibilities to choose these positions. If we fix these
positions, we get one function space $\G \circ \F$ for which we can bound its
VC dimension by Lemma \ref{le6new}.
                    Using Lemma \ref{le6new},
$V_{(T_{\beta} \G \circ \F)^+}
\leq
V_{(\G \circ \F)^+},$
and Theorem 9.4
in Gy\"orfi et al. (2002) we get
\begin{eqnarray*}
\mathcal{M}_1 \left(\epsilon,   T_{\beta} \G \circ \F,
   \bx_1^n\right)
&
   \leq
   &
3 \cdot \left(
\frac{4 e \cdot \beta}{\epsilon}
\cdot
\log
\frac{6 e \cdot \beta}{\epsilon}
\right)^{V_{(T_{\beta} \G \circ \F)^+}}
\\
&
\leq &
3 \cdot \left(
\frac{6 e \cdot \beta}{\epsilon}
\right)^{
2 \cdot  c_{20} \cdot \tau \cdot \max\{h \cdot I, d_{ff},J_n\} 
  \cdot N^3 \cdot \log( \max\{J_n,N,h,d_{ff},d_v,l,2\})
}.
\end{eqnarray*}
From this we conclude
\begin{eqnarray*}
  &&
  \sup_{z_1^n \in (\R^{d \cdot l})^n)} \log \Mu_1(\epsilon, T_\beta \G
     \circ \F, z_1^n)
  \\
  &&
  \leq
  \bar{L}_n \cdot
  \log
  \left( \lceil c_{22} \cdot ( N \cdot h^2+ J_n) \cdot I \cdot
  \max\{d_{key},d_{ff},d_v\} \rceil
 \right)
\\
&& \quad
 + 2 \cdot
 c_{20} \cdot \tau \cdot \max\{h \cdot I, d_{ff},J_n\} 
 \cdot N^3 \cdot \log( \max\{J_n,N,h,d_{ff},d_v,l,2\})
\cdot \log \left(
\frac{\beta}{\epsilon}
\right)
\\
&&
\leq
c_{23} \cdot \tau \cdot \max\{h \cdot  I, d_{ff} ,J_n\}
\cdot N^3 \cdot \log( \max\{J_n,N,h,I,d_{ff},d_{key},d_v,2\})
\cdot \log \left(
\frac{\beta}{\epsilon}
\right).
  \end{eqnarray*}
\hfill $\Box$

\noindent
    {\bf Proof of Lemma \ref{le4new}.}  
For $\delta_n>0$ we have
\begin{eqnarray*}
  &&
  \EXP \left\{
    \sup_{f \in \G \circ \F}
    \left|
    \frac{1}{n} \sum_{i=1}^n \epsilon_i \cdot (T_{\beta_n}(f(X_i))
    \right|
    \right\}
  \\
    &&
    =
    \int_0^\infty
    \PROB \left\{
    \sup_{f \in \G \circ \F}
    \left|
    \frac{1}{n} \sum_{i=1}^n \epsilon_i \cdot (T_{\beta_n}(f(X_i))
    \right|
    > t
    \right\}
    \, dt
    \\
    &&
    \leq
    \delta_n + \int_{\delta_n}^\infty
    \PROB \left\{    
    \sup_{f \in \G \circ \F}
    \left|
    \frac{1}{n} \sum_{i=1}^n \epsilon_i \cdot (T_{\beta_n}(f(X_i))
    \right|
    > t
    \right\}
    \, dt
.
\end{eqnarray*}
    Using a standard covering argument from empirical process theory we see that
    for any $\beta_n \geq t \geq \delta_n$ we have
    \begin{eqnarray*}
      &&
    \PROB \left\{    
    \sup_{f \in \G \circ \F}
    \left|
    \frac{1}{n} \sum_{i=1}^n \epsilon_i \cdot (T_{\beta_n}(f(X_i))
    \right|
    > t
    \right\}
    \\
    &&
    \leq
    \sup_{x_1^n \in (\R^{d \cdot l})^n }
    \Mu_1 \left(
    \frac{\delta_n}{2},
    \left\{
T_{\beta_n} f : f \in \G \circ \F
    \right\}, x_1^n
    \right)
    \\
    &&
    \hspace*{3cm}
    \cdot
    \sup_{f \in \G \circ \F}
    \PROB \left\{
    \left|
    \frac{1}{n} \sum_{i=1}^n \epsilon_i \cdot (T_{\beta_n}(f(X_i))
    \right|>\frac{t}{2}
    \right\}.
    \end{eqnarray*}
By Lemma \ref{le5new} we know
    \[
  \sup_{x_1^n \in (\R^{d \cdot l})^n}
        \Mu_1 \left(
    \frac{\delta_n}{2},
    \left\{
T_{\beta_n} f : f \in \G \circ \F
    \right\}, x_1^n
    \right)
    \leq c_{24} \cdot
    \left(
    \frac{c_{25}  \cdot \beta_n}{\delta_n}
    \right)^{c_{26}  \cdot \max\{h \cdot I \cdot d_{ff},J_n\}  \cdot \log (n)}.
    \]    
    By the inequality of Hoeffding (cf., e.g., Lemma A.3 in Gy\"orfi et al. (2002))
    and
    \[
    |T_{\beta_n}(f(x))| \leq  \beta_n
    \quad (x \in \R^d)
    \]
we have for any $f \in \G \circ \F$
\[
\PROB \left\{
     \left|
    \frac{1}{n} \sum_{i=1}^n \epsilon_i \cdot 
    T_{\beta_n}(f(X_i))
    \right|
    > t
    \right\}
    \leq 2 \cdot \exp \left(
- \frac{2 \cdot n \cdot t^2}{4  \cdot \beta_n^2}
    \right).
\]
Hence we get
\begin{eqnarray*}
  &&
  \EXP \left\{
    \sup_{f \in \G \circ \F}
    \left|
    \frac{1}{n} \sum_{i=1}^n \epsilon_i \cdot (T_{\beta_n}(f(X_i))
    \right|
    \right\}
  \\
    &&
  \leq
  \delta_n+ \int_{\delta_n}^{\beta_n}
c_{24} \cdot
    \left(
    \frac{c_{25}  \cdot \beta_n}{\delta_n}
    \right)^{
      c_{26}  \cdot \max\{h  \cdot I \cdot d_{ff},J_n\}   \cdot \log (n)
      }
    \cdot
    2 \cdot \exp \left(
- \frac{2 \cdot n \cdot t^2}{4 \cdot \beta_n^2}
\right) \, dt
\\
&&
\leq
  \delta_n+ \int_{\delta_n}^{\beta_n}
c_{24} \cdot
    \left(
    \frac{c_{25}  \cdot \beta_n}{\delta_n}
    \right)^{
      c_{26}  \cdot \max\{h \cdot I \cdot d_{ff} ,J_n\}   \cdot \log (n)
      }
    \cdot
    2 \cdot \exp \left(
- \frac{ n \cdot \delta_n \cdot t}{2  \cdot \beta_n^2}
\right) \, dt
\\
&&
\leq
\delta_n+
c_{24} \cdot
    \left(
    \frac{c_{25}  \cdot \beta_n}{\delta_n}
    \right)^{c_{26} \cdot \max\{h  \cdot I \cdot d_{ff} ,J_n\}  
    \cdot \log (n)}
    \frac{4  \cdot \beta_n^2}{ n \cdot \delta_n}
    \cdot
    \exp \left(
- \frac{ n \cdot \delta_n^2}{2  \cdot \beta_n^2}
\right). 
  \end{eqnarray*}
With
\[
\delta_n=
\sqrt{ \max\{h  \cdot I \cdot d_{ff},J_n\} } \cdot \log n
\cdot
\sqrt{
\frac{2 \cdot  \beta_n^2 }{ n}
}
\]
we get the assertion.
    \hfill $\Box$

\subsection{Approximation error}

\begin{lemma}
\label{le9new} 
Let $\tau \in \{l+1,l+2, \dots, l+d+1\}$.
Let $l,h,I \in \N$  and set $d_{model}=h \cdot I$.
Let $d_{key} \geq 3$.
Set $d_v=d_{model}/h=I$.
Let
$s_0 \in \{1, \dots, h\}$,
$s_1,s_2 \in \{1, \dots, d_{model}\}$,
$j \in \{1, \dots, l\}$, $k \in \{1, \dots, l\} \setminus \{j\}$,
$s_3 \in \{(s_0-1) \cdot d_v+1, \dots, s_0 \cdot d_v \}$,
$\beta \in \R$, $\delta \geq 0$ and
\begin{equation}
  \label{le9neweq1}
B \geq 168 \cdot d_{key} \cdot \tau^2 \cdot l \cdot (|\beta|+1)
  \cdot \|z_0\|_\infty^2 \cdot \max\{\delta^2,1\},
\quad
0 \leq \epsilon \leq
\min \left\{ 1, \frac{1}{36 \cdot \tau \cdot \|z_0\|_\infty^2}
\right\}.
\end{equation}
Then there exist
\[
\bW_{query,0,s_0},
\quad
\bW_{key,0,s_0}
\quad \mbox{and} \quad
\bW_{value,0,s_0}
\]
such that in each row of the above matrices
at most $\tau$ of its entries are not equal to zero,
such that in the last two rows of $\bW_{query,0,s_0}$ and
$\bW_{key,0,s_0}$ all entries in any column greater than $d+1+l$
are zero,
such that all entries are bounded in absolute value by $2 \cdot B$,
and such that we have for all 
$z_{0,r}, \tilde{z}_{0,r} \in \R^{d_{model}}$ satisfying
\begin{equation}
  \label{le9neweq4}
z_{0,r}^{(s)}
=
\tilde{z}_{0,r}^{(s)}
=
\begin{cases}
  x_r^{(s)} & \mbox{if } s\in \{1, \dots, d\}, \\
  1 & \mbox{if } s=d+1, \\
  \delta_{s-d-1,r} & \mbox{if } s \in \{d+2, \dots, d+1+l\}\\
  \end{cases}
\end{equation}
$(r \in \{1, \dots, l\})$ and
\[
\|\tilde{z}_{0,r} - z_{0,r}\|_\infty \leq \delta,
\]
$(r \in \{1, \dots, l\})$
and all
\[
\tilde{\bW}_{query,0,s},
\quad
\tilde{\bW}_{key,0,s}
\quad \mbox{and} \quad
\tilde{\bW}_{value,0,s}
\]
$(s \in \{1, \dots, h\})$
which satisfy
\[
\| \tilde{\bW}_{query,0,s_0}-\bW_{query,0,s_0}\|_{\infty} \leq \epsilon,
\]
\[
\| \tilde{\bW}_{key,0,s_0}-\bW_{key,0,s_0}\|_{\infty} \leq \epsilon,
\]
\[
\| \tilde{\bW}_{value,0,s_0}-\bW_{value,0,s_0}\|_{\infty} \leq \epsilon,
\]
and where in the last two rows
of $\tilde{\bW}_{query,0,s_0}$ and $\tilde{\bW}_{key,0,s_0}$
all entries in any column greater
than $d+l+1$ are zero and where
in  $\tilde{\bW}_{query,0,s_0}-\bW_{query,0,s_0}$,
$\tilde{\bW}_{key,0,s_0}-\bW_{key,0,s_0}$ and
$\tilde{\bW}_{value,0,s_0}-\bW_{value,0,s_0}$
in each row at most $\tau$ entries are nonzero,
that the following holds:

If we set for $s \in \{1, \dots, h\}$, $i \in \{1, \dots, l\}$
\[
q_{0,s,i} = \tilde{W}_{query,0,s} \cdot \tilde{z}_{0,i},
\quad
k_{0,s,i} = \tilde{W}_{key,0,s} \cdot \tilde{z}_{0,i}
\quad \mbox{and} \quad
v_{0,s,i} = \tilde{W}_{value,0,s} \cdot \tilde{z}_{0,i},
\]
\[
\hat{j}_{s,i} = \arg \max_{j \in \{1, \dots, l\}}
  <q_{0,s,i}, k_{0,s,j}>,
\]
\[
\bar{y}_{0,s,i} =v_{0,s,\hat{j}_{s,i}} \cdot  <q_{0,s,i}, k_{0,s,\hat{j}_{s,i}}>,
\]
\[
\bar{y}_{0,i}
= (\bar{y}_{0,1,i}, \dots, \bar{y}_{0,h,i})
\]
and
\[
\tilde{y}_{0,i}=\tilde{z}_{0,i}+\bar{y}_{0,i}
\]
then we have:
\begin{equation}
  \label{le9neweq2a}
\hat{j}_{s_0,1}=j, \quad \hat{j}_{s_0,r}=k \quad \mbox{if } r>1,
\end{equation}
\begin{eqnarray}
&&
\label{le9neweq2}
| \tilde{y}_{0,1}^{(s_3)}
-
(
z_{0,1}^{(s_3)}
+
z_{0,1}^{(s_1)}
\cdot
(z_{0,j}^{(s_2)} + \beta)
)
|
\nonumber \\
&&
\leq 136 \cdot d_{key} \cdot \tau^3 \cdot l \cdot (|\beta|+1)
\cdot \|z_0\|_\infty^3 \cdot B
\cdot \max\{ \delta^3,1 \} \cdot \epsilon
\nonumber \\
&&
\quad
+
25 \cdot \tau \cdot  (|\beta|+1) \cdot \|z_0\|_\infty \cdot \max\{ \delta,1 \}
\cdot \delta
\end{eqnarray}
and
\begin{eqnarray}
\label{le9neweq3}
| \tilde{y}_{0,r}^{(s)}
|
&\leq& 136 \cdot d_{key} \cdot \tau^3 \cdot l \cdot (|\beta|+1)
\cdot \|z_0\|_\infty^3 \cdot B
\cdot \max\{ \delta^3,1 \} \cdot \epsilon
\nonumber \\
&&
\quad
+
25 \cdot \tau \cdot  (|\beta|+1) \cdot \|z_0\|_\infty \cdot \max\{ \delta,1 \}
\cdot \delta
\end{eqnarray}
whenever $r > 1$ or $s \in \{ (s_0-1) \cdot d_v+1, \dots, s_0 \cdot d_v \}
\setminus \{ s_3 \}$.
\end{lemma}

\noindent
{\bf Proof of Lemma \ref{le9new}.} W.l.o.g. we assume $d_{key}=3$.

In the {\it first step of the proof} we define
$\bW_{query,0,s_0}$,
$\bW_{key,0,s_0}$
and
$\bW_{value,0,s_0}$
and present some of their properties.

Set
  \[
    W_{query,0,s_0} = \left(
    \begin{array}{cccccccccccccccc}
      0 & \dots & 0 & 0 & 0 & 0 & 0 &\dots & 0 &0 & \dots & 0 & 1 & 0 & \dots & 0 \\
      0 & \dots & 0 & -B & 0 & 0 & 0 & \dots & 0  & 0 & \dots & 0 & 0 & 0 & \dots & 0  \\
      0 & \dots & 0 & 0 & 0 & 1 & 1 & \dots & 1 & 0 & \dots & 0 & 0 & 0  & \dots & 0 \\
      \end{array}
    \right)
    \]
    where all columns are zero except columns number $d+1$, $d+3$, $d+4$, \dots,
    $d+1+l$ and $s_1$,
    \begin{eqnarray*}
      &&
      \hspace*{-0.8cm}
      W_{key,0,s_0} = \\
      &&       \hspace*{-0.8cm}
\left(
    \begin{array}{cccccccccccccccccccccc}
      0 & \dots & 0 & \beta & 0 & \dots & 0 & 0 & 0 & \dots & 0 & 0 & 0 & \dots & 0 & 0& \dots & 0 & 1 & 0 & \dots & 0 \\
      0 & \dots & 0 & 0 & 1 & \dots & 1 & 1 & 1& \dots & 1 &0 & 1 & \dots & 1 & 0 & \dots & 0 & 0 & 0 & \dots & 0 \\
      0 & \dots & 0 & 0 & 0 & \dots & 0 & 2 \cdot B & 0  & \dots & 0 & 0 & 0 & \dots & 0 & 0 & \dots & 0 & 0 & 0 & \dots & 0 
      \end{array}
    \right)
    \end{eqnarray*}
    where in the first row only the entries in columns $d+1$ and $s_2$
    are nonzero, where in the second row only the entries in columns
    $d+2$, $d+3$, \dots, $d+1+j-1$, $d+1+j+1$, $d+1+j+2$, \dots, $d+1+l$
    are nonzero, and where in the third row only the entry
    in column $d+1+k$ is nonzero, and
    \[
   W_{value,0,s_0} = \left(
    \begin{array}{ccccccc}
      0 & \dots & 0 & 0 & 0 & \dots & 0 \\
      \vdots & \dots & \vdots & \vdots & \vdots & \dots & \vdots\\
      0 & \dots & 0 & 0 & 0 & \dots & 0 \\
      0 & \dots & 0 & 1 & 0 & \dots & 0 \\
            0 & \dots & 0 & 0 & 0 & \dots & 0 \\
      \vdots & \dots & \vdots & \vdots & \vdots & \dots & \vdots\\
      0 & \dots & 0 & 0 & 0 & \dots & 0 \\
    \end{array}
    \right)
    \]
    where all rows and all columns are zero except column number $d+1+j$
    and row number $s_3-(s_0-1) \cdot d_v$.
  
Then we have
\[
W_{query,0,s_0} \cdot z_{0,r_1}=
\left(
\begin{matrix}
  z_{0,r_1}^{(s_1)} \\
    - B \\
    \sum_{i=2}^l \delta_{r_1,i}
\end{matrix}
\right)
, \quad
W_{key,0,s_0} \cdot z_{0,r_2}= \left(
\begin{matrix}
  \beta + z_{0,r_2}^{(s_3)} \\
  \sum_{i \in \{1, \dots, l\} \setminus \{j\}} \delta_{r_2,i} \\
  \delta_{r_2,k} \cdot 2 \cdot B
  \end{matrix}
\right)
\]
and
\[
W_{value,0,s_0} \cdot z_{0,r_2}= \delta_{r_2,j} \cdot \be_{s_3 -
  (s_0-1) \cdot d_v} 
\]
where $\be_r$ denotes the $r$-th unit vector in
$\R^{d_v}$. Hence
\begin{eqnarray*}
  &&
  < W_{query,0,s_0} z_{0,r_1}, W_{key,0,s_0} z_{0,r_2}>
  \\
  &&
  =
  z_{0,r_1}^{(s_1)} \cdot (\beta + z_{0,r_2}^{(s_3)})
  -
  B \cdot  \sum_{i \in \{1, \dots, l\} \setminus \{j\}} \delta_{r_2,i}
  +
  \sum_{i=2}^l \delta_{r_1,i} \cdot  \delta_{r_2,k} \cdot 2 \cdot B,
  \end{eqnarray*}
which implies
\begin{equation}
    \label{ple9neweq1}
<  W_{query,0,s_0} z_{0,1}, W_{key,0,s_0} z_{0,j}>
\quad = \quad
 z_{0,1}^{(s_1)} \cdot (\beta + z_{0,j}^{(s_2)}), 
\end{equation}
for $r_2 \neq j$
\[
<  W_{query,0,s_0} z_{0,1}, W_{key,0,s_0} z_{0, r_2 }>
\quad = \quad
 z_{0,1}^{(s_1)} \cdot (\beta + z_{0,r_2}^{(s_2)})
  -
  B, 
\]
for $r_1>1$
\[
<  W_{query,0,s_0} z_{0,r_1}, W_{key,0,s_0} z_{0,k}>
\quad = \quad
 z_{0,r_1}^{(s_1)} \cdot (\beta + z_{0,k}^{(s_2)})
  +
  B, 
\]
and for  $r_1>1$, $r_2 \in \{1, \dots, l\} \setminus \{k\}$
\[
<  W_{query,0,s_0} z_{0,r_1}, W_{key,0,s_0} z_{0,r_2}>
\quad = \quad
 z_{0,r_1}^{(s_1)} \cdot (\beta + z_{0,r_2}^{(s_2)})
  -
  B \cdot (1- \delta_{r_2,j}). 
\]
Because of
\[
B> 4 \cdot \max_{r_1,r_2} | z_{0,r_1}^{(s_1)} \cdot (\beta + z_{0,r_2}^{(s_2)})|
\]
we conclude
\begin{eqnarray}
  \label{ple9neweq2}
  &&
  < W_{query,0,s_0} z_{0,1}, W_{key,0,s_0} z_{0,j}>
\nonumber  \\
  &&
  >
  \frac{B}{2}
  +
  \max_{r_2 \neq j}
   < W_{query,0,s_0} z_{0,1}, W_{key,0,s_0} z_{0,r_2}>
  \end{eqnarray}
and for $r_1 > 1$
\begin{eqnarray}
    \label{ple9neweq3}
  &&
  < W_{query,0,s_0} z_{0,r_1}, W_{key,0,s_0} z_{0,k}>
  \nonumber \\
  &&
  >
  \frac{B}{2}
  +
  \max_{r_2 \neq k}
   < W_{query,0,s_0} z_{0,r_1}, W_{key,0,s_0} z_{0,r_2}>.
  \end{eqnarray}
Furthermore we have
\begin{equation}
    \label{ple9neweq4}
W_{value,0,s_0} z_{0,r} =
\delta_{r,j} \cdot \be_{s_3 -
  (s_0-1) \cdot d_v} .
\end{equation}

In the {\it second step of the proof} we bound the difference
between
\[
< \tilde{W}_{query,0,s_0} \tilde{z}_{0,r_1}, \tilde{W}_{key,0,s_0} \tilde{z}_{0,r_2} >
\quad
\mbox{and}
\quad
< W_{query,0,s_0} z_{0,r_1}, W_{key,0,s_0} z_{0,r_2} >
.
\]

We have
\begin{eqnarray*}
  &&
  |
  < \tilde{W}_{query,0,s_0} \tilde{z}_{0,r_1}, \tilde{W}_{key,0,s_0} \tilde{z}_{0,r_2} >
  -
  < W_{query,0,s_0} z_{0,r_1}, W_{key,0,s_0} z_{0,r_2} >
  |
  \\
  &&
  =
  |< (\tilde{W}_{query,0,s_0} - W_{query,0,s_0}) \tilde{z}_{0,r_1}
  + W_{query,0,s_0} (\tilde{z}_{0,r_1} - z_{0,r_1})
  + W_{query,0,s_0} z_{0,r_1},
  \\
  &&
  \hspace*{2cm}
   (\tilde{W}_{key,0,s_0} - W_{key,0,s_0}) \tilde{z}_{0,r_2}
  + W_{key,0,s_0} (\tilde{z}_{0,r_2} - z_{0,r_2})
  +W_{key,0,s_0}  z_{0,r_2}
  >
  \\
  &&
  \quad
  -
  < W_{query,0,s_0} z_{0,r_1}, W_{key,0,s_0} z_{0,r_2} >|
  \\
  &&
  \leq
  |< (\tilde{W}_{query,0,s_0} - W_{query,0,s_0}) \tilde{z}_{0,r_1}),
  (\tilde{W}_{key,0,s_0} - W_{key,0,s_0}) \tilde{z}_{0,r_2}>|
  \\
  &&
  \quad
  +|< (\tilde{W}_{query,0,s_0} - W_{query,0,s_0}) \tilde{z}_{0,r_1}),
  W_{key,0,s_0} (\tilde{z}_{0,r_2} - z_{0,r_2})>|
  \\
  &&
  \quad
  +|< (\tilde{W}_{query,0,s_0} - W_{query,0,s_0}) \tilde{z}_{0,r_1},
  W_{key,0,s_0}  z_{0,r_2}>|
  \\
  &&
  \quad
  +|< W_{query,0,s_0} (\tilde{z}_{0,r_1} - z_{0,r_1}),
  (\tilde{W}_{key,0,s_0} - W_{key,0,s_0}) \tilde{z}_{0,r_2}>|
  \\
  &&
  \quad
  +|< W_{query,0,s_0} (\tilde{z}_{0,r_1} - z_{0,r_1}),
  W_{key,0,s_0} (\tilde{z}_{0,r_2} - z_{0,r_2})>|
  \\
  &&
  \quad
  +|< W_{query,0,s_0} (\tilde{z}_{0,r_1} - z_{0,r_1}),
  W_{key,0,s_0}  z_{0,r_2}>|
  \\
  &&
  \quad
  +
|  < W_{query,0,s_0} z_{0,r_1},
(\tilde{W}_{key,0,s_0} - W_{key,0,s_0}) \tilde{z}_{0,r_2}>|
  \\
  &&
  \quad
  +
 | < W_{query,0,s_0} z_{0,r_1},
  W_{key,0,s_0} (\tilde{z}_{0,r_2} - z_{0,r_2})>|
  \\
  &&
  =: \sum_{i=1}^8 T_i
  .
  \end{eqnarray*}
We have
\[
T_1 \leq d_{key} \cdot \tau^2 \cdot (\|z_0\|_\infty + \delta)^2 \cdot \epsilon^2,
\]
\[
T_2 \leq   \tau \cdot (\|z_0\|_\infty + \delta) \cdot \epsilon \cdot \delta
\]
(where we have used the fact that $\tilde{z}_{0,r_2} - z_{0,r_2}$ is zero in components
less than $d+l+2$ and consequently only the first component of
$ W_{key,0,s_0} (\tilde{z}_{0,r_2} - z_{0,r_2})$
is nonzero),
\[
T_3 \leq   \tau  \cdot (\|z_0\|_\infty + \delta)
\cdot \epsilon
\cdot (|\beta| +1) \cdot   \|z_0\|_\infty 
+\tau  \cdot \|z_0\|_\infty \cdot \epsilon \cdot l \cdot \|z_0\|_\infty 
+
\tau  \cdot \|z_0\|_\infty \cdot \epsilon \cdot 2 \cdot B 
\]
(that is the consequence of the fact that the last two components of
$(\tilde{W}_{query,0,s_0} - W_{query,0,s_0}) \tilde{z}_{0,r_1}$
depend on $z_{0,r_1}$ and not on $\tilde{z}_{0,r_1}$, which follows from the
assumption that in the last two rows of
$\tilde{W}_{query,0,s_0}$ and $\tilde{W}_{key,0,s_0}$ all entries in columns
greater than $d+l+1$ are zero),
\[
T_4 \leq \delta \cdot   \tau \cdot (\|z_0\|_\infty + \delta)
\cdot \epsilon
\]
(where we have used the fact that only the first component of
$W_{query,0,s_0} (\tilde{z}_{0,r_1} - z_{0,r_1})$
is nonzero)
\[
T_5 \leq
\delta \cdot  \delta,
\]
\[
T_6 \leq
\delta \cdot  (|\beta|+1)  \cdot \|z_0\|_\infty,
\]
\[
T_7 \leq
\|z_0\|_\infty \cdot \tau \cdot \epsilon \cdot (\|z_0\|_\infty + \delta)
+
(B+l) \cdot \|z_0\|_\infty  \cdot \tau \cdot \epsilon \cdot \|z_0\|_\infty
\]
(where we have used the fact that  the last two components of
$(\tilde{W}_{key,0,s_0} - W_{key,0,s_0}) \tilde{z}_{0,r_2}$
depend on $z_{0,r_2}$ and not on $\tilde{z}_{0,r_2}$, which follows
as above from the
assumption that in the last two rows of
$\tilde{W}_{query,0,s_0}$ and $\tilde{W}_{key,0,s_0}$ all entries in columns
greater than $d+l+1$ are zero),
and
\[
T_8 \leq \|z_0\|_\infty  \cdot (|\beta|+1) \cdot \delta
\]
(where we have used the fact that $\tilde{z}_{0,r_2} - z_{0,r_2}$ is zero in components
greater $d+l+1$ and consequently only the first component of
$  W_{key,0,s_0} (\tilde{z}_{0,r_2} - z_{0,r_2})$
is nonzero).
This proves
\begin{eqnarray}
  &&
|
  < \tilde{W}_{query,0,s_0} \tilde{z}_{0,r_1}, \tilde{W}_{key,0,s_0} \tilde{z}_{0,r_2} >
  -
  < W_{query,0,s_0} z_{0,r_1}, W_{key,0,s_0} z_{0,r_2} >
  |
  \nonumber \\
  &&
  \leq
  14 \cdot d_{key} \cdot \tau^2 \cdot l \cdot (|\beta|+1) \cdot
  \|z_0\|_\infty^2 \cdot \max\{\delta^2,1\}
  \cdot \epsilon \nonumber \\
  && \quad
  +3 \cdot (|\beta|+1) \cdot \|z_0\|_\infty \cdot \max\{ \delta,1\} \cdot \delta
  \nonumber \\
  && \quad
  + 3 \cdot B \cdot \epsilon \cdot  \tau \cdot \|z_0\|_\infty^2.
  \label{ple9neweq5}
  \end{eqnarray}
Since we have
\[
\epsilon \leq \min \left\{ 1, \frac{1}{36 \cdot \tau \cdot \|z_0\|_\infty^2}
\right\},
\quad
B >
36 \cdot (|\beta|+1) \cdot \|z_0\|_\infty^2 \cdot \max\{ \delta,1\}  \cdot
\epsilon
\]
and
\[
B > 168 \cdot d_{key} \cdot \tau^2 \cdot l \cdot (|\beta|+1)
  \cdot \|z_0\|^2 \cdot \max\{\delta^2,1\} \cdot \epsilon
\]
the right-hand side of (\ref{ple9neweq5})
 is less than $B/4$.

In the {\it third step of the proof} we show
(\ref{le9neweq2a}).

To do this, we conclude from step 2 that
we have
\begin{eqnarray*}
  &&
  < \tilde{W}_{query,0,s_0} \tilde{z}_{0,r_1}, \tilde{W}_{key,0,s_0} \tilde{z}_{0,r_2}>
  -
  \max_{r_3 \neq r_2}
  < \tilde{W}_{query,0,s_0} \tilde{z}_{0,r_1}, \tilde{W}_{key,0,s_0} \tilde{z}_{0,r_3}>
  \\
  &&
  >
  \quad
  < W_{query,0,s_0} z_{0,r_1}, W_{key,0,s_0} z_{0,r_2}>
  -
  \max_{r_3 \neq r_2}
   < W_{query,0,s_0} z_{0,r_1}, W_{key,0,s_0} z_{0,r_3}>
- \frac{B}{2}.   
\end{eqnarray*}
The assertion follows from (\ref{ple9neweq2}) and (\ref{ple9neweq3}).

In the {\it fourth step of the proof} we show the assertion.
Because of (\ref{le9neweq2a}), (\ref{ple9neweq1}) and (\ref{ple9neweq4})
it suffices to show
\begin{eqnarray*}
  &&
  \Bigg\|
  \tilde{W}_{value,0,s_0} \tilde{z}_{0,r_2} \cdot
< \tilde{W}_{query,0,s_0} \tilde{z}_{0,r_1}, \tilde{W}_{key,0,s_0} \tilde{z}_{0,r_2} >
\\
&&
\hspace*{3cm}
-
W_{value,0,s_0} z_{0,r_2}
< W_{query,0,s_0} z_{0,r_1}, W_{key,0,s_0} z_{0,r_2} >
\Bigg\|_\infty
\\
&&
\leq 144 \cdot d_{key} \cdot \tau^3 \cdot l \cdot (|\beta|+1)
\cdot \|z_0\|_\infty^3 \cdot B
\cdot \max\{ \delta^3,1 \} \cdot \epsilon
\\
&&
\quad
+
24 \cdot \tau \cdot  (|\beta|+1) \cdot \|z_0\|_\infty \cdot \max\{ \delta,1 \}
\cdot \delta.
  \end{eqnarray*}
We have
\begin{eqnarray*}
  &&
  \Bigg\|
  \tilde{W}_{value,0,s_0} \tilde{z}_{0,r_2} \cdot
< \tilde{W}_{query,0,s_0} \tilde{z}_{0,r_1}, \tilde{W}_{key,0,s_0} \tilde{z}_{0,r_2} >
\\
&&
\hspace*{3cm}
-
W_{value,0,s_0} z_{0,r_2}
< W_{query,0,s_0} z_{0,r_1}, W_{key,0,s_0} z_{0,r_2} >
\Bigg\|_\infty
\\
&&
\leq
  \Bigg\|
  \tilde{W}_{value,0,s_0} \tilde{z}_{0,r_2} \cdot
\Big(< \tilde{W}_{query,0,s_0} \tilde{z}_{0,r_1}, \tilde{W}_{key,0,s_0} \tilde{z}_{0,r_2} >
\\
&&
\hspace*{5cm}
-< W_{query,0,s_0} z_{0,r_1}, W_{key,0,s_0} z_{0,r_2} > \Big)
\Bigg\|_\infty
\\
&&
\quad
+
  \Bigg\|
  (\tilde{W}_{value,0,s_0} \tilde{z}_{0,r_2} -
W_{value,0,s_0} z_{0,r_2}) \cdot
< W_{query,0,s_0} z_{0,r_1}, W_{key,0,s_0} z_{0,r_2} >
\Bigg\|_\infty
\\
&&
\leq
\|\tilde{W}_{value,0,s_0} \tilde{z}_{0,r_2}\|_\infty
\\
&&
\hspace*{1cm}
\cdot
|< \tilde{W}_{query,0,s_0} \tilde{z}_{0,r_1}, \tilde{W}_{key,0,s_0} \tilde{z}_{0,r_2} >
-< W_{query,0,s_0} z_{0,r_1}, W_{key,0,s_0} z_{0,r_2} >|
\\
&&
\quad
+
 \|
  \tilde{W}_{value,0,s_0} \tilde{z}_{0,r_2} -
W_{value,0,s_0} z_{0,r_2} \|_\infty
\cdot
|< W_{query,0,s_0} z_{0,r_1}, W_{key,0,s_0} z_{0,r_2} >|
\\
&&
\leq
(\tau+1) \cdot (1 + \epsilon) \cdot (\|z\|_\infty + \delta)
\cdot
(
  14 \cdot d_{key} \cdot \tau^2 \cdot l \cdot (|\beta|+1) \cdot
  \|z_0\|_\infty^2 \cdot \max\{\delta^2,1\}
  \cdot \epsilon \\
  && \quad
  \hspace*{1cm} +3 \cdot (|\beta|+1) \cdot \|z_0\|_\infty \cdot \max\{ \delta,1\} \cdot \delta
  + 3 \cdot B \cdot \epsilon \cdot  \tau \cdot \|z_0\|_\infty^2
  )
  \\
  &&
  \quad
  +
  \|
  \tilde{W}_{value,0,s_0} \tilde{z}_{0,r_2} -
W_{value,0,s_0} z_{0,r_2} \|_\infty
\cdot (\|z_0\|_\infty \cdot (|\beta|+\|z_0\|_\infty) + 2 \cdot B).
  \end{eqnarray*}
With
\begin{eqnarray*}
  &&
  \|
  \tilde{W}_{value,0,s_0} \tilde{z}_{0,r_2} -
W_{value,0,s_0} z_{0,r_2} \|_\infty
\\
&&
\leq
  \|
  (\tilde{W}_{value,0,s_0}- W_{value,0,s_0}) \tilde{z}_{0,r_2}  \|_\infty
  +
  \|
  W_{value,0,s_0} (\tilde{z}_{0,r_2} - z_{0,r_2}) \|_\infty
  \\
  &&
  \leq
  \tau \cdot \epsilon \cdot (\|z_0\|_\infty+ \delta)
  + 0
  \end{eqnarray*}
(where we have used 
that $\tilde{z}_{0,r_2} - z_{0,r_2}$ is zero in components
less than $d+l+2$) we get the assertion.
\hfill $\Box$

\begin{lemma}
\label{le10new}
Let $\epsilon \in [0,1)$, let $\delta \geq 0$ and let
  $\alpha \in \R$. Let $d_{ff},d_{model} \in \N$
and assume $d_{ff} \geq 4$.
Let $j_1, j_2 \in \{1, \dots, d_{model} \}$ with $j_1 \neq j_2$.
Then there exist
\[
 W_{r,1}\in \R^{d_{ff} \times d_{model}}, b_{r,1} \in \R^{d_{ff}},
  W_{r,2} \in \R^{d_{model} \times d_{ff}}, b_{r,2} \in \R^{d_{model}},
\]
where in $W_{r,1}$ in each row and in $W_{r,2}$ in each column
at most $2$ components are not equal to zero
and where all entries are bounded in absolute value by
$\max\{|\alpha|,1\}$, such that for all
$\tilde{W}_{r,1} \in \R^{d_{ff} \times d_{model}}$, 
$\tilde{b}_{r,1} \in \R^{d_{ff}}$,
$ \tilde{ W}_{r,2} \in \R^{d_{model} \times d_{ff}}$, 
$\tilde{b}_{r,2} \in \R^{d_{model}}$
with
\[
\|W_{r,1} - \tilde{W}_{r,1}\|_\infty < \epsilon,
\quad
\|b_{r,1} - \tilde{b}_{r,1}\|_\infty < \epsilon,
\quad
\|W_{r,2} - \tilde{W}_{r,2}\|_\infty < \epsilon,
\quad
\|b_{r,2} - \tilde{b}_{r,2}\|_\infty < \epsilon,
\]
and all  $y_{r,i}, \tilde{y}_{r,i} \in \R^{d_{model}}$ 
$(i \in \{1, \dots, l\})$
with
\[
\|y_{r,i}- \tilde{y}_{r,i}\|_\infty < \delta
\quad (i \in \{1, \dots, l\})
\]
and
\[
 \tilde{z}_{r,s}=\tilde{y}_{r,s}+\tilde{W}_{r,2} \cdot \sigma \left(
  \tilde{W}_{r,1} \cdot \tilde{y}_{r,s} + \tilde{b}_{r,1}
  \right) + \tilde{b}_{r,2} \quad (s \in \{1, \dots, l\})
\]
we have:
\[
\left|
\tilde{z}_{r,s}^{(j_1)}- \alpha \cdot \max\{y_{r,s}^{(j_2)},0\}
\right|
\leq 5 \cdot d_{ff} \cdot \max\{|\alpha|,1\} \cdot (\|y_{r,s}\|_\infty+1) \cdot
d_{model} \cdot (\delta +  \epsilon),
\]
\[
|\tilde{z}_{r,s}^{(j_2)}| \leq 5 \cdot d_{ff} \cdot \max\{|\alpha|,1\}
\cdot (\|y_{r,s}\|_\infty+1) \cdot d_{model} \cdot (\delta +  \epsilon)
\]
and
\[
|\tilde{z}_{r,s}^{(j)} - y_{r,s}^{(j)}| \leq 5 \cdot d_{ff}
\cdot \max\{|\alpha|,1\}
\cdot (\|y_{r,s}\|_\infty+1) \cdot d_{model} \cdot (\delta +  \epsilon)
\]
whenever $j \in \{1, \dots, d_{model}\} \setminus \{j_1,j_2\}.$

Furthermore, the assertion of the lemma holds also if we replace
$\alpha \cdot \max\{y_{r,s}^{(j_2)},0\}$ by $\alpha \cdot y_{r,s}^{(j_2)}$.
\end{lemma}

\noindent
{\bf Proof.}
Set
\[
  z_{r,s}=y_{r,s}+W_{r,2} \cdot \sigma \left(
  W_{r,1} \cdot y_{r,s} + b_{r,1}
 \right) + b_{ r,2} \quad (s \in \{1, \dots, l\}).
\]
In the {\it first step of the proof} we show that we can choose
\[
 W_{r,1}\in \R^{d_{ff} \times d_{model}}, b_{r,1} \in \R^{d_{ff}},
  W_{r,2} \in \R^{d_{model} \times d_{ff}}, b_{r,2} \in \R^{d_{model}},
\]
such that at most $9$ components are not equal to zero
and such that
\[
z_{r,s}^{(j_1)}= \alpha \cdot \max\{y_{r,s}^{(j_2)},0\}
,
\quad
z_{r,s}^{(j_2)}=0
\quad
\mbox{and}
\quad
z_{r,s}^{(j)}= y_{r,s}^{(j)}
\]
hold whenever $j \in \{1, \dots, d_{model}\} \setminus \{j_1,j_2\}.$

W.l.o.g. we assume $d_{ff}=4$  and $j_1<j_2$.
        We choose $b_{r,1}=0$, $b_{r,2}=0$,
        \[
        W_{r,1}=
        \left(
    \begin{array}{ccccccccccc}
      0 & \dots & 0& 1 &  0 &\dots & 0 & 0 &  0 & \dots & 0 \\
      0 & \dots & 0& -1 &  0 &\dots & 0 & 0 &  0 & \dots & 0 \\
      0 & \dots & 0& 0 &  0 &\dots & 0 & 1 &  0 & \dots & 0 \\
      0 & \dots & 0& 0 &  0 &\dots & 0 & -1 &  0 & \dots & 0 \\
      \end{array}
    \right),
        \]
where  all columns
except columns number $j_1$ and $j_2$ are zero,
and
        \[
        W_{r,2}=
        \left(
    \begin{array}{cccc}
      0 & 0 & 0 & 0  \\
      \vdots & && \vdots \\
      0 & 0 & 0 & 0  \\
     -1 & 1 & \alpha & 0  \\
      0 & 0 & 0 & 0  \\
      \vdots & && \vdots \\
      0 & 0 & 0 & 0  \\
      0 & 0 &  -1 & 1 \\
      0 & 0 &  0 & 0 \\
      \vdots & && \vdots \\
      0 & 0 & 0 & 0  
     \end{array}
    \right),
        \]
where all rows except row number $j_1$ and $j_2$ are zero.
Then we have
\begin{eqnarray*}
          &&
W_{2,r} \cdot \sigma \left(
  W_{1,r} \cdot y_{r,s} + b_{1,r}
  \right) + b_{2,r}
  \\
  &&
  =
  \left(
  \begin{array}{c}
    0 \\
    \vdots \\
    0 \\
  \alpha \cdot \sigma(y_{r,s}^{(j_2)})
    -   ( \sigma(y_{r,s}^{(j_1)})-    \sigma(-y_{r,s}^{(j_1})) \\
    0 \\
    \vdots \\
    0 \\
   -   ( \sigma(y_{r,s}^{(j_2)})-    \sigma(-y_{r,s}^{(j_2)})) \\
    0 \\
    \vdots \\
    0 \\
    \end{array}
  \right).
        \end{eqnarray*}
        Because of
        \[
\sigma(u)-\sigma(-u)=u
        \]
        for $u \in \R$ this implies the assertion of the first step.

In the {\it second step of the proof} we show
\[
| \tilde{z}_{r,s}^{(j)} - z_{r,s}^{(j)}|\leq 5 \cdot d_{ff}
\cdot \max\{|\alpha|,1\}
\cdot (\|y_{r,s}\|_\infty+1) \cdot d_{model} \cdot  (\delta + \epsilon).
\]
We have
\begin{eqnarray*}
&&
\|  \tilde{W}_{r,1} \cdot \tilde{y}_{r,s} + \tilde{b}_{r,1}
-
(
 W_{r,1} \cdot y_{r,s} + b_{r,1}
)
\|_\infty
\\
&&
\leq
\|  (\tilde{W}_{r,1}-W_{r,1}) \cdot \tilde{y}_{r,s}
\|_\infty
+
\|
 W_{r,1} \cdot (\tilde{y}_{r,s}  - y_{r,s}) 
\|_\infty
+ \epsilon\\
&&
\leq
d_{model} \cdot \epsilon \cdot (\|y_{r,s}\|_\infty+ \delta) +
   \delta+\epsilon
\\
&&
\leq \delta + d_{model} \cdot \epsilon \cdot (\|y_{r,s}\|_\infty+1+\delta), 
\end{eqnarray*}
which implies
\begin{eqnarray*}
&&
\| \sigma ( \tilde{W}_{r,1} \cdot \tilde{y}_{r,s} + \tilde{b}_{r,1} )
-
\sigma(  W_{r,1} \cdot y_{r,s} + b_{r,1})
\|_\infty
\\
&&
\leq \delta + d_{model} \cdot \epsilon \cdot (\|y_{r,s}\|_\infty+1+\delta)
\end{eqnarray*}
and
\begin{eqnarray*}
&&
\| \sigma ( \tilde{W}_{r,1} \cdot \tilde{y}_{r,s} + \tilde{b}_{r,1} )
\|_\infty
\\
&&
\leq \delta + d_{model} \cdot \epsilon \cdot (\|y_{r,s}\|_\infty+1+\delta)
+
\|
\sigma(  W_{r,1} \cdot y_{r,s} + b_{r,1})
\|_\infty
\\
&&
\leq \delta + d_{model} \cdot \epsilon \cdot (\|y_{r,s}\|_\infty+1+\delta)
+
\|y_{r,s}\|_\infty
.
\end{eqnarray*}
From this we conclude
\begin{eqnarray*}
&&
| \tilde{z}_{r,s}^{(j)} - z_{r,s}^{(j)}|
\\
&&
\leq
\bigg\|
\tilde{y}_{r,s}+\tilde{W}_{r,2} \cdot \sigma \left(
  \tilde{W}_{r,1} \cdot \tilde{y}_{r,s} + \tilde{b}_{r,1}
  \right) + \tilde{b}_{r,2}
\\
&&
\hspace*{3cm}
-
\left(
y_{r,s}+W_{r,2} \cdot \sigma \left(
  W_{r,1} \cdot y_{r,s} + b_{r,1}
 \right) + b_{ r,2}
\right)
\bigg\|_\infty \\
&&
\leq
\delta
+
\left\|
\tilde{W}_{r,2} \cdot \sigma \left(
  \tilde{W}_{r,1} \cdot \tilde{y}_{r,s} + \tilde{b}_{r,1}
  \right)
-
W_{r,2} \cdot \sigma \left(
  W_{r,1} \cdot y_{r,s} + b_{r,1}
 \right)
\right\|_\infty + \epsilon
\\
&&
\leq
\delta
+
\left\|
(\tilde{W}_{r,2} -
W_{r,2}) \cdot \sigma \left(
  \tilde{W}_{r,1} \cdot \tilde{y}_{r,s} + \tilde{b}_{r,1}
  \right)
\right\|_\infty 
\\
&&
\quad
+
\left\|
W_{r,2} \cdot 
\left(
\sigma \left(
  \tilde{W}_{r,1} \cdot \tilde{y}_{r,s} + \tilde{b}_{r,1}
  \right)
-
\sigma \left(
  W_{r,1} \cdot y_{r,s} + b_{r,1}
 \right)
\right)
\right\|_\infty + \epsilon
\\
&&
\leq
\delta
+ d_{ff} \cdot
\left\|
\tilde{W}_{r,2} -
W_{r,2}
\right\|_\infty
 \cdot 
\left\|
\sigma \left(
  \tilde{W}_{r,1} \cdot \tilde{y}_{r,s} + \tilde{b}_{r,1}
  \right)
\right\|_\infty 
\\
&&
\quad
+ d_{ff} \cdot
\left\|
W_{r,2} \right\|_\infty
\cdot 
\left\|
\sigma \left(
  \tilde{W}_{r,1} \cdot \tilde{y}_{r,s} + \tilde{b}_{r,1}
  \right)
-
\sigma \left(
  W_{r,1} \cdot y_{r,s} + b_{r,1}
 \right)
\right\|_\infty + \epsilon
\\
&&
\leq
\delta
+ d_{ff} \cdot
\epsilon \cdot
(\delta + d_{model} \cdot \epsilon \cdot (\|y_{r,s}\|_\infty+1+\delta)
+
\|y_{r,s}\|_\infty
)\\
&&
\quad
+
d_{ff} \cdot \max\{|\alpha|,1\} \cdot (\delta + d_{model} \cdot \epsilon \cdot
   (\|y_{r,s}\|_\infty+1+\delta))
+ \epsilon
\\
&&
\leq
5 \cdot d_{ff} \cdot \max\{|\alpha|,1\} \cdot (\|y_{r,s}\|_\infty+1) \cdot
d_{model} \cdot (\delta + \epsilon).
\end{eqnarray*}

By slightly changing the matrix $W_{r,2}$ we conclude 
that the assertion also holds upon replacing 
$\alpha \cdot \max\{y_{r,s}^{(j_2)},0\}$ by $\alpha \cdot y_{r,s}^{(j_2)}$.
\hfill $\Box$

\begin{lemma}
  \label{le11new}
  Let
$I \geq d+l+4$, let
  $\tau \in \{l+1,l+2, \dots, l+d+1\}$.
          Let $c \geq A \geq 1$, let $K \in \N$, let $u_k \in [-A,A]$ $(k=1, \dots, K-1)$, set
          \[
B_j(x)=x^j \quad \mbox{for } j=0,1,\dots, M
\]
and set
\[
B_j(x)=(x-u_{j-M})_+^M  \quad \mbox{for } j=M+1,M+2,\dots, M+K-1.
\]
Let $i \in \{d+l+4, \dots, I\}$. 
Let $h \in \N$ with $1 \leq h \leq n$
and for $s \in \{2, \dots, h\}$ let
              $j_{s,1}, \dots, j_{s,d} \in \{0,1, \dots, M+K-1\}$
              and $\alpha_s \in \R$.
              Let $d_{key} \geq 3$ and set $d_{ff} = 2 \cdot h + 2$.
              Then there exists a transformer encoder consisting of
               $ M \cdot (d \cdot l)+1$ pairs of layers, where the first layer is a
              multi-head attention layer with $h$ attention units and
              the second layer is a pointwise feedforward
              neural network, and
              where all matrices in the attention heads have in each row at most
              $\tau$ nonzero entries, and where all matrices $W_{r,1}$ and $W_{r,2}$
              in the pointwise feedforward neural networks have the property that
              in each row in $W_{r,1}$ and in each column in $W_{r,2}$ at
              most $\tau$ of the entries are nonzero
              and where all matrices
        and vectors depend only on $(u_k)_k$ and
        $\alpha_s$ $(s \in \{1, \dots, k\})$ and all entries are
        bounded
        in absolute value by
        \[
        c_{27} \cdot n^{12^{M \cdot (d \cdot l)}} \cdot (d_{model})^{12^{M \cdot (d \cdot l)}}
        \cdot c^{2 \cdot 12^{M \cdot (d \cdot l)}},
        \]
        and where the matrices $\tilde{W}_{query,0,s}$,
        $\tilde{W}_{key,0,s}$ and
        $\tilde{W}_{value,0,s}$ satisfy the assumptions of Lemma \ref{le9new},
which has the following property:
Any transformer network which has the same structure as the above
Transformer network and whose weights are in supremum norm no
further than $\epsilon$ away from the weights of the above network
for some
\[
0 \leq \epsilon \leq \min \left\{ 1, \frac{1}{36 \cdot \tau \cdot
  (2 \cdot A)^{M \cdot d \cdot l}}
\right\}
\]
has the property that if it gets
        as input $\tilde{z}_0$ 
which satisfies (\ref{le9neweq4}) and
\[
\|\tilde{z}_0-z_0\|_\infty \leq c \cdot \epsilon
\]
for some $z_0 \in [-A,A]^{l \cdot d_{model}}$ defined as in
Subsection \ref{se2sub1} (which encodes in particular
$x=(x_1^T, \dots, x_l^T) \in \R^{d \cdot l}$)
it  produces
        as output $\tilde{z}_{M \cdot d+1}$, which satisfies
for $n$ sufficiently large
        \[
|\tilde{z}_{M \cdot d,1}^{(i)} -         
\sum_{s=2}^h
        \alpha_s \prod_{k=1}^{d \cdot l} B_{j_{s,k}}(x^{(k)})|
\leq  c_{28}  \cdot n^{6^{M \cdot (d \cdot l)}+1} \cdot (d_{model})^{6^{M \cdot (d \cdot l)}+1} \cdot c^{8^{M \cdot (d \cdot l)+2}+ 2 \cdot (M \cdot d \cdot l+1)}  \cdot \epsilon
\]
and
\[
|\tilde{z}_{M \cdot d,j}^{(l)} - z _{M \cdot d,j}^{(l)}|
\leq  c_{28} \cdot n^{6^{M \cdot (d \cdot l)}+1} \cdot (d_{model})^{6^{M \cdot (d \cdot l)}+1} \cdot c^{8^{M \cdot (d \cdot l)+2} \cdot (M \cdot d \cdot l+1)} \cdot \epsilon
\]
\quad
whenever $j>1$ or
\begin{eqnarray*}
  &&
l \in \{1, \dots, d_{model}\} \setminus
\{
i, I+d+l+2, I+d+l+3,
2 \cdot I+d+l+2, 
\\
&&
\hspace*{2.5cm}
2 \cdot I+d+l+3,
\dots,
(h-1) \cdot I+d+l+2, (h-1) \cdot I+d+l+3
\}.
\end{eqnarray*}
\end{lemma}

\noindent
{\bf Proof.} 
In the {\it first step of the proof} we show that the $h-1$
products of the
B-splines
in the sum above 
can be computed in the first $M \cdot (d \cdot l)$ pairs of attention
heads and pointwise feedforward network.

The basic idea is as follows.
                Each attention head of the network works only on one
                of the
parts $2, \dots, h$ of length $I$ of the input.
                It uses the fact that
each $B_j(x)$ can be written as
            \[
B_j(x)= \prod_{k=1}^M B_{j,k}(x)
            \]
where $B_{j,k}(x)$ is one of the functions
            \[
            x \mapsto 1, \quad x \mapsto x \quad
            \mbox{and} \quad x \mapsto (x -u_r)_+. 
            \]            
            Using Lemma \ref{le9new} (with a suitable value for $B$, which will
            be chosen in the third step of the proof) and Lemma \ref{le10new}
            we can combine an attention layer and
            a pointwise feedforward layer such that
            the following holds:
            They get as input an approximation
            $\tilde{z}_{0,j}$ of $z_{0,j}$ where
            $z_{0,j}$ is given as in Lemma \ref{le9new}
            and where the component $(s-1) \cdot I + d+l+2$ of $z_{0,j}$
            is zero, and they modify the components
            $(s-1) \cdot I + d+l+2$  and $(s-1) \cdot I + d+l+3$ of $z_{0,j}$.
            More precisely, they combine the attention head of Lemma \ref{le9new}
            and the pointwise feedforward neural network of Lemma \ref{le10new}
            such that they produce
            an output $\tilde{y}_j$ where $\tilde{y}_1^{( (s-1) \cdot I + d+l+3)}$ is
            an approximation of the product
            of an approximation of either
$z_{0,1}^{(d+1)}=1$ or $z_{0,1}^{((s-1) \cdot I +d+l+3)}$ and one of the functions
            \[
            x^{(k)} \mapsto 1, \quad x^{(k)} \mapsto x^{(k)} \quad
            \mbox{and} \quad x^{(k)} \mapsto (x^{(k)} -u_r)_+
            \quad (r \in \{1, \dots, K-1\})
            \]
            and where $\tilde{y}_j^{(r)}$ is approximately equal to $z_{0,j}^{(r)}$ otherwise $(r \in \{ (s-1) \cdot I +1, \dots, s \cdot I\}
\setminus \{(s-1) \cdot I + d+l+2, (s-1) \cdot I +d+l+3\}
            )$.
             Using this $M \cdot (d \cdot l)$ times  we get an approximation of
             \begin{equation}
               \label{ple11neweq1}
       \alpha_s \prod_{k=1}^{d \cdot l} B_{j_{s,k}}(x^{(k)})
\end{equation}
in $z_{M \cdot d,1}^{((s-1) \cdot I +(d+l+3))}$ for $s=2, \dots, h$.

In the {\it second step of the proof} we show how one pair
of attention head and pointwise feedforward neural network
can be used to compute the sum of the values in (\ref{ple11neweq1}).
To do this, we choose
                $W_{value,M \cdot (d \cdot l)+1}=0$ (which results in $y_{M \cdot (d \cdot l) +
                  1} = z_{M \cdot (d \cdot l)}$
and $\tilde{y}_{M \cdot (d \cdot l) +
                  1} \approx \tilde{z}_{M \cdot (d \cdot l)}$)
                and choose $W_{1, M \cdot (d \cdot l)+1}$, $b_{1, M \cdot (d \cdot l)+1}$,
                $W_{2, M \cdot (d \cdot l)+1}$, $b_{2, M \cdot (d \cdot l)+1}$, 
                such that
                \begin{eqnarray*}
                  &&
                  \hspace*{-0.6cm}
W_{2, M \cdot (d \cdot l)+1} \cdot \sigma \left(
  W_{1, M \cdot (d \cdot l)+1} \cdot y_{M \cdot (d \cdot l)+1} + b_{1, M \cdot (d \cdot l)+1}
  \right) + b_{2, M \cdot (d \cdot l)+1}
  \\[0.3cm]
  &&
                  \hspace*{-0.6cm}
  =
  \left(
  \begin{array}{c}
    0 \\
    \vdots \\
    0 \\
    -   ( \sigma(y_{M \cdot d+1, 1}^{(i) })-    \sigma(-y_{M \cdot d+1,1}^{(i) })) +
    \sigma(
\sum_{s=2}^h y_{M \cdot d+1, 1}^{(s-1) \cdot I +  (d+l+3)  }
)
-
   \sigma(-
\sum_{s=2}^h y_{M \cdot d+1, 1}^{(s-1) \cdot I +  (d+l+3)  }
) 
\\
    0 \\
    \vdots \\
    0 \\
    \end{array}
  \right)
  \end{eqnarray*}
                holds, where the nonzero entry is in row number $i$.

                In the {\it third step of the proof} we analyze the error
                occurring in the above approximation. Here we describe in
                particular how the values of $B$ in the application of Lemma \ref{le9new} need to be chosen.
                In the applications of Lemma \ref{le9new} we will have
                \[
                \|z_0\|_\infty \leq (2 \cdot A)^{M \cdot d \cdot l}
                \quad \mbox{and} \quad
                |\beta| \leq A.
                \]

                In the first layer we set
                \[
B=B_1= c_{29} \cdot c^{2 \cdot M \cdot d \cdot l +3},
                \]
                which implies
                \begin{eqnarray*}
                  &&
168 \cdot d_{key} \cdot \tau^2 \cdot l \cdot (|\beta|+1)
\cdot \|z_0\|_\infty^2 \cdot \max\{(c \cdot \epsilon)^2,1\}
\leq c_{30} \cdot A^{2 \cdot M \cdot d \cdot l+1} \cdot \max\{(c \cdot \epsilon)^2,1\} \\
&& \leq
c_{30} \cdot c^{2 \cdot M \cdot d \cdot l+3} \cdot \max\{\epsilon^2,1\} \leq B_1.
                \end{eqnarray*}
                From this we can conclude by Lemma \ref{le9new} that the output
                of the first attention head has an error not exceeding 
                \begin{eqnarray*}
                  &&
136 \cdot d_{key} \cdot \tau^3 \cdot l \cdot (|\beta|+1)
\cdot \|z_0\|_\infty^3 \cdot B_1
\cdot \max\{ (c \cdot \epsilon)^3,1 \} \cdot \epsilon
\\
&&
\quad
+
25 \cdot  (|\beta|+1) \cdot \|z_0\|_\infty \cdot \max\{ c \cdot \epsilon,1 \}
\cdot (c \cdot \epsilon)
\\
&&
\leq
c_{31} \cdot c^{5 \cdot M \cdot d \cdot l +7} \cdot \epsilon.
                \end{eqnarray*}
                Application of Lemma \ref{le10new}
                (where the input is bounded in absolute value by
                $c_{32} \cdot A^{M \cdot d \cdot l} \leq
                c_{32} \cdot c^{M \cdot d \cdot l}$)
                yields
                that after the pointwise feedforward neural network
                the error is maximal
                \[
\delta_1 = c_{33} \cdot n \cdot d_{model} \cdot c^{6 \cdot M \cdot d \cdot l +7} \cdot \epsilon.
                \]
                In the second level we set
                \[
B=B_2 = c_{34} \cdot n^2 \cdot d_{model}^2 \cdot c^{14 \cdot M \cdot d \cdot l +14}
                \]
                which implies
                \begin{eqnarray*}
                  &&
168 \cdot d_{key} \cdot \tau^2 \cdot l \cdot (|\beta|+1)
\cdot \|z_0\|_\infty^2 \cdot \max\{\delta_1^2,1\}
\\
&&
\leq  c_{35} \cdot
  c^{2 \cdot M \cdot d \cdot l}
\max\{(n \cdot d_{model} \cdot c^{6 \cdot M \cdot d \cdot l +7} \cdot \epsilon)^2,1\}
\\
&&
 \leq
c_{35} \cdot n^2 \cdot d_{model}^2 \cdot c^{14 \cdot M \cdot d \cdot l +14} \cdot \max\{\epsilon^2,1\} \leq B_2.
                \end{eqnarray*}
                From this we can conclude by Lemma \ref{le9new} that the output
                of the second attention head has error not exceeding
                \begin{eqnarray*}
                  &&
136 \cdot d_{key} \cdot \tau^3 \cdot l \cdot (|\beta|+1)
\cdot \|z_0\|_\infty^3 \cdot B_2
\cdot \max\{ (c_{33} \cdot n \cdot d_{model} \cdot c^{14 \cdot M \cdot d \cdot l +14} \cdot \epsilon)^3,1 \} \cdot \epsilon
\\
&&
\quad
+
25 \cdot  (|\beta|+1) \cdot \|z_0\|_\infty \cdot \max\{ c_{33} \cdot n \cdot d_{model}\cdot c^{14 \cdot M \cdot d \cdot l +14} \cdot \epsilon,1 \}
\cdot (c_{33} \cdot n \cdot d_{model} \\
&&
\quad
\cdot c^{14 \cdot M \cdot d \cdot l +14} \cdot \epsilon)
\\
&&
\leq
c_{36}  \cdot n^{5} \cdot (d_{model})^{5} \cdot c^{59\cdot M \cdot d \cdot l +57} \cdot \epsilon.
                \end{eqnarray*}
                Application of Lemma \ref{le10new}
(where the input is bounded in absolute value by $c_{37} \cdot c^{M \cdot d \cdot l}$)
                yields
                that after the second pointwise feedforward neural network
                the error is bounded above by
                \[
                c_{38} \cdot n \cdot c^{M \cdot d \cdot l} \cdot d_{model} \cdot n^5
                \cdot d_{model} \cdot c^{59 \cdot M \cdot d \cdot l +57} \cdot \epsilon
                \leq
                c_{38} \cdot n^{6^1} \cdot (d_{model})^{6^1} \cdot c^{8^2 \cdot (M \cdot d \cdot l +1)} \cdot \epsilon
                =: \delta_2.
                \]
                Arguing recursively with value
                \[
                B_r= c_{39,r}  \cdot n^{2 \cdot 6^{r-1}} \cdot (d_{model})^{2 \cdot 6^{r-1}} \cdot
                c^{4 \cdot 8^r \cdot (M \cdot d \cdot l +1)}
                \]
                on level $r \in \{1, \dots, M \cdot (d \cdot l)\}$ we see that after level
                $r$ the error of the output is at most
                \[
                \delta_r = c_{40,r}  \cdot n^{6^r} \cdot (d_{model})^{6^r} \cdot c^{8^{r+1}
                \cdot (M \cdot d \cdot l +1)} \cdot \epsilon.
                \]
                The last pair of attention head and pointwise feedforward neural network
                in level $M \cdot (d \cdot l) +1$, where the entries in all
                matrices are bounded by a constant and all entries in all
                matrices in the attention head are all close to zero, increases
                this error at most by a factor
                \[
\| z_{M \cdot (d \cdot l)}\|_\infty \cdot c_{41} \cdot h \cdot d_{model}
\]
(cf., Step 2 in the proof of Lemma \ref{le10new}),
                which implies that the error of the output of our transformer network
                is bounded by
                \[
 c_{42}  \cdot n^{ 6^{M \cdot d \cdot l}+1} \cdot (d_{model})^{6^{M \cdot d \cdot l}+1} \cdot c^{8^{M \cdot d \cdot l+2} \cdot (M \cdot d \cdot l +1)} \cdot  \epsilon.
                \]
                \quad \hfill $\Box$

                    Next we show how we can approximate a function
                    which satisfies
                    a hierarchical composition model
                    by a Transformer encoder.
                    In order to formulate this result, we introduce some
                    additional notation. In order to
compute a function $h_1^{(\kappa)} \in \mathcal{H}(\kappa, \mathcal{P})$
                    one has to compute different hierarchical composition models of some level $i$ $(i\in \{1, \dots, \kappa-1\})$. Let $\tilde{N}_i$ denote the number of hierarchical composition models of level $i$, needed to compute $h_1^{(\kappa)}$. 
Let
\begin{align}
\label{h}
h_j^{(i)}: \R^{d \cdot l} \to \R 
\end{align}
be the $j$--th hierarchical composition model of some level $i$ ($j \in \{1, \ldots, \tilde{N}_i\}, i \in \{1, \ldots, \kappa\}$),
that applies a $(p_j^{(i)}, C)$--smooth function $g_j^{(i)}: \R^{K_j^{(i)}} \to \R$ with $p_j^{(i)} = q_j^{(i)} + s_j^{(i)}$, $q_j^{(i)} \in \N_0$ and $s_j^{(i)} \in (0,1]$, where $(p_j^{(i)}, K_j^{(i)}) \in \mathcal{P}$
(and $K_j^{(1)}=d \cdot l$ $(j=1, \dots, \tilde{N}_1)$).
  With this notation we can describe
 the computation of $h_1^{(\kappa)}(\bold{x})$ recursively as follows:
    \begin{equation}\label{hji}
  h_j^{(i)}(\bold{x}) =  g_{j}^{(i)}\left(h^{(i-1)}_{\sum_{t=1}^{j-1} K_t^{(i)}+1}(\bold{x}), \dots, h^{(i-1)}_{\sum_{t=1}^j K_t^{(i)}}(\bold{x}) \right)
  \end{equation}
holds for $j \in \{1, \dots, \tilde{N}_i\}$ and $i \in \{2, \dots, \kappa\}$,
  and
    \begin{equation}\label{hj1}
      h_j^{(1)}(\bold{x}) = g_j^{(1)}\left(x_{\pi(\sum_{t=1}^{j-1} K_t^{(1)}+1)}, \dots,
      x_{\pi(\sum_{t=1}^{j} K_t^{(1)})}\right) 
    \end{equation}
    holds
        for $j \in \{1, \dots, \tilde{N}_1 \}$
  for some function $\pi: \{1, \dots, \tilde{N}_1\} \to \{1, \dots, d\}$. 
  Here
  the recursion
\begin{align}
\label{N}
\tilde{N}_l = 1 \ \text{and} \ \tilde{N}_{i} = \sum_{j=1}^{\tilde{N}_{i+1}} K_j^{(i+1)} 
\quad (i \in \{1, \dots, \kappa-1\})
\end{align}
holds.

                    \begin{theorem}
                      \label{th3}
                        Let $\tau \in \{l+1,l+2, \dots, l+d+1\}$.
Let $A \geq 1$,
                      let $m: \mathbb{R}^{d \cdot l} \to \mathbb{R}$ be contained in the class
                      $\mathcal{H}(\kappa, \mathcal{P})$ for some $\kappa \in \N$ and $\mathcal{P} \subseteq [1,\infty) \times \N$.  Let $\tilde{N}_i$ be defined as in \eqref{N}. Each $m$ consists of different functions $h_j^{(i)}$ $(j \in \{1, \ldots, \tilde{N}_i\},$
 $ i\in \{1, \dots, \kappa\})$ defined as in \eqref{h}, \eqref{hji} and \eqref{hj1}. 
 Assume that the corresponding functions $g_j^{(i)}$ are Lipschitz continuous with Lipschitz constant $C_{Lip} \geq 1$ and satisfy
  \begin{equation*}
  \|g_j^{(i)}\|_{C^{q_j^{(i)}}(\R^{K_j^{(i)}})} \leq c_{43}
  \end{equation*}
  for some constant $c_{43} >0$. Denote by $K_{max} = \max_{i,j} K_j^{(i)} < \infty$ the maximal input dimension and set $q_{max} = \max_{i,j} q_j^{(i)} < \infty$,
  where $q_j^{(i)}$ is the integer part of the smoothness $p_j^{(i)}$
  of $g_j^{(i)}$.  Let $A \geq 1$.
  Choose $h \in \N$ such that
  \begin{equation}
    \label{th3eq1}
c_{44} \leq h \leq n 
\end{equation}
holds for $n$ large and for some sufficiently large constant $c_{44}$, choose
  \[
I \geq 
\sum_{i=1}^\kappa \tilde{N}_i+
d+l+4
 \quad \mbox{and} \quad d_{key} \geq 3
\]
  and set
  \[
  N= I \cdot (q_{max} \cdot K_{max} +1), \quad
  d_{model} =h \cdot I , \quad d_v=I.
  \]
  Then there exists a transformer network $f_{\vartheta}$,
  where the matrices in the attention heads  have
in each row at most nonzero $\tau$
entries,
where all matrices $W_{r,1}$ and $W_{r,2}$ in the pointwise feedforward
neural networks have the property that in each row of $W_{r,1}$ and
in each column of $W_{r,2}$ there are at most $\tau$ nonzero
components,
and where all parameters are bounded
in absolute value by $c_{45} \cdot n^{c_{46}}$ provided $c_{45}, c_{46}>0$
are sufficiently large,
such that for each Transformer network $f_{\tilde{\vartheta}}$ which has the same structure
and which weights are in supremum norm not further away than
\[
0 \leq \epsilon \leq \frac{1}{c_{47}}
\]
from the
weights of this network for some suitable large constant $c_{47} \geq 1$
and where the matrices $\tilde{W}_{query,r,s}$, $\tilde{W}_{key,r,s}$
and $\tilde{W}_{value,r,s}$ satisfy the assumptions of Lemma \ref{le9new},
satisfies for $n$ large
\begin{eqnarray*}
  \|f_{\tilde{\vartheta}}-m\|_{\infty, [-A,A]^{d \cdot l}} 
  & \leq &  c_{48} \cdot (K_{max}+1)^\kappa \cdot
  \max_{j,i} h^{-p_j^{(i)}/K_j^{(i)}} \\
  && +
  c_{49} \cdot n^{(I+1) \cdot (q_{max} \cdot K_{max} +2)} \cdot d_{model}^{(I+1) \cdot (q_{max} \cdot K_{max} +1)} \cdot \epsilon
  \end{eqnarray*}
  provided $\epsilon \geq 0$ satisfies
  \begin{equation}
    \label{th3eq1}
    \epsilon \leq \frac{1}{
2 \cdot  c_{50} \cdot n^{8^{(\sum_{i=1}^\kappa \tilde{N}_s) \cdot (2 \cdot q_{max} \cdot K_{max} +4)}} \cdot d_{model}^{8^{(\sum_{i=1}^\kappa \tilde{N}_s) \cdot (2 \cdot q_{max} \cdot K_{max} +4)}}
      }.
    \end{equation}
                    \end{theorem}

\noindent
{\bf Proof.}
                  From the Lipschitz continuity of the
                        $g_j^{(i)}$ and the recursive
                        definition of the $h_j^{(i)}$ we conclude
                        that there exists $1 \leq \bar{A}  \leq c_{51} \cdot A$ such that
                        \begin{equation}
                          \label{pth3eq1}
h_j^{(i)}(x) \in [-\bar{A},\bar{A}]
\end{equation}
holds for all $x \in [-A,A]^{d \cdot l}$, $j \in \{1, \dots, \tilde{N}_i\}$
and $i \in \{1, \dots, \kappa-1\}$.

                        Our transformer encoder successively
                        approximates 
                        $h_1^{(1)}(x)$, \dots, $h_{N_1}^{(1)}(x)$,
                        $h_1^{(2)}(x)$, \dots, $h_{N_2}^{(2)}(x)$,
                        \dots, $h^{(\kappa)}_1(x)$
                        and saves the computed values successively in
                        $z_{r,1}^{(d+l+5)}$, $z_{r,1}^{(d+l+6)}$, \dots,
                        $z_{r,1}^{(d+l+4 + \sum_{i=1}^\kappa \tilde{N}_i)}$. 
                        Here $h_i^{(j)}$ is
                        approximated by computing
                        in a first step a truncated power basis
                        of a tensor product spline space
                        of degree $q_{i}^{(j)}$ on an equidistant grid in
                        \[
                          [-\bar{A}-1,\bar{A}+1]^{K_i^{(j)}}
                          \]
                          consisting of $h-1$ basis
                        functions, which are evaluated at the arguments
                        of $h_i^{(j)}$ in (\ref{hj1}),
                        and by using in a second step a linear combination of these basis functions to approximate
                        \[
                        g_{j}^{(i)}\left(h^{(i-1)}_{\sum_{t=1}^{j-1} K_t^{(i)}+1}(\bold{x}), \dots, h^{(i-1)}_{\sum_{t=1}^j K_t^{(i)}}(\bold{x}) \right)
                        .
                        \]
                        The approximate computation of this truncated power basis can be done as in Lemma \ref{le11new} using layers $(\tilde{N}_{i-1}+j-1) \cdot ( q_{max} \cdot K_{max} +1)+1$
                        till
$(\tilde{N}_{i-1}+j) \cdot ( q_{max} \cdot K_{max} +1)$
                        of our transformer encoder.
                        Here the computed values of this basis will have an error not exceeding
                        \[
                        c_{51} \cdot n^{8^{(\sum_{s=1}^{i-1} \tilde{N}_s +j) \cdot (2 \cdot q_{max} \cdot K_{max}+4)}} \cdot
                        d_{model}^{8^{(\sum_{s=1}^{i-1} \tilde{N}_s +j) \cdot (2 \cdot q_{max} \cdot K_{max}+4)}} \cdot \epsilon.
                        \]
                       Using standard approximation results from spline
                        theory (cf., e.g., Theorem 15.2 and
                        proof of Theorem 15.1 in Gy\"orfi et al. (2002)
                        and Lemma 1 in Kohler (2014))
                        and the Lipschitz continuity of $g_j^{(i)}$
                        this results in an approximation
                        \[
\tilde{g}_{j}^{(i)}
                        \]
                        of $g_{j}^{(i)}$ which satisfies
                        \begin{eqnarray}
                          \label{pth3eq2}
                          &&
                          \| \tilde{g}_{j}^{(i)} - g_{j}^{(i)} \|_{\infty, [-\bar{A}-1,\bar{A}+1]^{K_j^{(i)}}}
                          \\
                          && \nonumber
                        \leq
                        c_{52}  \cdot h^{-p_{j}^{(i)}/K_j^{(i)}}+ c_{51} \cdot h \cdot n^{8^{(\sum_{s=1}^{i-1} \tilde{N}_s +j) \cdot (2 \cdot q_{max} \cdot K_{max}+4)}} \cdot d_{model}^{8^{(\sum_{s=1}^{i-1} \tilde{N}_s +j) \cdot (2 \cdot q_{max} \cdot K_{max}+4)}} \cdot \epsilon
                          \\
                          && \nonumber
                        \leq
                        c_{52}  \cdot h^{-p_{j}^{(i)}/K_j^{(i)}}+ c_{51}  \cdot n^{8^{(\sum_{s=1}^{i-1} \tilde{N}_s +j) \cdot (2 \cdot q_{max} \cdot K_{max}+4)}} \cdot d_{model}^{8^{(\sum_{s=1}^{i-1} \tilde{N}_s +j) \cdot (2 \cdot q_{max} \cdot K_{max}+4)}}  \cdot \epsilon
.
                        \end{eqnarray}
                        The approximation
                        $\tilde{h}_1^{(\kappa)}(\bold{x})$
                        of $h_1^{(\kappa)}(\bold{x})$ which our transformer encoder computes
                        is defined  as follows:
    \[
  \tilde{h}_j^{(1)}(\bold{x}) = \tilde{g}_j^{(1)}\left(x_{\pi(\sum_{t=1}^{j-1} K_t^{(1)}+1)}, \dots, x_{\pi(\sum_{t=1}^{j} K_t^{(1)})}\right)
  \]
for  $j \in \{1, \dots, \tilde{N}_1\}$
  and
  \[
    \tilde{h}_j^{(i)}(\bold{x}) =  \tilde{g}_{j}^{(i)}\left(
    \tilde{h}^{(i-1)}_{\sum_{t=1}^{j-1} K_t^{(i)}+1}(\bold{x}), \dots,
    \tilde{h}^{(i-1)}_{\sum_{t=1}^j K_t^{(i)}}(\bold{x}) \right)
  \]
  for $j \in \{1, \dots, \tilde{N}_i\}$ and $i \in \{2, \dots, \kappa\}$.

  Assume that (\ref{pth3eq1}) holds.
  From (\ref{th3eq1}), (\ref{pth3eq1}) and (\ref{pth3eq2}) we 
  conclude 
  \[
  |\tilde{h}_j^{(i)}(\bold{x})| \leq
  | \tilde{h}_j^{(i)}(\bold{x}) - h_j^{(i)}(\bold{x})|
  + |h_j^{(i)}(\bold{x})| \leq \bar{A}+1.
  \]
  Consequently we get from  (\ref{pth3eq2})
for $n$ sufficiently large
  \begin{eqnarray*}
    &&
    |
    \tilde{h}_j^{(i)}(\bold{x})
    -
    h_j^{(i)}(\bold{x})
    |
    \\
    &&
\leq
|
\tilde{g}_{j}^{(i)}\left(
    \tilde{h}^{(i-1)}_{\sum_{t=1}^{j-1} K_t^{(i)}+1}(\bold{x}), \dots,
    \tilde{h}^{(i-1)}_{\sum_{t=1}^j K_t^{(i)}}(\bold{x}) \right)
    -
g_{j}^{(i)}\left(
    \tilde{h}^{(i-1)}_{\sum_{t=1}^{j-1} K_t^{(i)}+1}(\bold{x}), \dots,
    \tilde{h}^{(i-1)}_{\sum_{t=1}^j K_t^{(i)}}(\bold{x}) \right)
    |
    \\
    &&
    \quad
    +
    |
g_{j}^{(i)}\left(
    \tilde{h}^{(i-1)}_{\sum_{t=1}^{j-1} K_t^{(i)}+1}(\bold{x}), \dots,
    \tilde{h}^{(i-1)}_{\sum_{t=1}^j K_t^{(i)}}(\bold{x}) \right)
    -
    g_{j}^{(i)}\left(
    h^{(i-1)}_{\sum_{t=1}^{j-1} K_t^{(i)}+1}(\bold{x}), \dots,
    h^{(i-1)}_{\sum_{t=1}^j K_t^{(i)}}(\bold{x}) \right)
    |
    \\
    &&
    \leq
                            c_{52} \cdot h^{-p_{j}^{(i)}/K_j^{(i)}}
                            + c_{51} \cdot n^{8^{(\sum_{s=1}^{i-1} \tilde{N}_s +j) \cdot (2 \cdot q_{max} \cdot K_{max}+4)}} \cdot d_{model}^{8^{(\sum_{s=1}^{i-1} \tilde{N}_s +j) \cdot (2 \cdot q_{max} \cdot K_{max}+4)}}  \cdot \epsilon
                            \\
                            &&
                            \quad
                            +
    |
g_{j}^{(i)}\left(
    \tilde{h}^{(i-1)}_{\sum_{t=1}^{j-1} K_t^{(i)}+1}(\bold{x}), \dots,
    \tilde{h}^{(i-1)}_{\sum_{t=1}^j K_t^{(i)}}(\bold{x}) \right)
    -
    g_{j}^{(i)}\left(
    h^{(i-1)}_{\sum_{t=1}^{j-1} K_t^{(i)}+1}(\bold{x}), \dots,
    h^{(i-1)}_{\sum_{t=1}^j K_t^{(i)}}(\bold{x}) \right)
    |
    \\
    &&
    \leq
                            c_{52} \cdot h^{-p_{j}^{(i)}/K_j^{(i)}} 
                            +
                             c_{51} \cdot  n^{8^{(\sum_{s=1}^{i-1} \tilde{N}_s +j) \cdot (2 \cdot q_{max} \cdot K_{max}+4)}} \cdot d_{model}^{8^{(\sum_{s=1}^{i-1} \tilde{N}_s +j) \cdot (2 \cdot q_{max} \cdot K_{max}+4)}}  \cdot \epsilon
                             \\
                             &&
                             \quad
                            + c_{53} \cdot \sum_{s=1}^{K_j^{(i)}}
                            |\tilde{h}^{(i-1)}_{\sum_{t=1}^{j-1} K_t^{(i)}+s}
                            (\bold{x})
                            - h^{(i-1)}_{\sum_{t=1}^{j-1} K_t^{(i)}+s}
                            (\bold{x})
         |,
    \end{eqnarray*}
  where the last inequality follows from 
  the Lipschitz continuity of $g_{j}^{(i)}$. Together with
  \[
      |
    \tilde{h}_j^{(1)}(\bold{x})
    -
    h_j^{(1)}(\bold{x})
    |
    \leq
     c_{54} \cdot h^{-p_{j}^{(1)}/K_j^{(1)}}+c_{55} \cdot  n^{8^{j \cdot (2 \cdot q_{max} \cdot K_{max}+4)}} \cdot d_{model}^{8^{j \cdot (2 \cdot q_{max} \cdot K_{max}+4)}}  \cdot \epsilon,
\]
which follows again from (\ref{pth3eq2}), an easy induction shows
\begin{eqnarray*}
|    \tilde{h}_1^{(\kappa)}(\bold{x})
    -
    h_1^{(\kappa)}(\bold{x})
    |
    &
    \leq &
    c_{56} \cdot (K_{max}+1)^\kappa \cdot
    \max_{j,i} h^{-p_j^{(i)}/K_j^{(i)}} \\
    && +
  c_{57} \cdot n^{8^{(\sum_{s=1}^{\kappa} \tilde{N}_s) \cdot (2 \cdot q_{max} \cdot K_{max} +4)}} \cdot d_{model}^{8^{(\sum_{s=1}^{\kappa} \tilde{N}_s) \cdot (2 \cdot q_{max} \cdot K_{max} +4)}} \cdot \epsilon
.   
\end{eqnarray*}

                        \hfill $\Box$

                        \noindent
                        \begin{remark}
                            It follows from the proof of Theorem \ref{th3}
                            (i.e., in particular from the proof of Lemma \ref{le9new}) that even if $\epsilon$ does not satisfy (\ref{pth3eq2}) then
                            the all maximal attentions are attended
                            at some data-independent indices provided
                            $0 \leq \epsilon \leq 1/c_{58}$ holds.
                            \end{remark}

\begin{lemma}
  \label{le12new}
  Set
  \[
f(z)=\begin{cases}
\infty & ,z=1 \\
\log \frac{z}{1-z} & ,0<z<1 \\
- \infty & ,z=0,
\end{cases}
\]
let $K \in \N$ with $K \geq 6$ and let $A \geq 1$. Let $m:\R^{d \cdot l} \rightarrow [0,1]$
and let
$\bar{g}:\R^{d \cdot l} \rightarrow \R$ such that
$\|\bar{g}-m\|_{\infty,[-A,A]^{d \cdot l}} \leq \epsilon$ for some
\[
0 \leq \epsilon \leq \frac{1}{K}.
\]
Then there exists a neural
network $\bar{f}:\R \rightarrow \R$ with ReLU activation function,
and one hidden layer with $3 \cdot K + 9$ neurons,
where all the weights are bounded in absolute value by $K$,
such that for each network
$\tilde{f}:\R \rightarrow \R$
which has the same structure and which
has weights which are in supremum norm not more than
\[
0 \leq  \bar{\epsilon} \leq 1
\]
away from the weights of the above network
we have
\begin{eqnarray*}
  &&
\sup_{x \in [-A,A]^{d \cdot l}}
\Bigg(
\left|
m(x) \cdot
\left(
\varphi(\tilde{f}(\bar{g}( x))-
\varphi(f(m(x))
\right)
\right|
\\
&&
\hspace*{4cm}
+
\left|
(1-m(x)) \cdot
\left(
\varphi(-\tilde{f}(\bar{g}(x)))-
\varphi(-f(m(x))
\right)
\right|
\Bigg)
\\
&&
\leq
c_{58} \cdot \left(\frac{\log K}{K} + \epsilon\right) +
132 \cdot K^2 \cdot \bar{\epsilon}
\end{eqnarray*}
and
\[
\sup_{x \in [-A,A]^{d \cdot l}}
|\tilde{f}(\bar{g}(x))| \leq \log K + 11 \cdot (3 \cdot K+9) \cdot K \cdot \bar{\epsilon}.
\]
\end{lemma}

\noindent
    {\bf Proof.}
In the {\it first part part of the proof} we show the assertion
in case $\bar{\epsilon}=0$.

    For $k \in \{-1,0, \dots, K+1\}$ define
    \[
    B_k(z) =
    \begin{cases}
      0 & ,z < \frac{k-1}{K} \\
     K \cdot( z- \frac{k-1}{K})  & ,\frac{k-1}{K} \leq z < \frac{k}{K} \\
     K \cdot (\frac{k+1}{K}-z)  & ,\frac{k}{K} \leq z < \frac{k+1}{K} \\
     0 &, z \geq \frac{k+1}{K},
      \end{cases}
    \]
    (which implies $B_k(k/K)=1$ and $B_k(j/K)=0$ for $j \in \Z \setminus \{k\}$)
    and set
    \begin{eqnarray*}
    \bar{f}(z)&=&f(1/K) \cdot (B_{-1}(z)+B_0(z)) + \sum_{k=1}^{K-1} f(k/K) \cdot B_k(z)
    + f(1-1/K) \cdot (B_K(z)+ B_{K+1}(z))\\
    &=:&
    \sum_{k=-1}^{K+1} a_k \cdot B_k(z).
    \end{eqnarray*}
    Then $\bar{f}$ interpolates the points $(-1/K, f(1/K))$, $(0,f(1/K))$, $(1/K,f(1/K))$,
    $(2/K,f(2/K))$, \dots, 
    $((K-1)/K,f((K-1)/K))$, $(1,f((K-1)/K))$ and $(1+1/K,f((K-1)/K))$,
    is zero outside of $(-2/K,1+2/K)$
    and is linear on each interval $[k/K, (k+1)/K]$
    $(k \in \{-2, \dots, K+1\})$.
    In particular this implies
    \[
    \sup_{x \in [-A,A]^{d \cdot l}}
|\tilde{f}(\bar{g}(x))|
\leq
\max_{k=1, \dots, K-1} |f(k/K)| \leq \log K.
\]
    Because of
    \[
    B_k(z)=\sigma \left( K \cdot\left( z- \frac{k-1}{K}\right)\right) - 2
    \cdot \sigma \left( K \cdot \left(z-\frac{k}{K}\right)\right) +
    \sigma\left(K \cdot \left(z-\frac{k+1}{K}\right)\right),
    \]
    $\bar{f}$ can be computed by a neural network with
    ReLU activation function and one hidden layer with $3 \cdot (K+3)=3 \cdot K + 9$
    neurons.     
   Set
\[
h_1(z)=\varphi(f(z)) = \log \left( 1 + \exp \left(
- \log \frac{z}{1-z} \right) \right)
= \log \left( 1 + \frac{1-z}{z}  \right) = - \log z
\]
and
\[
h_2(z)=\varphi(-f(z))
=
 \log \left( 1 + \exp \left(
 \log \frac{z}{1-z} \right) \right)
 =
 - \log (1-z).
\]
First we consider the case $m(\bx) \in [0,2/K]$, which implies
\[
f(m(\bx)) \leq f(2/K)=- \log(K/2-1)<0.
\]
In this case we have $-1/K \leq \bar{g}(\bx) \leq 3/K$ and
\[
- \log (K-1)=f(1/K) \leq \bar{f}(\bar{g}(\bx)) \leq f(3/K) = - \log (K/3-1)
\]
(where we have used that $\bar{f}$ is monotone increasing and
satisfies $\bar{f}(- \frac{1}{K})=f(\frac{1}{K})$ and
$\bar{f}(\frac{3}{K})=f(\frac{3}{K})$).
Consequently we get 
\begin{eqnarray*}
\left|
m(\bx) \cdot
\left(
\varphi(\bar{f}(\bar{g}(\bx))-
\varphi(f(m(\bx))
\right)
\right|
&\leq&  m(\bx) \cdot \varphi(\bar{f}(\bar{g}(\bx))
+
m(\bx) \cdot h_1(m(\bx)) \\
& \leq &
\frac{2}{K} \cdot \log(1+\exp(\log(K-1)))
+
m(\bx) \cdot \log (\frac{1}{m(\bx)}) \\
& \leq & 4 \cdot \frac{\log K}{K} 
\end{eqnarray*}
(where we have used the inequality $z \cdot \log(1/z) \leq (2/K) \cdot \log (K/2)$ for $0<z<2/K$)
and
\begin{eqnarray*}
  &&
\left|
(1-m(\bx)) \cdot
\left(
\varphi(-\bar{f}(\bar{g}(\bx))-
\varphi(-f(m(\bx))
\right)
\right|
\\
&&\leq  \varphi(-\bar{f}(\bar{g}(\bx))
+
\varphi(-f(m(\bx)) \\
&&=
\log(1+\exp(\bar{f}(\bar{g}(\bx))))
+
\log(1+\exp(f(m(\bx))))
\\
&&\leq
\log (1 + \exp(- \log(K/3-1)))
+
\log(1+\exp(-\log(K/2-1)))
\\
&& \leq  2 \cdot \exp(-\log(K/3-1)) = \frac{
6
}{
K-3
}.
\end{eqnarray*}
Similarly we get in case $m(\bx) \geq 1-2/K$
\begin{eqnarray*}
  &&
  \left|
m(\bx) \cdot
\left(
\varphi(\bar{f}(\bar{g}(\bx))-
\varphi(f(m(\bx))
\right)
\right|
+
\left|
(1-m(\bx)) \cdot
\left(
\varphi(-\bar{f}(\bar{g}(\bx))-
\varphi(-f(m(\bx))
\right)
\right|
\\
&&
\leq
12 \cdot \frac{\log K}{K-3}.
\end{eqnarray*}
Hence it suffices to show
\begin{eqnarray*}
  &&
\sup_{\bx \in \Rd, \atop m(\bx) \in [2/K,1-2/K]}
\Bigg(
\left|
m(\bx) \cdot
\left(
\varphi(\bar{f}(\bar{g}(\bx))-
\varphi(f(m(\bx))
\right)
\right|
\\
&&
\hspace*{3cm}
+
\left|
(1-m(\bx)) \cdot
\left(
\varphi(-\bar{f}(\bar{g}(\bx))-
\varphi(-f(m(\bx))
\right)
\right|
\Bigg)
\\
&&
\leq
c_{59} \cdot (\frac{\log K}{K} + \epsilon).
\end{eqnarray*}

By the monotonicity of $f$, $|f^\prime(z)|=\frac{1}{z \cdot (1-z)} \geq 1$ for
$z \in (0,1)$, the mean value theorem and the definition of $\bar{f}$ 
    we conclude that for any $\bx \in \R^{d \cdot l}$ with
    $m(\bx) \in [2/K,1-2/K]$
    we find $\xi_x, \delta_{\bx} \in \R$
    with $|\xi_x| \leq \frac{1}{K}$,
    $|\delta_{\bx}| \leq \frac{1}{K} + \epsilon$ and
    $m(\bx)+\delta_{\bx} \in [1/K,1-1/K]$
    such that 
    \begin{equation}
      \label{ple12neweq1}
      \bar{f}(\bar{g}(\bx))
      =
      f(\bar{g}(\bx)+\xi_x)
      =
      f(m(\bx) + \delta_{\bx}).
    \end{equation}
This implies
\begin{eqnarray*}
  &&
\sup_{\bx \in \Rd, \atop m(\bx) \in [2/K,1-2/K]}
\Bigg(
\left|
m(\bx) \cdot
\left(
\varphi(\bar{f}(\bar{g}(\bx))-
\varphi(f(m(\bx))
\right)
\right|
\\
&&
\hspace*{3cm}
+
\left|
(1-m(\bx)) \cdot
\left(
\varphi(-\bar{f}(\bar{g}(\bx))-
\varphi(-f(m(\bx))
\right)
\right|
\Bigg)
\\
&&
=
\sup_{\bx \in \Rd, \atop m(\bx) \in [2/K,1-2/K]}
\Bigg(
|m(\bx)| \cdot |h_1(m(\bx)+\delta_{\bx})-h_1(m(\bx))|
\\
&&
\hspace*{3cm}
+
|1-m(\bx)| \cdot |h_2(m(\bx)+\delta_{\bx})-h_2(m(\bx))|
\Bigg).
\end{eqnarray*}
Consequently it suffices to show that there exist constants
$c_{60}, c_{61}>0$ such that we have for any $z \in [2/K,1-2/K]$ and any $\delta \in \R$
with $|\delta| \leq \frac{1}{K} + \epsilon$ and $z+\delta \in [1/K,1-1/K]$
\begin{equation}
  \label{ple12neweq2}
  |z| \cdot |h_1(z+\delta)-h_1(z)| \leq c_{60} \cdot
  \left( \frac{1}{K} + \epsilon \right)
  \end{equation}
and
\begin{equation}
  \label{ple12neweq3}
    |1-z| \cdot |h_2(z+\delta)-h_2(z)| \leq c_{61} \cdot
  \left( \frac{1}{K} + \epsilon \right).
  \end{equation}
Obviously
\[
h_1^\prime (z) = - \frac{1}{z}.
\]
By the mean value theorem we get for some $\xi \in [\min\{ z+\delta, z\},
  \max\{ z+\delta,z\}] $
\[
|z| \cdot |h_1(z+\delta)-h_1(z)|
=
|z| \cdot \frac{1}{|\xi|} \cdot |\delta|
\leq
4 \cdot |\delta|
\leq
4 \cdot
  \left( \frac{1}{K} + \epsilon \right),
\]
where we have used that $z, z+\delta \in [1/K,1-1/K]$ and $|\delta| \leq 2/K$
imply $4 |\xi| \geq |z|$.

In the same way we get
\[
h_2^\prime (z) = \frac{1}{1-z}
\]
and
\[
|1-z| \cdot |h_2(z+\delta)-h_2(z)|
=
|1-z| \cdot \frac{1}{|1-\xi|} \cdot |\delta|
\leq
4 \cdot |\delta|
\leq
4 \cdot
  \left( \frac{1}{K} + \epsilon \right).
\]

In the {\it second step of the proof} we show
that if a network $\tilde{f}$ has the same structure as the network
$\bar{f}$ in the first step of the proof and if the supremum norm
distance between the weights of $\tilde{f}$ and $\bar{f}$ is at most
$\epsilon$, then we have:
\[
\sup_{x \in \R^{d \cdot l}}
\left|
\tilde{f}(\bar{g}(x)-\bar{f}(\bar{g}(x))
\right|
\leq
11 \cdot (3 \cdot K+9) \cdot K \cdot \bar{\epsilon}
.
\]

Let
\[
f(z)=
\sum_{j=1}^{J_n} v_{j}^{(1)} \cdot
\sigma \left(
v_{j,1}^{(0)} \cdot z + v_{j,0}^{(0)}
\right)
\]
be a neural network with one hidden layer with $J_n$ neurons,
where all the weights are bounded in absolute
value by $\beta=K$.
It suffices to show that for any $z \in [-1,2]$
and any network $\tilde{f}$
which has the same structure as $f$ and where the weights are
in supremum norm not further away from the weights of $f$ than
$\bar{\epsilon}$, it holds
\[
| \tilde{f}(z)-f(z)| \leq 11 \cdot \beta \cdot J_n \cdot \bar{\epsilon}.
\] 
To prove this we observe
\begin{eqnarray*}
&&
|
\tilde{v}_{i,1}^{(0)} \cdot z
+ \tilde{v}_{i,0}^{(0)}
-
(v_{i,1}^{(0)} \cdot z 
+ v_{i,0}^{(0)}
)|
\leq |(\tilde{v}_{i,1}^{(0)} - v_{i,1}^{(0)} ) \cdot z| +
\bar{\epsilon}
\leq
2 \cdot \bar{\epsilon} +
\bar{\epsilon} = 3 \cdot \bar{\epsilon},
\end{eqnarray*}
which implies
\[
|
\sigma(\tilde{v}_{i,1}^{(0)} \cdot \tilde{z}
+ \tilde{v}_{i,0}^{(0)})
-
\sigma
(v_{i,1}^{(0)} \cdot z 
+ v_{i,0}^{(0)}
)|
\leq
3 \cdot \bar{\epsilon},
\]
and
\[
|
\sigma(\tilde{v}_{i,1}^{(0)} \cdot \tilde{z}
+ \tilde{v}_{i,0}^{(0)})
|
\leq
3 \cdot \bar{\epsilon}
+
| \sigma
(v_{i,1}^{(0)} \cdot z 
+ v_{i,0}^{(0)}
)|
\leq
3 \cdot \bar{\epsilon}
+
5 \beta
\leq 8 \beta.
\]
Hence we have
\begin{eqnarray*}
  | \tilde{f}(z)-f(z)|
  &=&
  \left|
\sum_{j=1}^{J_n} \tilde{v}_{j}^{(1)} \cdot
\sigma \left(
\tilde{v}_{j,1}^{(0)} \cdot z + \tilde{v}_{j,0}^{(0)}
\right)
  -
\sum_{j=1}^{J_n} v_{j}^{(1)} \cdot
\sigma \left(
v_{j,1}^{(0)} \cdot z + v_{j,0}^{(0)}
\right)  
\right|
\\
& \leq &
\sum_{j=1}^{J_n} |\tilde{v}_{j}^{(1)}-v_j^{(1)}| \cdot
\sigma \left(
\tilde{v}_{j,1}^{(0)} \cdot z + \tilde{v}_{j,0}^{(0)}
\right)
\\
&&
+
\sum_{j=1}^{J_n} |v_{j}^{(1)}| \cdot
\left|
\sigma \left(
\tilde{v}_{j,1}^{(0)} \cdot z + \tilde{v}_{j,0}^{(0)}
\right)
-
\sigma \left(
v_{j,1}^{(0)} \cdot z + v_{j,0}^{(0)}
\right)  
\right|
\\
& \leq &
8 \cdot \beta \cdot J_n \cdot \bar{\epsilon}
+
\beta \cdot J_n \cdot 3 \cdot \bar{\epsilon} 
\leq
11 \cdot \beta \cdot J_n \cdot \bar{\epsilon},
  \end{eqnarray*}
which yields the assertion.

Since $\varphi$ is Lipschitz continuous the assertion of
Lemma \ref{le12new} follows from steps 1 and 2.
\quad \quad \hfill $\Box$

\begin{lemma}
  \label{le13}
  Let $A \geq 1$ and let $0 \leq \epsilon \leq 1/(2 \cdot c_{62})$.
  Let $f_{\btheta}$ be a transformer classifier defined as in Section \ref{se2}
  where the weights in all attention units and in all piecewise
  feedforward networks are in supremum norm not further away than
  $1/(2 \cdot c_{62})$ from
  the weights of the transformer network in Theorem \ref{th3},
  and where the weights in the feedfoward network are not further
  away than $\epsilon$ from the weights of the feedforward neural network
  in Lemma \ref{le12new}. Let $f_{\tilde{\btheta}}$ be a transformer classifier
  of the same form which satisfies
  \[
  \|\tilde{\btheta}-\btheta\|_\infty \leq  \epsilon.
  \]
  Assume $d_{ff}=2 \cdot h + 2$, $h \leq c_{64} \cdot n$,
  $d_{model}=h \cdot I$ and $I= \lceil \log n \rceil$.
  Then we have
  for $c_{65}, c_{66}>0$ sufficiently large
  \[
  \| f_{\tilde{\btheta}} - f_{\btheta}\|_{[-A,A]^{d \cdot l},\infty} \leq c_{65} \cdot n^{c_{66}} \cdot
    \|\tilde{\btheta}-\btheta\|_\infty.
    \]
  \end{lemma}

\noindent
    {\bf Proof.}
    Since the weights in the transformer classifiers  $f_{\btheta}$
    and $f_{\btheta^*}$ are not further away than
    $1/c_{63}$
    from the weights of the transformer network in Theorem \ref{th3},
    it follows from the proof of Theorem \ref{th3} that in both transformer network all maximal
    attention are attained at the same indices, namely at the indices
    where the transformer network in Theorem \ref{th3} attains its
    maximal attentions (cf., Remark 6). Consequently we can ignore the selection of the
    maximal attention in the rest of the proof.

    Let
    \[
      q_{k,r-1,s,i} = W_{query,k,r,s} \cdot z_{k,r-1,i}, \quad
      k_{k,r-1,s,i} = W_{key,k,r,s} \cdot z_{k,r-1,i},
      \]
      \[
  v_{k,r-1,s,i} = W_{value,k,r,s} \cdot z_{k,r-1,i}, \quad
  \tilde{q}_{k,r-1,s,i} = \tilde{W}_{query,k,r,s} \cdot \tilde{z}_{k,r-1,i},
  \]
  \[
  \tilde{k}_{k,r-1,s,i} = \tilde{W}_{key,k,r,s} \cdot \tilde{z}_{k,r-1,i}, \quad
  \tilde{v}_{k,r-1,s,i} = \tilde{W}_{value,k,r,s} \cdot \tilde{z}_{k,r-1,i} 
  \]
  where all the weights in the matrices above are bounded in absolute value
  by $B \geq 1$, and set
  \[
y_{k,r,i}=z_{k,r-1,i}+
v_{k,r-1,s,r_1} \cdot  <q_{k,r-1,s,i}, k_{k,r-1,s,r_1}>
\]
and
\[
\tilde{y}_{k,r,i}=\tilde{z}_{k,r-1,i}+
\tilde{v}_{k,r-1,s,r_1} \cdot  <\tilde{q}_{k,r-1,s,i}, \tilde{k}_{k,r-1,s,r_1}>.
\]
In the {\it first step of the proof} we show
\begin{eqnarray*}
  &&
  \|\tilde{y}_{k,r,i} - y_{k,r,i} \|_\infty
  \\
  &&
  \leq
  c_{67} \cdot d_{key} \cdot d_{model}^3 \cdot B^2
  \cdot \left( \max\{\|z_{k,r-1,i}\|_\infty, \|\tilde{z}_{k,r-1,i}\|_\infty,1\}
  \right)^3
  \\
  &&
  \hspace*{1cm}
  \cdot
  \max\Big\{
  \| \tilde{W}_{query,k,r-1,s} - W_{query,k,r-1,s} \|_\infty,
  \| \tilde{W}_{key,k,r-1,s} - W_{key,k,r-1,s} \|_\infty,
  \\
  &&
  \hspace*{2cm}
  \| \tilde{W}_{value,k,r-1,s} - W_{value,k,r-1,s} \|_\infty
  \Big\}
  \\
  &&
  \quad
  +
   c_{68} \cdot d_{key} \cdot d_{model}^3 \cdot B^3
\cdot \left( \max\{\|z_{k,r-1,i}\|_\infty, \|\tilde{z}_{k,r-1,i}\|_\infty,1\}
  \right)^2
  \cdot \| \tilde{z}_{k,r-1} - z_{k,r-1} \|_{\infty}.
\end{eqnarray*}

We have
\begin{eqnarray*}
  &&
  \|\tilde{y}_{k,r,i} - y_{k,r,i} \|_\infty
  \\
  &&
  \leq
  \|\tilde{z}_{k,r,i} - z_{k,r,i} \|_\infty
  +
  \|\tilde{v}_{k,r,r_1} - v_{k,r,r_1} \|_\infty
  \cdot
  |<\tilde{q}_{k,r,i}, \tilde{k}_{k,r,r_1}>|
  \\
  &&
  \quad
  +
  \|v_{k,r,r_1} \|_\infty
  \cdot
  |<\tilde{q}_{k,r,i}, \tilde{k}_{k,r,r_1}>
  -
  <q_{k,r,i}, k_{k,r,r_1}>|.
  \end{eqnarray*}
With
\begin{eqnarray*}
  &&
  \|\tilde{v}_{k,r,r_1} - v_{k,r,r_1} \|_\infty
  \\
  &&
  \leq
  \|
(\tilde{W}_{value,k,r,s} - W_{value,k,r,s}) \cdot \tilde{z}_{k,r-1,r_1}
  \|_\infty
  +
  \|
W_{value,k,r,s} \cdot (\tilde{z}_{k,r-1,r_1}-z_{k,r-1,r_1})
  \|_\infty
  \\
  &&
  \leq
  d_{model} \cdot \|\tilde{W}_{value,k,r,s} - W_{value,k,r,s}\|_\infty \cdot
  \|\tilde{z}_{k,r-1,r_1}\|_\infty
  \\
  &&
  \quad
  + d_{model} \cdot B \cdot \|\tilde{z}_{k,r-1,r_1}-z_{k,r-1,r_1}\|_\infty,
  \end{eqnarray*}
\begin{eqnarray*}
  &&
    |<\tilde{q}_{k,r,i}, \tilde{k}_{k,r,r_1}>|
    \leq
    d_{key} \cdot B^2 \cdot d_{model} \cdot \|\tilde{z}_{k,r-1,r_1}\|_\infty^2,
\end{eqnarray*}
\[
\|v_{k,r,i}\|_\infty \leq d_{model} \cdot B \cdot \|z_{k,r-1,i}\|_\infty
\]
and
\begin{eqnarray*}
  &&
|<\tilde{q}_{k,r,i}, \tilde{k}_{k,r,r_1}>
  -
  <q_{k,r,i}, k_{k,r,r_1}>|
  \\
  &&
  \leq
  |<\tilde{q}_{k,r,i} - q_{k,r,i}, \tilde{k}_{k,r,r_1}>|
  +
  |<q_{k,r,i}, \tilde{k}_{k,r,r_1}- k_{k,r,r_1}>|
  \\
  &&
  \leq
  d_{key} \cdot \Big(
  \|
(\tilde{W}_{query,k,r,s} - W_{query,k,r,s}) \cdot \tilde{z}_{k,r-1,i}
  \|_\infty
  \\
  &&
  \hspace*{2cm}
  +
  \|
W_{query,k,r,s} \cdot (\tilde{z}_{k,r-1,i} - z_{k,r-1,i})
  \|_\infty
  \Big) \cdot \| \tilde{k}_{k,r,r_1}\|_\infty
  \\
  &&
  \quad
  +
  d_{key} \cdot \|q_{k,r,i}\|_\infty
  \cdot
  \Big(
  \|
(\tilde{W}_{key,k,r,s} - W_{key,k,r,s}) \cdot \tilde{z}_{k,r-1,i}
  \|_\infty
  \\
  &&
  \hspace*{2cm}
  +
  \|
  W_{key,k,r,s}
  \cdot (\tilde{z}_{k,r-1,i} - z_{k,r-1,i})
  \|_\infty
  \Big)
  \\
  &&
  \leq
  d_{key} \cdot
  ( d_{model} \cdot \|\tilde{W}_{query,k,r,s} - W_{query,k,r,s}\|_\infty
  \cdot
  \|\tilde{z}_{k,r-1}\|_\infty
  \\
  &&
  \hspace*{2cm}
  + d_{model} \cdot B \cdot \| \tilde{z}_{k,r-1} - z_{k,r-1}\|_\infty)
  \cdot d_{model} \cdot B \cdot \| \tilde{z}_{k,r-1} \|_\infty
  \\
  &&
  \quad
  +
  d_{key} \cdot d_{model} \cdot B \cdot  \| z_{k,r-1} \|_\infty
  \cdot
  (d_{model} \cdot \|\tilde{W}_{key,k,r,s} - W_{key,k,r,s}\|_\infty
  \cdot \| \tilde{z}_{k,r-1} \|_\infty
  \\
  &&
  \hspace*{2cm}
  +
  d_{model} \cdot B \cdot \| \tilde{z}_{k,r-1} - z_{k,r-1}\|_\infty)
  \end{eqnarray*}
we get the assertion.

Set
\[
  z_{r,s}=y_{r,s}+W_{r,2} \cdot \sigma \left(
  W_{r,1} \cdot y_{r,s} + b_{r,1}
 \right) + b_{ r,2}
\]
and
\[
 \tilde{z}_{r,s}=\tilde{y}_{r,s}+\tilde{W}_{r,2} \cdot \sigma \left(
  \tilde{W}_{r,1} \cdot \tilde{y}_{r,s} + \tilde{b}_{r,1}
  \right) + \tilde{b}_{r,2},
\]
where all weights of the neural networks above are bounded
in absolute value by $B \geq 1$.
In the {\it second step of the proof} we show
\begin{eqnarray*}
  &&
  \| z_{r,s}-  \tilde{z}_{r,s}\|_\infty
  \\
  &&
\leq
c_{69} \cdot d_{ff} \cdot d_{model} \cdot B
\cdot
\max\{\|y_{r,s}\|_\infty, \|\tilde{y}_{r,s}\|_\infty,1\}
\\
&&
\hspace*{2cm}
\cdot
\max\left\{
\|\tilde{W}_{r,2} - W_{r,2}\|_\infty,
\|\tilde{W}_{r,1} - W_{r,1}\|_\infty,
\|\tilde{b}_{r,2} - b_{r,2}\|_\infty,
\|\tilde{b}_{r,1} - b_{r,1}\|_\infty
\right\}
\\
&&
\quad
+
c_{70} \cdot d_{ff} \cdot d_{model} \cdot B^2
\cdot \|\tilde{y}_{r,s} - y_{r,s} \|_\infty.
\end{eqnarray*}

We have
\begin{eqnarray*}
  &&
  \| z_{r,s}-  \tilde{z}_{r,s}\|_\infty
  \\
  &&
  \leq
  \| y_{r,s}-  \tilde{y}_{r,s}\|_\infty
  +
  \|
\tilde{W}_{r,2} \cdot \sigma \left(
  \tilde{W}_{r,1} \cdot \tilde{y}_{r,s} + \tilde{b}_{r,1}
  \right)
  -
W_{r,2} \cdot \sigma \left(
  W_{r,1} \cdot y_{r,s} + b_{r,1}
 \right)  
 \|_\infty
 \\
 &&
 \hspace*{1cm}
 +
 \|
\tilde{b}_{r,2} - b_{r,2}
\|_\infty
\\
&&
\leq
  \| y_{r,s}-  \tilde{y}_{r,s}\|_\infty
  +
  \|
(\tilde{W}_{r,2}- W_{r,2}) \cdot \sigma \left(
  \tilde{W}_{r,1} \cdot \tilde{y}_{r,s} + \tilde{b}_{r,1}
  \right)
  \|_\infty
  \\
  &&
  \quad
  +
  \|
W_{r,2} \cdot (\sigma \left(
  \tilde{W}_{r,1} \cdot \tilde{y}_{r,s} + \tilde{b}_{r,1}
  \right)
  -
  \sigma \left(
  W_{r,1} \cdot y_{r,s} + b_{r,1}
 \right)  )
 \|_\infty
 +
 \|
\tilde{b}_{r,2} - b_{r,2}
\|_\infty
\\
&&
\leq
  \| y_{r,s}-  \tilde{y}_{r,s}\|_\infty
  +
  d_{ff} \cdot
  \|
  \tilde{W}_{r,2}- W_{r,2}
  \|_\infty
  \cdot
  \|
  \tilde{W}_{r,1} \cdot \tilde{y}_{r,s} + \tilde{b}_{r,1}
  \|_\infty
  \\
  &&
  \quad
  + d_{ff} \cdot B \cdot
  \| \tilde{W}_{r,1} \cdot \tilde{y}_{r,s}
  - W_{r,1} \cdot y_{r,s}+ \tilde{b}_{r,1} - b_{r,1}
  \|_\infty + \| \tilde{b}_{r,2} - b_{r,2}\|_\infty.
\end{eqnarray*}
Using
\[
\|
  \tilde{W}_{r,1} \cdot \tilde{y}_{r,s} + \tilde{b}_{r,1}
  \|_\infty
  \leq
  d_{model} \cdot B \cdot \| \tilde{y}_{r,s}\|_\infty + B
\]
and
\begin{eqnarray*}
  &&
\|\tilde{W}_{r,1} \cdot \tilde{y}_{r,s}
- W_{r,1} \cdot y_{r,s}\|_\infty
\\
&&
\leq
\|( \tilde{W}_{r,1} - W_{r,1}) \cdot \tilde{y}_{r,s} \|_\infty
+
\| W_{r,1} \cdot (\tilde{y}_{r,s}-y_{r,s}) \|_\infty
\\
&&
\leq
d_{model} \cdot \| \tilde{W}_{r,1} - W_{r,1} \|_\infty
\cdot \|\tilde{y}_{r,s} \|_\infty
+
d_{model} \cdot B \cdot \| \tilde{y}_{r,s}-y_{r,s} \|_\infty
  \end{eqnarray*}
we get the assertion.

Let
\[
f(z)=
\sum_{j=1}^{J_n} v_{j}^{(1)} \cdot
\sigma \left(
v_{j,1}^{(0)} \cdot z + v_{j,0}^{(0)}
\right)
\]
and
\[
\tilde{f}(\tilde{z})=
\sum_{j=1}^{J_n} \tilde{v}_{j}^{(1)} \cdot
\sigma \left(
\tilde{v}_{j,1}^{(0)} \cdot \tilde{z} + \tilde{v}_{j,0}^{(0)}
\right),
\]
where all the weights of the networks above are bounded in absolute
value by $B \geq 1$.
In the {\it third part of the proof} we show
\begin{eqnarray*}
|\tilde{f}(\tilde{z})-f(z)|
&\leq&
c_{71} \cdot J_n \cdot B \cdot \max\{\tilde{z},z, 1\}
\cdot \max\{|\tilde{v}_{j}^{(1)}-v_{j}^{(1)}|,
|\tilde{v}_{j,1}^{(0)}-v_{j,1}^{(0)}|,
|\tilde{v}_{j,0}^{(0)}-v_{j,0}^{(0)}|
\}
\\
&&
+
c_{72} \cdot J_n \cdot B^2 
\cdot |\tilde{z} - z|.
\end{eqnarray*}

We have
\begin{eqnarray*}
  &&
  |\tilde{f}(\tilde{z})-f(z)|
  \\
  &&
  \leq
  \sum_{j=1}^{J_n}
|
  \tilde{v}_{j}^{(1)} \cdot
\sigma \left(
\tilde{v}_{j,1}^{(0)} \cdot \tilde{z} + \tilde{v}_{j,0}^{(0)}
\right)
-
v_{j}^{(1)} \cdot
\sigma \left(
v_{j,1}^{(0)} \cdot z + v_{j,0}^{(0)}
\right)
|
\\
&&
\leq
J_n \cdot \max_{j=1, \dots, J_n}
\Big(
|\tilde{v}_{j}^{(1)} -v_{j}^{(1)} |\cdot
\sigma \left(
\tilde{v}_{j,1}^{(0)} \cdot \tilde{z} + \tilde{v}_{j,0}^{(0)}
\right) 
\\
&&
\hspace*{3cm}
+
 |v_{j}^{(1)}| \cdot
|\sigma \left(
\tilde{v}_{j,1}^{(0)} \cdot \tilde{z} + \tilde{v}_{j,0}^{(0)}
\right)
-
\sigma \left(
v_{j,1}^{(0)} \cdot z + v_{j,0}^{(0)}
\right)
|
\Big).
\end{eqnarray*}
With
\[
\sigma \left(
\tilde{v}_{j,1}^{(0)} \cdot \tilde{z} + \tilde{v}_{j,0}^{(0)}
\right) 
\leq
|\tilde{v}_{j,1}^{(0)} \cdot \tilde{z} + \tilde{v}_{j,0}^{(0)}|
\leq
B \cdot |\tilde{z}| + B
\]
and
\begin{eqnarray*}
  &&
|\sigma \left(
\tilde{v}_{j,1}^{(0)} \cdot \tilde{z} + \tilde{v}_{j,0}^{(0)}
\right)
-
\sigma \left(
v_{j,1}^{(0)} \cdot z + v_{j,0}^{(0)}
\right)
|
\\
&&
\leq
|
\tilde{v}_{j,1}^{(0)} \cdot \tilde{z} 
-
v_{j,1}^{(0)} \cdot z
+ \tilde{v}_{j,0}^{(0)}
- v_{j,0}^{(0)}
|
\leq
|\tilde{v}_{j,1}^{(0)} \cdot (\tilde{z}-z)|
+
|(\tilde{v}_{j,1}^{(0)} -v_{j,1}^{(0)}) \cdot z|
+
| \tilde{v}_{j,0}^{(0)}
- v_{j,0}^{(0)}|
\\
&&
\leq
B \cdot |\tilde{z}-z|
+
|\tilde{v}_{j,1}^{(0)} -v_{j,1}^{(0)}| \cdot |z|
+
| \tilde{v}_{j,0}^{(0)}
- v_{j,0}^{(0)}|
  \end{eqnarray*}
we get the assertion.

In the {\it fourth part of the proof} we use the above results to show the
assertion of the lemma.

All weights in the above transformer classifiers are bounded in absolute
value by
\[
c_{73} \cdot n^{c_{74}} + \epsilon \leq 2 \cdot c_{73} \cdot n^{c_{74}} =:B.
\]
Because of
\[
\|y_{k,r}\|_\infty \leq 2 \cdot d_{ff} \cdot d_{key}^2 \cdot d_{model} \cdot B^3
\cdot \max\{\|z_{k,r-1}\|_\infty^3,1\},
\]
\[
\|z_{k,r}\|_\infty \leq 4 \cdot d_{ff} \cdot d_{model} \cdot B^2
\cdot \max\{ \|y_{k,r}\|_\infty,1\}
\]
and
\[
\|z_{0,r}\|_\infty \leq A
\]
an easy induction shows
\[
\|y_{k,r}\|_\infty \leq
128^{3^{2 \cdot r -2}}
\cdot d_{ff}^{4^{2 \cdot r -2}}
\cdot d_{model}^{4^{2 \cdot r -2}}
\cdot B^{9^{2 \cdot r -2}}
\cdot A^{3^r}
\]
and
\[
\|z_{k,r}\|_\infty
\leq
128^{3^{2 \cdot r -2}}
\cdot d_{ff}^{4^{2 \cdot r -2}+1}
\cdot d_{model}^{4^{2 \cdot r -2}+1}
\cdot B^{9^{2 \cdot r -2}+2}
\cdot A^{3^r}
\]
for $r \geq 1$.
The same inequalities also hold for $\tilde{y}_{k,r}$ and $\tilde{z}_{k,r}$.
This implies
\[
\max\left\{
\|y_{k,r}\|_\infty, \|\tilde{y}_{k,r}\|_\infty,
\|z_{k,r}\|_\infty, \|\tilde{z}_{k,r}\|_\infty  
\right\}
\leq c_{75} \cdot n^{c_{76}}
\]
for $r \leq N$, where $c_{75}=c_{75}(N),c_{76}=c_{76}(N)>0$ are finite constants.

Consequently we can conclude from Step 1
\begin{eqnarray*}
  &&
  \|\tilde{y}_{k,r,i} - y_{k,r,i} \|_\infty
  \\
  &&
  \leq
  c_{77} \cdot n^{c_{78}} \cdot
  \max\Big\{
  \| \tilde{W}_{query,k,r-1,s} - W_{query,k,r-1,s} \|,
  \| \tilde{W}_{key,k,r-1,s} - W_{key,k,r-1,s} \|,
  \\
  &&
  \hspace*{1cm}
  \| \tilde{W}_{value,k,r-1,s} - W_{value,k,r-1,s} \|,
  \| \tilde{z}_{k,r-1} - z_{k,r-1} \|_{\infty}
  \Big\},
\end{eqnarray*}
from Step 2
\begin{eqnarray*}
  &&
  \| z_{r,s}-  \tilde{z}_{r,s}\|_\infty
  \\
  &&
\leq
c_{79} \cdot n^{c_{80}} 
\cdot
\max\Big\{
\|\tilde{W}_{r,2} - W_{r,2}\|_\infty,
\|\tilde{W}_{r,1} - W_{r,1}\|_\infty,
\|\tilde{b}_{r,2} - b_{r,2}\|_\infty,
\\
&&
\hspace*{3cm}
\|\tilde{b}_{r,1} - b_{r,1}\|_\infty,
\|\tilde{y}_{r,s} - y_{r,s} \|_\infty
\Big\},
\end{eqnarray*}
and from Step 3
\begin{eqnarray*}
  &&
  \| f_{\tilde{\btheta}} - f_{\btheta}\|_{[-A,A]^{d \cdot l},\infty} \\
  &&
  \leq
  c_{81} \cdot n^{c_{82}} 
  \cdot \max\{|\tilde{v}_{j}^{(1)}-v_{j}^{(1)}|,
|\tilde{v}_{j,1}^{(0)}-v_{j,1}^{(0)}|,
|\tilde{v}_{j,0}^{(0)}-v_{j,0}^{(0)}|,
\|\tilde{z}_{N,1}-z_{N,1}\|_\infty
\}
.
  \end{eqnarray*}
Using these relations recursively we conclude
\[
\| f_{\tilde{\btheta}} - f_{\btheta}\|_{[-A,A]^{d \cdot l},\infty} \leq
  c_{83} \cdot n^{c_{84}}
\]
for $c_{83}, c_{84} >0$ sufficiently large (and depending
on $N$).    \hfill $\Box$

\end{appendix}

\end{document}